\documentclass[11pt]{article}

\usepackage[preprint]{acl}

\usepackage{times}
\usepackage{latexsym}
\usepackage{makecell} 
\usepackage{etoc}
\usepackage{enumitem}
\usepackage[T1]{fontenc}
\usepackage{hyperref}
\usepackage{url}
\usepackage{multirow}
\usepackage{amsmath,amssymb}
\usepackage{algorithm, algorithmic}
\usepackage{booktabs}
\usepackage{tabularx}
\usepackage{makecell}
\usepackage{graphicx}          
\usepackage{xcolor}
\usepackage[table]{xcolor}
\usepackage{enumitem}
\usepackage{listings}
\usepackage{url}
\usepackage{caption}
\usepackage{capt-of}
\usepackage{placeins}
\usepackage[most]{tcolorbox}
\definecolor{promptplaceholder}{RGB}{200,40,40}

  {\par\medskip\noindent
   \begin{tabular}{@{}!{\color{loopred}\vrule width 2.5pt}@{\hspace{11pt}}%
                     p{\dimexpr\linewidth-15pt\relax}@{}}
   \scriptsize\ignorespaces}%
  {\end{tabular}\par\medskip}

  {\par\scriptsizeskip\begin{list}{}{%
     \setlength{\leftmargin}{3.4cm}\setlength{\labelwidth}{3.1cm}%
     \setlength{\labelsep}{0.3cm}\setlength{\itemsep}{1pt}%
     \setlength{\parsep}{0pt}\setlength{\topsep}{2pt}%
     }%
   \footnotesize}%
  {\end{list}\par\scriptsizeskip}

\renewcommand{\arraystretch}{1.15}

\lstdefinestyle{pystyle}{
 language=Python,
 basicstyle=\ttfamily\footnotesize,
 commentstyle=\itshape\color{black!55},
 keywordstyle=\bfseries,
 stringstyle=\color{black!70},
 showstringspaces=false,
 breaklines=true,
 breakatwhitespace=true,
 frame=single,
 framesep=5pt,
 rulecolor=\color{black!25},
 xleftmargin=0pt,
 aboveskip=9pt,
 belowskip=9pt,
 columns=fullflexible,
}

\lstdefinestyle{promptstyle}{
    basicstyle=\ttfamily\scriptsize,
    breaklines=false,
    columns=fullflexible,
    keepspaces=true,
    escapeinside={(*}{*)},
    frame=none,
    xleftmargin=0pt,
    showstringspaces=false
}

\definecolor{keybg}{gray}{0.90}
\newcommand{\thd}[1]{\textbf{#1}}

\newcommand{\ethink}{\texttt{</think>}}
\newcommand{\ckpt}[1]{\texttt{obc\_#1}}
\newcommand{\dns}{\texttt{dense}}

\newcommand{\tabsmall}{\scriptsize\setlength{\tabcolsep}{3.5pt}\renewcommand{\arraystretch}{0.95}}
\newcommand{\num}[1]{\ensuremath{#1}}

\newcommand{\INPUT}{\item[\textbf{Input:}]}
\newcommand{\OUTPUT}{\item[\textbf{Output:}]}

\title{To Reason but Not to Stop:\\ Understanding and Recovering Pruned Large Reasoning Models}

\definecolor{table_green}{RGB}{34,139,34}

\author{
  Minh-Anh Nguyen$^{1}$ \quad
  Nguyen T. Ha Lan$^{2}$ \quad
  Xiyan Fu$^{1}$ \quad
  Dung D. Le$^{2}$ \quad
  Wei Lu$^{1}$ \\
  $^{1}$Nanyang Technological University, Singapore\\
  $^{2}$VinUniversity, Vietnam
}

\definecolor{tgasrow}{RGB}{231,240,253}
\newcommand{\sd}[1]{{\tiny$\pm$#1}}
\newcommand{\gain}[1]{\textsubscript{\textcolor{table_green}{$\uparrow$#1}}}
\begin{document}
\etocdepthtag.toc{mainbody}
\maketitle
 
\begin{abstract}
Weight pruning can degrade the accuracy of large reasoning models (LRMs), yet it remains unclear whether it damages reasoning itself. We find that a substantial fraction of this degradation stems from a failure to \emph{stop} rather than to \emph{reason}. As sparsity increases, pruned LRMs increasingly fail to terminate within the token budget, often after the correct answer has appeared in their reasoning, instead falling into repetitive loops, even though they still recognize where the dense model would stop. We further find that the hidden-state difference between terminating and looping continuations is consistent across pruned checkpoints, suggesting a shared direction associated with committing to an answer. Building on this, we propose Termination-Guided Activation Steering (TGAS), a training-free method that adds this direction to the residual stream at inference time. Across six LRMs, three pruning pipelines, and math and code benchmarks, TGAS improves termination and overall accuracy while reducing repetitive reasoning and generation length, showing that this failure is actionable.
\end{abstract}


\section{Introduction}
\label{sec:intro}

Large reasoning models (LRMs) achieve strong performance by generating long chain-of-thought traces, but this makes inference expensive \citep{zhang2025system,deepseek_r1}. Weight pruning offers a natural approach to reduce cost by zeroing out selected parameters \citep{sparsegpt,wanda}. However, as consistently observed across pruning methods, pruning LRMs can substantially degrade accuracy on reasoning tasks \citep{beware_calibration,prune-RAC,zhang2026when}. Existing efforts have primarily focused on mitigating this degradation through improved pruning calibration \citep{beware_calibration,prune-RAC,zhang2026when}, alongside post-pruning recovery via retraining or fine-tuning \citep{liu2025pat,muralidharan2024compact,kim2024shortened,xia2024sheared}. Yet, it remains unclear whether this degradation reflects a loss of reasoning capability or a different kind of failure.

\begin{figure}[t]
\centering
\includegraphics[width=0.9\linewidth]{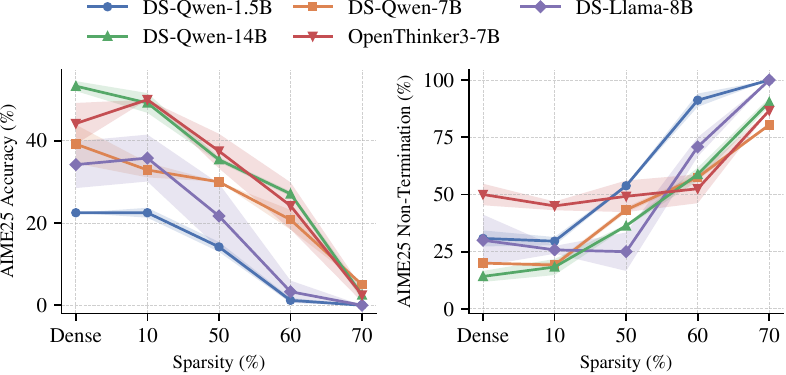}
\caption{Accuracy (left) and non-termination rate (right) on AIME-25 across pruning settings (SparseGPT with OBC calibration, 16{,}384-token budget, averaged over 4 runs). Accuracy decreases as non-termination becomes more frequent.}
\label{fig:acc_and_non-termination}
\end{figure}

We find that much of this degradation stems from a failure to \emph{stop} rather than a failure to \emph{reason}. As sparsity increases, pruned LRMs increasingly fail to terminate and commit to a final answer within the token budget, and accuracy falls accordingly (Figure~\ref{fig:acc_and_non-termination}). Strikingly, many of these non-terminating traces already contain the correct answer, yet the model keeps reasoning until the budget is exhausted. We refer to this behavior as \textbf{\textit{reasoning divergence}}. On problems solved by the dense model, pruned models remain largely accurate whenever they terminate, and non-termination becomes the dominant source of errors as sparsity grows (Section~\ref{sec:obs_1}).

Why do pruned models fail to stop? Pruning does not erase the stopping signal: when replaying dense traces, pruned models still assign high probability to the end-of-thinking token where the dense model stops. Instead, during free-running generation they increasingly revisit earlier reasoning rather than moving forward, and fall into repetitive loops. Moreover, the hidden-state difference between continuations that terminate and those that loop points in a similar direction across pruned checkpoints, suggesting that the shift between looping and committing to an answer is captured by a shared direction in representation space.

Based on this diagnosis, we propose \textit{Termination-Guided Activation Steering} (TGAS), which adds this direction to the residual stream at inference time to steer pruned models toward termination without retraining or weight updates. TGAS consistently improves termination and accuracy across six LRMs, three pruning pipelines, and math and code benchmarks by reducing repetitive reasoning. Our contributions are threefold: (i) We show that a substantial part of pruning-induced degradation in LRMs arises from reasoning divergence, a failure to commit rather than a loss of reasoning ability; (ii) We identify a hidden-state direction that separates terminating from looping continuations, and build on it TGAS, a training-free steering method; (iii) We show that TGAS consistently improves termination and overall accuracy across models, pruning methods, and benchmarks by reducing looping.

\section{Background and Related Work}
\label{sec:background}

\paragraph{Pruning for LRMs.} One-shot unstructured pruning formulates compression as layer-wise reconstruction~\citep{obc}: given a weight matrix $\mathbf{W}_\ell$ and calibration inputs $\mathbf{X}_\ell$, it solves $\min_{\|\hat{\mathbf{W}}_\ell\|_0 \le S} \|(\mathbf{W}_\ell - \hat{\mathbf{W}}_\ell)\mathbf{X}_\ell\|_F^2$, where $S$ is the number of retained weights. We focus on two retraining-free methods: SparseGPT~\citep{sparsegpt}, which uses second-order information to compensate for pruning errors, and Wanda~\citep{wanda}, which scores weights by magnitude and input-activation norm. To better capture long reasoning traces, recent LRM pruning methods improve calibration via on-policy chain-of-thought RAC~\citep{prune-RAC}, reasoning-step importance OBC~\citep{OBC-prune}, or self-generated traces~\citep{beware_calibration,resp,ssgr}, detailed preliminaries in Appendix~\ref{app:preliminaries}.


\paragraph{Related Works.} LRMs can overthink and enter repetitive loops~\citep{chen2026towards,pipis2025wait}, which quantization can further prolong~\citep{lotfi2026quantized}. We show that pruning amplifies looping while largely preserving the stopping signal, distinguishing failures to stop from failures to reason. Activation steering offers a way to modify behavior by adding directions to the residual stream, with difference-in-means directions shown to control behaviors of LLMs ~\citep{arditi2024refusal}. For reasoning, SEAL~\citep{chen2025seal} suppresses reflection, while Plan-CoT~\citep{sheng2025reasoning} modulates reasoning length. In contrast, TGAS targets pruning-induced non-termination using contrasting continuations of the same prefix, yielding a direction weakly aligned with SEAL's (Appendix~\ref{app:seal-comparison}). 
\begin{figure*}[t]
    \centering
    \includegraphics[width=0.95\linewidth]{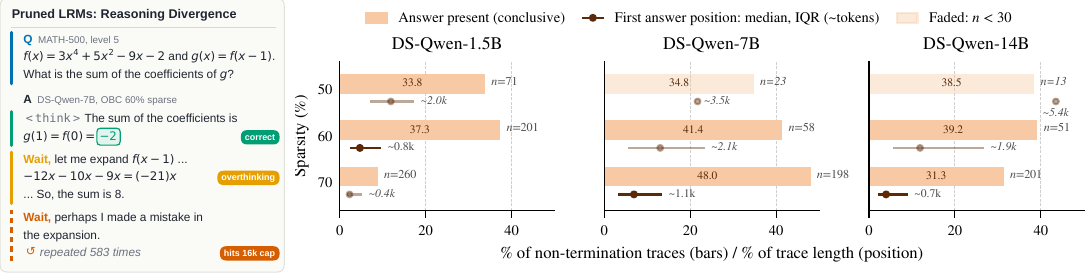}
\caption{Gold answers in non-terminating reasoning traces of DeepSeek models on MATH-500. Left: an example of reasoning divergence, where the pruned model continues reasoning after reaching the correct answer. Right: the fraction of traces containing the gold answer and the median position of its first occurrence. $n$ denotes the number of non-terminating samples.}
    \label{fig:answer_in_think}
\end{figure*}


\section{Understanding Pruning-Induced Reasoning Degradation}
\label{sec:observation}
We investigate three questions: (i) Do pruned models lose reasoning ability or only the ability to stop? (ii) Is non-termination caused by limited generation budgets? (iii) Why do models fail to stop despite reaching the correct answer? We address the first two questions in Section~\ref{sec:obs_1} and the third in Section~\ref{subsec:behavior}. Recall that we refer to this phenomenon as \emph{reasoning divergence}, characterized by \emph{non-termination}, driven by repetitive \emph{looping} and an \emph{failure to commit} to final answers.

\textbf{Setting.} We evaluate DeepSeek-R1-Distill-Qwen (DS-Qwen), DeepSeek-R1-Distill-Llama (DS-Llama)~\citep{deepseek_r1}, and OpenThinker3 (OT3)~\citep{guha2026openthoughts} under one-shot unstructured pruning of decoder linear layers at varying sparsity levels, using SparseGPT or Wanda with OBC, RAC, or C4 calibration. Benchmarks include MATH-500 Levels 4-5 (hereafter MATH-500)~\citep{math-500}, AIME-25~\citep{aime25}, and LiveCodeBench-v6 (LCB)~\citep{livecodebench}; OpenR1-Math-220K (OpenR1-Math)~\citep{openr1} is used only for calibration and steering-direction construction. We use a 16,384-token budget, $T=0.6$, and top-$p=0.95$, averaging $k=4$ responses on three benchmarks. Our default uses SparseGPT with OBC calibration on model-generated reasoning traces, counting a response as correct only if it commits to a final answer, detailed in Appendix~\ref{app:experimental_setup}.

\subsection{Do Pruned Models Lose the Ability to Reason?}
\label{sec:obs_1}

\paragraph{Accuracy on Dense-Correct Problems.} To determine whether reasoning divergence reflects a loss of reasoning capability, we evaluate pruned models on problems correctly solved by their dense counterparts. Table~\ref{tab:dense-correct-main} shows that as sparsity grows, the additional errors come mainly from non-termination rather than from incorrect answers. On MATH-500, going from 50\% to 60\% sparsity raises the non-termination rate from 3.8\% to 16.8\% for DS-Qwen-7B, and from 2.4\% to 16.4\% for the 14B counterpart, while the terminated-but-wrong rate stays nearly unchanged. Terminated generations largely preserve their accuracy at high sparsity, despite fewer dense-correct samples remaining correct. These results suggest that substantial reasoning capability remains intact under moderate pruning, while termination becomes the primary bottleneck as sparsity grows (full results in Appendix~\ref{app:corrected_dense_acc_full}).

\begin{table}[t]
\centering
\scriptsize
\setlength{\tabcolsep}{2.5pt}
\resizebox{0.8\columnwidth}{!}{
\begin{tabular}{ll rrrr}
\toprule
\thd{Model} & \thd{Ckpt} & \thd{Kept} & \thd{No-Term.} & \thd{Wrong} & \thd{Kept$\mid$term} \\
\midrule
\multirow{2}{*}{\makecell{DS-Qwen\\-7B}} & \ckpt{50} & 91.7\sd{0.8} & 3.8\sd{0.3} & 4.5\sd{0.6} & 95.3\sd{0.6} \\
 & \ckpt{60} & 77.8\sd{0.7} & 16.8\sd{0.1} & 5.4\sd{0.6} & 93.5\sd{0.7} \\
\midrule
\multirow{2}{*}{\makecell{DS-Qwen\\-14B}} & \ckpt{50} & 89.2\sd{0.6} & 2.4\sd{1.5} & 8.4\sd{0.9} & 91.4\sd{0.8} \\
 & \ckpt{60} & 74.3\sd{2.8} & 16.4\sd{1.9} & 9.3\sd{0.9} & 88.9\sd{1.4} \\
\midrule
\multirow{2}{*}{OT3-7B} & \ckpt{50} & 79.3\sd{2.9} & 13.1\sd{2.4} & 7.6\sd{1.9} & 91.3\sd{2.2} \\
 & \ckpt{60} & 63.6\sd{3.4} & 29.8\sd{3.3} & 6.6\sd{1.8} & 90.6\sd{2.5} \\
\bottomrule
\end{tabular}}
\caption{Accuracy on MATH-500 problems correctly solved by the dense model. \thd{Kept}: correct; \thd{No-Term.}: reasoning fails to terminate; \thd{Wrong}: terminated but incorrect; \thd{Kept$\mid$term}: accuracy among terminated generations. From 50\% to 60\% sparsity, non-termination grows sharply while the terminated-but-wrong rate stays nearly flat, so accuracy among terminated generations remains high.}
\label{tab:dense-correct-main}
\end{table}

\begin{figure}[t]
    \centering
    \includegraphics[width=0.85\columnwidth]{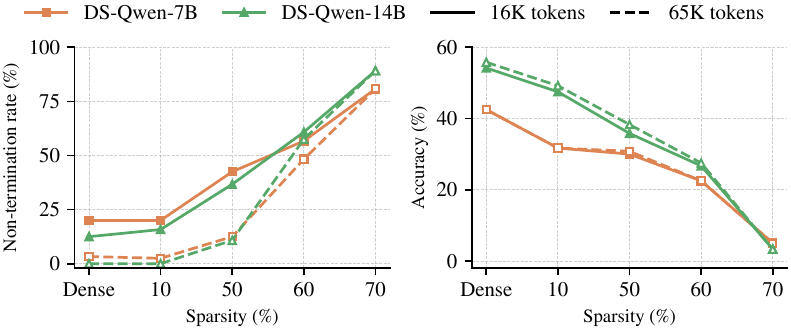}
\caption{Non-termination rate and accuracy of DeepSeek models on AIME-25 under 16K- and 65K-token budgets.}
    \label{fig:budget}
\end{figure}

\paragraph{Larger Generation Budget.} We ask whether reasoning divergence is simply caused by an insufficient generation budget. As shown in Figure~\ref{fig:budget}, increasing the budget from 16,384 to 65,536 tokens substantially reduces non-termination at low and moderate sparsity, e.g., from 42.5\% to 12.5\% for 7B at 50\% sparsity, but leaves accuracy nearly unchanged for DS-Qwen 7B and 14B, while at 60--70\% sparsity non-termination barely decreases. Appendix~\ref{app:larger_budget} shows that the larger budget recovers only 0--3 additional correct traces across settings, suggesting that pruning-induced failures are not primarily due to insufficient generation length.

\paragraph{Correct Answer Presence in Reasoning Traces.} Since a larger budget rarely helps, we next ask what the non-terminating traces contain: does the reasoning already reach the correct answer even when the model fails to terminate and finalize it? To answer this question, we use text-based rules, termed \emph{Conclusive}, that detect answer-like statements following cues such as \emph{"thus the answer"} or \emph{"so we get"}(Appendix~\ref{app:presence}).\footnote{We validate \emph{Conclusive} against an independent LLM judge: on 250 sampled traces, the two agree on 96.0\% of cases ($\kappa=0.898$; Appendix~\ref{app:judge}).} 
Figure~\ref{fig:answer_in_think} shows that a substantial fraction of non-terminating traces already contain the correct answer: at 60\% sparsity, this holds for $41.4\%$ and $39.2\%$ of non-terminating traces for DS-Qwen 7B and 14B, respectively, with the answer appearing early at a median of $13.0\%$ and $11.9\%$ of the trace length, roughly 2.1k and 1.9k tokens. Despite reaching the correct answer, models continue generating to the token limit, often entering repetitive or corrective reasoning. This suggests that non-termination often reflects reasoning divergence rather than an inability to reach the correct answer. We provide a headroom analysis in Appendix~\ref{app:headroom}, quantifying the recoverable accuracy loss by counting non-terminating traces that already contain the correct answer as correct.

\subsection{Why Do Pruned Models Fail to Terminate?}
\label{subsec:behavior}
Section~\ref{sec:obs_1} shows that pruned LRMs largely retain the ability to reach correct answers but increasingly fail to terminate. We show that the stopping signal remains intact, yet during free-running generation, models struggle to commit to final answers and fall into repetitive reasoning loops.


\paragraph{Does pruning damage the stopping mechanism?}
We test whether pruning impairs stopping recognition via teacher forcing. Let $\mathcal{D}_{\mathrm{cov}}$ denote dense traces that terminate correctly. We replay these traces through each pruned checkpoint and measure $p(\ethink{})$ and its rank at the dense model's termination position (\textbf{STOP}). As shown in Table~\ref{tab:exp3}, at 50--60\% sparsity, the signal remains strong: for 7B, $p(\ethink{})>0.92$ with median rank 0 on both AIME-25 and MATH-500, while 14B exceeds $0.85$ with the same rank. At 70\%, it drops to $0.036$--$0.040$ for 7B and $0.073$--$0.130$ for 14B. Comparing STOP with positions more than 50 tokens earlier, we find that $\ethink{}$ remains substantially more likely at STOP at all sparsities, but at 70\% its log-probability falls below that of the ordinarily generated tokens, detailed in Appendix~\ref{app:stopping_mechanism}. Thus, the stopping signal remains largely intact under moderate pruning, suggesting that the main failure is reaching, rather than recognizing, the stopping point. At 70\% sparsity, the signal remains localized but weakens, consistent with severe pruning also impairing stopping recognition.


\begin{table}[t]
\centering
\scriptsize
\setlength{\tabcolsep}{4pt}
\resizebox{\columnwidth}{!}{
\begin{tabular}{ll rrr rrr}
\toprule
 & & \multicolumn{3}{c}{\thd{DS-Qwen-7B}} & \multicolumn{3}{c}{\thd{DS-Qwen-14B}} \\
\cmidrule(lr){3-5}\cmidrule(lr){6-8}
 & & \ckpt{50} & \ckpt{60} & \ckpt{70} & \ckpt{50} & \ckpt{60} & \ckpt{70} \\
\midrule
\multirow{2}{*}{AIME-25} & $p(\ethink{})$ & 0.998 & 0.957 & 0.036 & 0.995 & 0.854 & 0.073 \\
 & Med.\ rank & 0 & 0 & 5.5 & 0 & 0 & 2 \\
\midrule
\multirow{2}{*}{MATH-500} & $p(\ethink{})$ & 0.996 & 0.924 & 0.040 & 0.998 & 0.907 & 0.130 \\
 & Med.\ rank & 0 & 0 & 7 & 0 & 0 & 3 \\
\bottomrule
\end{tabular}}
\caption{Teacher-forced stopping signal at the dense model's STOP position: probability of \ethink{} and its median rank (0 means \ethink{} is the top prediction).}
\label{tab:exp3}
\end{table}

\begin{table}[t]
\centering
\scriptsize
\setlength{\tabcolsep}{4pt}
\resizebox{\columnwidth}{!}{
\begin{tabular}{l rrrr rrrr}
\toprule
 & \multicolumn{4}{c}{\thd{DS-Qwen-7B}} & \multicolumn{4}{c}{\thd{OT3-7B}} \\
\cmidrule(lr){2-5}\cmidrule(lr){6-9}
 & \textcolor{gray}{\dns} & \ckpt{50} & \ckpt{60} & \ckpt{70} & \textcolor{gray}{\dns} & \ckpt{50} & \ckpt{60} & \ckpt{70} \\
\midrule
$n$ & \textcolor{gray}{794} & 788 & 796 & 782 & \textcolor{gray}{800} & 800 & 796 & 790 \\
$\Delta_{\mathrm{cyc}}$ & \textcolor{gray}{$-15.3$} & $-11.9$ & $-9.6$ & $-5.0$ & \textcolor{gray}{$-23.1$} & $-15.8$ & $-13.2$ & $-5.6$ \\
Entropy & \textcolor{gray}{1.55} & 1.51 & 1.41 & 1.44 & \textcolor{gray}{1.03} & 1.22 & 1.33 & 1.60 \\
\bottomrule
\end{tabular}}
\caption{Log-probability ratio between cyclic and progressive continuations at decision points on AIME-25. $n$: number of decision points; \thd{Entropy}: predictive entropy at the decision points. Less negative (larger) $\Delta_{\mathrm{cyc}}$ values indicate a stronger preference for the repetitive continuation.}
\label{tab:exp4-logprob}
\end{table}

\begin{figure}[t]
\centering
\includegraphics[width=0.8\linewidth]{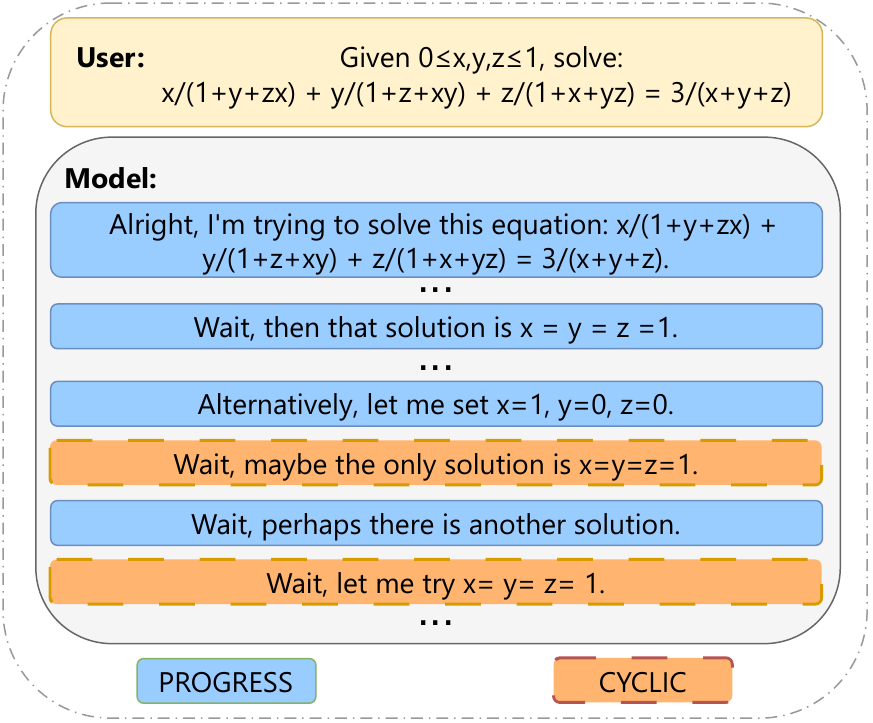}
\caption{\textsc{Progress} and \textsc{Cyclic} behaviors illustration.}
\label{fig:loop_example}
\end{figure}

\paragraph{Definition.} We investigate whether pruning alters reasoning trajectories at moments of reconsideration, where a model may either move forward or revisit previous reasoning. We define a \emph{decision point} as the start of a paragraph beginning with a predefined reconsideration cue (e.g., \emph{"wait"}, \emph{"hmm"}, full list in Appendix~\ref{app:full_transition_pruned}), following \citet{lotfi2026quantized}. We label each paragraph \textsc{Cyclic} if its 8-word n-gram set has Jaccard similarity $\geq 0.5$ with any earlier paragraph in the same trace, and \textsc{Progress} otherwise, following \citet{pipis2025wait}, as illustrated in Figure~\ref{fig:loop_example}. 

\paragraph{Teacher-Forced Preferences.} To measure whether a model prefers to repeat or to move forward, we compare two candidate next paragraphs at each decision point of a dense trace: repeating the preceding paragraph (\textsc{Cyclic}) or continuing with the next paragraph that adds new content (\textsc{Progress}). We score the first 32 tokens of each candidate under teacher forcing with the same prefix, and reuse the same decision points and candidates across all pruning checkpoints. We restrict evaluation to decision points that are followed by a \textsc{Progress} paragraph, and define:
\begin{equation*}
\begin{split}
\Delta_{\mathrm{cyc}} = \sum_{t=1}^{32} \Big[ &\log p_{\theta}\big(\mathbf{x}_t^{\text{cyclic}}\mid\text{prefix},\mathbf{x}_{<t}^{\text{cyclic}}\big) \\
&- \log p_{\theta}\big(\mathbf{x}_t^{\text{progress}}\mid\text{prefix},\mathbf{x}_{<t}^{\text{progress}}\big)\Big]
\end{split}
\label{eq:delta-cyc}
\end{equation*}
where a larger $\Delta_{\mathrm{cyc}}$ indicates a stronger relative preference for \textsc{Cyclic} over \textsc{Progress}. 

As shown in Table~\ref{tab:exp4-logprob}, pruning progressively shifts the model's preference from progressive toward cyclic continuations at decision points; Appendix~\ref{app:full_transition_pruned} shows a similar shift at randomly sampled paragraph, indicating a general preference for repetition rather than one specific to reconsideration cues. For DS-Qwen-7B, $\Delta_{\mathrm{cyc}}$ increases from $-15.3$ for the dense model to $-5.0$ at 70\% sparsity; OT3-7B shows a similar shift from $-23.1$ to $-5.6$, other models in Appendix~\ref{app:full_transition_pruned}. Predictive entropy changes little for DS-Qwen-7B but increases for OT3-7B, whereas $\Delta_{\mathrm{cyc}}$ shifts consistently for both, suggesting that the shift is not merely a by-product of increased uncertainty. Overall, pruning makes models increasingly favor repetitive over progressive reasoning, which we confirm in free-running generation below.


\begin{table}[t]
\centering
\scriptsize
\setlength{\tabcolsep}{2.2pt}
\resizebox{\columnwidth}{!}{
\begin{tabular}{l rrrr rrrr}
\toprule
 & \multicolumn{4}{c}{DS-Qwen-7B} & \multicolumn{4}{c}{DS-Qwen-14B} \\
\cmidrule(lr){2-5}\cmidrule(lr){6-9}
 & \textcolor{gray}{\dns} & \ckpt{50} & \ckpt{60} & \ckpt{70}
 & \textcolor{gray}{\dns} & \ckpt{50} & \ckpt{60} & \ckpt{70} \\
\midrule
$P$(cyc $\mid$ cyc)  & \textcolor{gray}{0.029} & 0.283 & 0.327 & 0.307 & \textcolor{gray}{0.000} & 0.203 & 0.271 & 0.324 \\
$P$(cyc $\mid$ prog) & \textcolor{gray}{0.026} & 0.043 & 0.064 & 0.085 & \textcolor{gray}{0.029} & 0.033 & 0.059 & 0.081 \\
Count (k)            & \textcolor{gray}{1.3}   & 51.5  & 85.7  & 113.4 & \textcolor{gray}{0.8}   & 47.3  & 68.3  & 136.8 \\
\bottomrule
\end{tabular}}
\caption{Transition dynamics between \textsc{Cyclic} and \textsc{Progress} paragraphs on AIME-25 from shared branch points. $P(\mathrm{cyc}\mid\mathrm{cyc})$ measures loop persistence, while $P(\mathrm{cyc}\mid\mathrm{prog})$ measures loop entry; \thd{Count}: number of adjacent paragraph pairs, in thousands; dense rows have fewer pairs due to early termination, while pruned checkpoints retain reasoning divergence despite using dense correct traces.}
\label{tab:exp5-branch-transitions-aime25-main}
\end{table}


\paragraph{Free-Running Continuations.} The preceding analysis shows that pruning shifts model behavior toward repetitive loops rather than progressing new reasoning steps. We next examine how pruned models continue dense reasoning traces. Following Section~\ref{sec:obs_1}, we revisit the \emph{Conclusive} point. For each trace, we define the branch point $b$ as the first eligible paragraph start after \emph{Conclusive} such that the prefix $\mathbf{x}_{<b}$ is sufficient to identify the correct answer. We teacher-force the prefix through the pruned checkpoint and sample $N$ continuations, discarding traces without eligible branch points. Thus, continuations share the same prefix, checkpoint, and problem, differing only beyond the branch point.

Table~\ref{tab:exp5-branch-transitions-aime25-main} reveals a shift toward cyclic reasoning in pruned checkpoints. Both loop persistence, $P(\mathrm{cyc}\mid\mathrm{cyc})$, and loop entry, $P(\mathrm{cyc}\mid\mathrm{prog})$, generally increase with pruning. For DS-Qwen 7B and 14B at 70\% sparsity, $P(\mathrm{cyc}\mid\mathrm{cyc})$ reaches $0.307$ and $0.324$, while $P(\mathrm{cyc}\mid\mathrm{prog})$ reaches $0.085$ and $0.081$, respectively, compared with at most $0.029$ for the dense models (other models in Appendix~\ref{app:full_transition_pruned}). This indicates that pruning is associated with both more persistent cyclic reasoning and more frequent transitions from progressive to cyclic states.

\section{Methodology}
\label{sec:method}

If looping reflects a failure to commit rather than a loss of capability, a pruned model should be steerable toward committing without weight updates. We use a hidden-state direction that separates terminating from looping continuations under the same prefix.

\subsection{Constructing the Termination Direction}
\label{sec:method_direction}

We reuse the dense correct traces $\mathcal{D}_{\mathrm{cov}}$ and branch points $b$ from Section~\ref{subsec:behavior}; at each branch point, the prefix $\mathbf{x}_{<b}$ already suffices to derive the correct answer. From each prefix, we sample $N$ continuations with the pruned checkpoint and split them into $\mathcal{C}^{+}_{b}$, continuations that emit \ethink{} within the token budget, and $\mathcal{C}^{-}_{b}$, continuations that reach the budget without terminating ($N$, set sizes and median lengths are reported in Appendix~\ref{app:construct_steer_vector}). For each layer $\ell \in \{1,\ldots,L\}$, we represent a continuation $c$ by its mean hidden state over the span from the branch point $b$ to the dense trace's \ethink{} position $t_{\mathrm{end}}$:
\begin{equation*}
\mathbf{s}_\ell(c) = \frac{1}{t_{\mathrm{end}}-b}\sum_{t=b}^{t_{\mathrm{end}}-1}\mathbf{h}_{\ell,t}(c),
\end{equation*}
For continuations that end before $t_{\mathrm{end}}$, we average hidden states from $b$ to the continuation's end. We use the dense trace's $t_{\mathrm{end}}$ as a common reference to exclude the looping tail from the mean and ensure comparable representations. We compute the class means as:
\begin{equation*}
\boldsymbol{\mu}^{+}_{\ell,b} = \frac{1}{|\mathcal{C}^{+}_{b}|}\sum_{c \in \mathcal{C}^{+}_{b}} \mathbf{s}_\ell(c),
\quad
\boldsymbol{\mu}^{-}_{\ell,b} = \frac{1}{|\mathcal{C}^{-}_{b}|}\sum_{c \in \mathcal{C}^{-}_{b}} \mathbf{s}_\ell(c).
\end{equation*}
Following the difference-in-means approach of \citet{arditi2024refusal}, we then average the class difference over branch points:
\begin{equation}
\mathbf{r}_\ell = \frac{1}{|\mathcal{B}|} \sum_{b \in \mathcal{B}} \bigl( \boldsymbol{\mu}^{+}_{\ell,b} - \boldsymbol{\mu}^{-}_{\ell,b} \bigr),
\label{eq:classmeans}
\end{equation}
where $\mathcal{B}$ denotes the set of eligible branch points collected from $\mathcal{D}_{\mathrm{cov}}$. Because the difference is taken within each branch point, the two classes share the same problem and prefix, so problem-specific content cancels and $\mathbf{r}_\ell$ isolates the contrast between committing and looping. This yields a candidate direction $\mathbf{r}_\ell$ for each layer. Since layers lie in different residual spaces, we compare two checkpoints $A$ and $B$ by the layer-averaged cosine similarity $\mathrm{sim}(A,B)=\frac{1}{L}\sum_{\ell=1}^{L}\cos(\mathbf{r}^{A}_{\ell},\mathbf{r}^{B}_{\ell})$ (details in Appendix~\ref{app:commitment-direction-analysis}). For the analysis in this section we use MATH-500 traces; the directions used in all experiments (Section~\ref{sec:experiment}) are constructed from OpenR1-Math traces disjoint from every evaluation set. 

\begin{figure}[t]
    \centering
    \includegraphics[width=0.9\columnwidth]{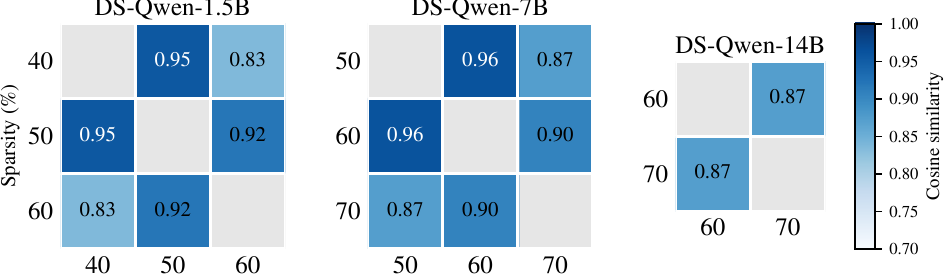}
    \caption{Cosine similarity between termination-guided directions constructed from pruned DeepSeek models on MATH-500. We compare only checkpoints with sufficient positive-terminated and negative non-terminating samples; for example, all 1.5B checkpoints have at least 50 samples in each class. Directions constructed from different pruned checkpoints exhibit high cosine similarity.}
    \label{fig:consine_directions}
\end{figure}

\subsection{Direction Consistency Across Checkpoints}
\label{sec:method_consistency}
Figure~\ref{fig:consine_directions} reports $\mathrm{sim}(A,B)$ between directions from different pruning levels of the same model, and shows that they are highly consistent. For DS-Qwen-7B on MATH-500, the cosine similarity between directions from 50--70\% pruning ratios remains at least $0.87$; for DS-Qwen-1.5B, directions from 40--60\% have pairwise cosine similarities of at least $0.83$ (sample counts and results on OpenR1-Math are in Appendix~\ref{app:construct_steer_vector}). These results suggest that the difference between repetitive reasoning and commitment is represented in a relatively stable direction in hidden-state space rather than being specific to an individual pruning checkpoint, enabling a single direction per model to generalize across sparsity levels. Appendix~\ref{app:commitment-direction-analysis} shows that this similarity is not an artifact of comparing checkpoints of the same model: across 15 checkpoint pairs, it exceeds both a label-shuffled baseline and a shared contrast unrelated to commitment in every pair.

\subsection{Steering at Inference}
\label{sec:method_steer}
TGAS constructs the direction at an intermediate sparsity level, where both $\mathcal{C}^{+}_{b}$ and $\mathcal{C}^{-}_{b}$ contain enough trajectories, and reuses it for all sparsity levels of the same model. We select the steering layer $\ell^\star$ on held-out branch points as the layer whose direction best separates the two classes (highest ROC AUC of the projection; Appendix~\ref{app:construct_steer_vector}). The steering vector is this direction rescaled to the typical activation norm at that layer,
$\mathbf{v}=\bar{n}_{\ell^\star}\,\mathbf{r}_{\ell^\star}/\|\mathbf{r}_{\ell^\star}\|_2$, where $\bar{n}_{\ell^\star}$ is the mean $\ell_2$ norm of residual-stream representations over calibration trajectories. At layer $\ell^\star$, we add it to the residual stream at every token position $t$ during autoregressive generation:
$$\tilde{\mathbf{h}}_{\ell^\star,t}
\leftarrow
\mathbf{h}_{\ell^\star,t}
+\alpha\,\mathbf{v},$$
where the coefficient $\alpha$ scales the intervention relative to the typical activation norm; positive values steer toward commitment, while negative values steer in the opposite direction. We select $\alpha$ per checkpoint by accuracy on a held-out OpenR1-Math split, detailed in Table~\ref{tab:hparams}.


\section{Experiments}
\label{sec:experiment}

\begin{table*}[t]
\centering
\scriptsize
\setlength{\tabcolsep}{3pt}
\renewcommand{\arraystretch}{1}
\resizebox{0.7\textwidth}{!}{%
\begin{tabular}{lll rr rr rr r}
\toprule
& & & \multicolumn{2}{c}{\thd{MATH-500}} & \multicolumn{2}{c}{\thd{AIME-25}} & \multicolumn{2}{c}{\thd{LiveCodeBench}} & \multicolumn{1}{c}{\thd{Avg}} \\
\cmidrule(lr){4-5}\cmidrule(lr){6-7}\cmidrule(lr){8-9}\cmidrule(lr){10-10}
\thd{Model} & \thd{Ckpt} & \thd{Method} & \thd{Acc}$\uparrow$ & \thd{Term}$\uparrow$ & \thd{Acc}$\uparrow$ & \thd{Term}$\uparrow$ & \thd{Acc}$\uparrow$ & \thd{Term}$\uparrow$ & \thd{Acc}$\uparrow$ \\
\midrule
\multirow{11}{*}{\makecell{OT3-1.5B}}
& \dns & Base & \textcolor{gray}{80.2\sd{1.6}} & \textcolor{gray}{87.2\sd{1.9}} & \textcolor{gray}{32.5\sd{1.7}} & \textcolor{gray}{39.2\sd{3.2}} & \textcolor{gray}{43.2\sd{2.9}} & \textcolor{gray}{57.4\sd{3.3}} & \textcolor{gray}{52.0} \\
\cmidrule(lr){2-10}
& \multirow{5}{*}{\ckpt{10}} & Base & \underline{79.6}\sd{1.3} & 87.0\sd{0.5} & 34.2\sd{8.3} & 40.0\sd{2.7} & \underline{43.4}\sd{2.3} & 57.1\sd{2.6} & 52.4 \\
& & No-Repeat & 79.2\sd{1.9} & 86.6\sd{2.2} & 35.0\sd{6.9} & 41.7\sd{6.9} & 38.9\sd{2.6} & 55.5\sd{1.6} & 51.0 \\
& & Forced-Answer & 77.3\sd{3.0} & --- & 29.2\sd{6.3} & --- & 40.8\sd{2.6} & --- & 49.1 \\
& & SEAL & 77.1\sd{1.9} & 87.0\sd{2.0} & \textbf{41.7}\sd{5.8} & 52.5\sd{5.7} & 38.7\sd{2.0} & 61.3\sd{5.8} & \underline{52.5} \\
& & \cellcolor{tgasrow}\textbf{TGAS} & \cellcolor{tgasrow}\textbf{81.7}\sd{1.9} & \cellcolor{tgasrow}88.5\sd{0.8} & \cellcolor{tgasrow}\underline{37.5}\sd{3.2} & \cellcolor{tgasrow}46.7\sd{9.8} & \cellcolor{tgasrow}\textbf{45.5}\sd{1.6} & \cellcolor{tgasrow}61.3\sd{2.8} & \cellcolor{tgasrow}\textbf{54.9}\gain{2.4} \\
\cmidrule(lr){2-10}
& \multirow{5}{*}{\ckpt{50}} & Base & 56.9\sd{3.8} & 67.4\sd{4.0} & \textbf{20.8}\sd{1.7} & 32.5\sd{5.0} & 10.0\sd{0.6} & 22.4\sd{1.3} & \underline{29.2} \\
& & No-Repeat & 54.4\sd{0.3} & 67.4\sd{1.3} & 16.7\sd{4.7} & 25.8\sd{3.2} & 6.3\sd{0.9} & 23.2\sd{4.5} & 25.8 \\
& & Forced-Answer & \underline{57.1}\sd{3.5} & --- & 17.5\sd{5.0} & --- & \underline{11.1}\sd{1.8} & --- & 28.6 \\
& & SEAL & 55.0\sd{5.1} & 73.7\sd{2.1} & 18.3\sd{4.3} & 34.2\sd{1.7} & 9.2\sd{1.0} & 26.8\sd{2.8} & 27.5 \\
& & \cellcolor{tgasrow}\textbf{TGAS} & \cellcolor{tgasrow}\textbf{58.4}\sd{2.2} & \cellcolor{tgasrow}74.4\sd{0.3} & \cellcolor{tgasrow}\underline{19.2}\sd{7.4} & \cellcolor{tgasrow}44.2\sd{5.7} & \cellcolor{tgasrow}\textbf{13.9}\sd{2.3} & \cellcolor{tgasrow}29.2\sd{5.6} & \cellcolor{tgasrow}\textbf{30.5}\gain{1.3} \\
\midrule
\multirow{11}{*}{\makecell{DS-Llama\\-8B}}
& \dns & Base & \textcolor{gray}{82.4\sd{2.7}} & \textcolor{gray}{97.9\sd{0.8}} & \textcolor{gray}{34.2\sd{5.7}} & \textcolor{gray}{70.0\sd{11.2}} & \textcolor{gray}{58.7\sd{2.5}} & \textcolor{gray}{85.5\sd{1.8}} & \textcolor{gray}{58.4} \\
\cmidrule(lr){2-10}
& \multirow{5}{*}{\ckpt{50}} & Base & \underline{65.5}\sd{0.3} & 88.5\sd{2.2} & \underline{21.7}\sd{7.9} & 75.0\sd{8.4} & \underline{42.6}\sd{0.6} & 66.8\sd{1.8} & \underline{43.3} \\
& & No-Repeat & 64.7\sd{0.8} & 90.6\sd{1.3} & 20.8\sd{8.8} & 76.7\sd{2.7} & 33.2\sd{3.0} & 75.3\sd{4.9} & 39.6 \\
& & Forced-Answer & 64.5\sd{0.5} & --- & 16.7\sd{6.1} & --- & 40.8\sd{1.3} & --- & 40.7 \\
& & SEAL & 63.0\sd{0.6} & 89.3\sd{1.1} & \textbf{23.3}\sd{5.4} & 76.7\sd{9.4} & 41.3\sd{4.1} & 75.3\sd{3.9} & 42.5 \\
& & \cellcolor{tgasrow}\textbf{TGAS} & \cellcolor{tgasrow}\textbf{67.6}\sd{1.6} & \cellcolor{tgasrow}88.5\sd{1.1} & \cellcolor{tgasrow}\textbf{23.3}\sd{3.8} & \cellcolor{tgasrow}80.8\sd{4.2} & \cellcolor{tgasrow}\textbf{43.9}\sd{3.0} & \cellcolor{tgasrow}72.1\sd{2.0} & \cellcolor{tgasrow}\textbf{44.9}\gain{1.6} \\
\cmidrule(lr){2-10}
& \multirow{5}{*}{\ckpt{60}} & Base & 30.0\sd{3.0} & 50.6\sd{2.4} & 3.3\sd{2.7} & 29.2\sd{4.2} & \underline{10.8}\sd{1.6} & 26.3\sd{2.3} & 14.7 \\
& & No-Repeat & 33.4\sd{0.8} & 74.4\sd{1.1} & 2.5\sd{3.2} & 50.8\sd{10.7} & 9.5\sd{1.5} & 42.1\sd{3.7} & 15.1 \\
& & Forced-Answer & \underline{35.7}\sd{1.3} & --- & 3.3\sd{2.7} & --- & \textbf{11.1}\sd{1.4} & --- & \underline{16.7} \\
& & SEAL & 32.8\sd{1.6} & 61.8\sd{1.3} & \underline{4.2}\sd{5.0} & 45.0\sd{4.3} & 10.0\sd{3.0} & 32.6\sd{1.9} & 15.7 \\
& & \cellcolor{tgasrow}\textbf{TGAS} & \cellcolor{tgasrow}\textbf{36.6}\sd{1.6} & \cellcolor{tgasrow}65.6\sd{0.5} & \cellcolor{tgasrow}\textbf{7.5}\sd{1.7} & \cellcolor{tgasrow}45.0\sd{4.3} & \cellcolor{tgasrow}\underline{10.8}\sd{1.8} & \cellcolor{tgasrow}33.7\sd{1.5} & \cellcolor{tgasrow}\textbf{18.3}\gain{1.6} \\
\midrule
\multirow{11}{*}{\makecell{DS-Qwen\\-14B}}
& \dns & Base & \textcolor{gray}{88.4\sd{0.3}} & \textcolor{gray}{99.0\sd{0.3}} & \textcolor{gray}{54.2\sd{5.0}} & \textcolor{gray}{87.5\sd{5.0}} & \textcolor{gray}{73.2\sd{3.9}} & \textcolor{gray}{90.8\sd{2.2}} & \textcolor{gray}{71.9} \\
\cmidrule(lr){2-10}
& \multirow{5}{*}{\ckpt{50}} & Base & 82.4\sd{0.5} & 95.4\sd{0.5} & 35.8\sd{5.0} & 63.3\sd{6.1} & \underline{62.4}\sd{1.0} & 75.8\sd{2.3} & 60.2 \\
& & No-Repeat & 83.6\sd{1.1} & 95.0\sd{1.6} & \underline{37.5}\sd{3.2} & 64.2\sd{6.3} & 46.6\sd{2.9} & 80.3\sd{3.6} & 55.9 \\
& & Forced-Answer & 81.1\sd{1.9} & --- & 36.7\sd{4.7} & --- & 56.6\sd{3.3} & --- & 58.1 \\
& & SEAL & \underline{84.4}\sd{0.8} & 98.1\sd{0.5} & \textbf{42.5}\sd{5.7} & 70.0\sd{3.8} & 61.3\sd{2.8} & 78.7\sd{2.9} & \underline{62.7} \\
& & \cellcolor{tgasrow}\textbf{TGAS} & \cellcolor{tgasrow}\textbf{86.3}\sd{0.5} & \cellcolor{tgasrow}95.4\sd{0.3} & \cellcolor{tgasrow}\textbf{42.5}\sd{3.2} & \cellcolor{tgasrow}70.8\sd{1.7} & \cellcolor{tgasrow}\textbf{62.9}\sd{3.5} & \cellcolor{tgasrow}77.1\sd{3.5} & \cellcolor{tgasrow}\textbf{63.9}\gain{1.2} \\
\cmidrule(lr){2-10}
& \multirow{5}{*}{\ckpt{60}} & Base & 67.6\sd{1.6} & 80.2\sd{0.5} & \underline{26.7}\sd{4.7} & 39.2\sd{5.0} & 37.6\sd{5.6} & 46.6\sd{3.9} & 44.0 \\
& & No-Repeat & \underline{71.6}\sd{1.3} & 91.4\sd{0.8} & 25.0\sd{3.3} & 70.0\sd{10.5} & 33.9\sd{1.0} & 60.3\sd{3.0} & 43.5 \\
& & Forced-Answer & 70.8\sd{0.8} & --- & \textbf{27.5}\sd{3.2} & --- & \underline{38.9}\sd{4.3} & --- & 45.7 \\
& & SEAL & 71.0\sd{0.8} & 86.6\sd{1.8} & 24.2\sd{4.2} & 45.0\sd{6.4} & \textbf{42.9}\sd{2.9} & 60.0\sd{0.9} & \underline{46.0} \\
& & \cellcolor{tgasrow}\textbf{TGAS} & \cellcolor{tgasrow}\textbf{72.5}\sd{1.3} & \cellcolor{tgasrow}88.9\sd{2.2} & \cellcolor{tgasrow}25.8\sd{3.2} & \cellcolor{tgasrow}44.2\sd{4.2} & \cellcolor{tgasrow}\textbf{42.9}\sd{2.6} & \cellcolor{tgasrow}53.7\sd{3.1} & \cellcolor{tgasrow}\textbf{47.1}\gain{1.1} \\
\bottomrule
\end{tabular}}
\caption{Training-free remedies for SparseGPT-pruned LRMs with OBC calibration. \thd{Acc} is accuracy (\%) and \thd{Term} is the percentage of generations that terminate before the 16,384 token cap. For each checkpoint, the best accuracy is in \textbf{bold} and the second best is \underline{underlined}. TGAS rows are shaded; the \textcolor{table_green}{$\uparrow$} subscript in Avg denotes its gain over the best competitor.}
\label{tab:main-results}
\end{table*}

\begin{table*}[t]
\centering
\scriptsize
\setlength{\tabcolsep}{4pt}
\begin{minipage}[t]{0.48\textwidth}
\vspace{0pt}\centering
\resizebox{0.92\linewidth}{!}{%
\begin{tabular}{ll rr rr}
\toprule
\multicolumn{6}{c}{\textbf{(a) RAC (SparseGPT)}} \\
\midrule
& & \multicolumn{2}{c}{\thd{AIME-25}} & \multicolumn{2}{c}{\thd{LiveCodeBench}} \\
\cmidrule(lr){3-4}\cmidrule(lr){5-6}
\thd{Ckpt} & \thd{Method} & \thd{Acc} & \thd{Term} & \thd{Acc} & \thd{Term} \\
\midrule
\multirow{4}{*}{\texttt{rac\_10}} & Base & 35.8\sd{5.0} & 79.2\sd{4.2} & \underline{56.3}\sd{1.8} & 85.0\sd{3.8} \\
 & No-Repeat & \underline{38.3}\sd{5.8} & 80.0\sd{5.4} & 47.4\sd{3.1} & 85.8\sd{1.1} \\
 & Forced-Answer & 33.3\sd{4.7} & --- & 53.7\sd{1.2} & --- \\
 & \cellcolor{tgasrow}\textbf{TGAS} & \cellcolor{tgasrow}\textbf{39.2}\sd{3.2} & \cellcolor{tgasrow}85.0\sd{6.9} & \cellcolor{tgasrow}\textbf{58.7}\sd{2.6} & \cellcolor{tgasrow}89.7\sd{3.0} \\
\cmidrule(lr){1-6}
\multirow{4}{*}{\texttt{rac\_50}} & Base & 24.2\sd{5.7} & 32.5\sd{5.7} & 32.6\sd{1.7} & 37.4\sd{2.2} \\
 & No-Repeat & \underline{25.0}\sd{3.3} & 79.2\sd{7.9} & 27.9\sd{2.8} & 80.0\sd{3.6} \\
 & Forced-Answer & 23.3\sd{3.8} & --- & \textbf{37.1}\sd{1.8} & --- \\
 & \cellcolor{tgasrow}\textbf{TGAS} & \cellcolor{tgasrow}\textbf{25.8}\sd{1.7} & \cellcolor{tgasrow}59.2\sd{7.4} & \cellcolor{tgasrow}\underline{34.5}\sd{4.0} & \cellcolor{tgasrow}47.9\sd{1.1} \\
\bottomrule
\end{tabular}}
\end{minipage}\hfill
\begin{minipage}[t]{0.48\textwidth}
\vspace{0pt}\centering
\resizebox{0.92\linewidth}{!}{%
\begin{tabular}{ll rr rr}
\toprule
\multicolumn{6}{c}{\textbf{(b) C4 calibration (Wanda)}} \\
\midrule
& & \multicolumn{2}{c}{\thd{AIME-25}} & \multicolumn{2}{c}{\thd{LiveCodeBench}} \\
\cmidrule(lr){3-4}\cmidrule(lr){5-6}
\thd{Ckpt} & \thd{Method} & \thd{Acc} & \thd{Term} & \thd{Acc} & \thd{Term} \\
\midrule
\multirow{4}{*}{\texttt{c4\_40}} & Base & \underline{34.2}\sd{3.2} & 67.5\sd{6.9} & 49.5\sd{1.7} & 73.7\sd{3.5} \\
 & No-Repeat & 31.7\sd{4.3} & 75.8\sd{3.2} & 41.6\sd{4.6} & 77.9\sd{1.2} \\
 & Forced-Answer & 30.0\sd{4.7} & --- & \underline{49.7}\sd{2.0} & --- \\
 & \cellcolor{tgasrow}\textbf{TGAS} & \cellcolor{tgasrow}\textbf{35.0}\sd{1.9} & \cellcolor{tgasrow}75.8\sd{5.0} & \cellcolor{tgasrow}\textbf{52.1}\sd{0.6} & \cellcolor{tgasrow}81.6\sd{4.7} \\
\cmidrule(lr){1-6}
\multirow{4}{*}{\texttt{c4\_50}} & Base & \underline{19.2}\sd{1.7} & 47.5\sd{9.2} & 31.8\sd{5.3} & 52.9\sd{5.6} \\
 & No-Repeat & 15.8\sd{3.2} & 61.7\sd{1.9} & 26.6\sd{4.7} & 67.4\sd{2.4} \\
 & Forced-Answer & \underline{19.2}\sd{1.7} & --- & \underline{34.7}\sd{4.3} & --- \\
 & \cellcolor{tgasrow}\textbf{TGAS} & \cellcolor{tgasrow}\textbf{21.7}\sd{4.3} & \cellcolor{tgasrow}64.2\sd{7.4} & \cellcolor{tgasrow}\textbf{39.2}\sd{4.0} & \cellcolor{tgasrow}58.7\sd{3.6} \\
\bottomrule
\end{tabular}}
\end{minipage}
\caption{Training-free remedies for checkpoints pruned with (a) RAC and (b) C4-calibrated Wanda on DS-Qwen-7B. \thd{Acc} is accuracy (\%) and \thd{Term} the share of generations that terminated (\%).}
\label{tab:ablation-pruners-main}
\end{table*}

\begin{table}[t]
\centering
\scriptsize
\setlength{\tabcolsep}{4pt}
\resizebox{0.8\columnwidth}{!}{%
\begin{tabular}{ll rr rr}
\toprule
& & \multicolumn{2}{c}{\thd{$\Delta_{\mathrm{cyc}}$ at decisions}} & \multicolumn{2}{c}{\thd{DPs/1k}} \\
\cmidrule(lr){3-4}\cmidrule(lr){5-6}
\thd{Model} & \thd{Ckpt} & \thd{U} & \thd{S} & \thd{U} & \thd{S} \\
\midrule
\multirow{2}{*}{\makecell{DS-Qwen\\-7B}} & \ckpt{50} & $-15.3$\sd{0.5} & $-17.1$\sd{0.5} & 9.0 & 5.4 \\
 & \ckpt{60} & $-12.0$\sd{0.4} & $-15.4$\sd{0.5} & 12.3 & 5.8 \\
\midrule
\multirow{2}{*}{\makecell{DS-Qwen\\-14B}} & \ckpt{50} & $-14.7$\sd{0.4} & $-17.9$\sd{0.5} & 7.2 & 5.5 \\
 & \ckpt{60} & $-12.7$\sd{0.4} & $-14.0$\sd{0.4} & 11.5 & 6.8 \\
\bottomrule
\end{tabular}}
\caption{$\Delta_{\mathrm{cyc}}$ at decision points, defined in Section \ref{subsec:behavior}, on MATH-500, unsteered (\thd{U}) vs.\ steered with TGAS (\thd{S}). \thd{DPs/1k}: decision points per 1{,}000 tokens.}
\label{tab:steer-deltacyc-main}
\end{table}

\begin{figure*}[t]
\centering
\begin{minipage}[t]{0.6\textwidth}
\vspace{0pt}\centering
\includegraphics[width=0.95\linewidth,height=92pt]{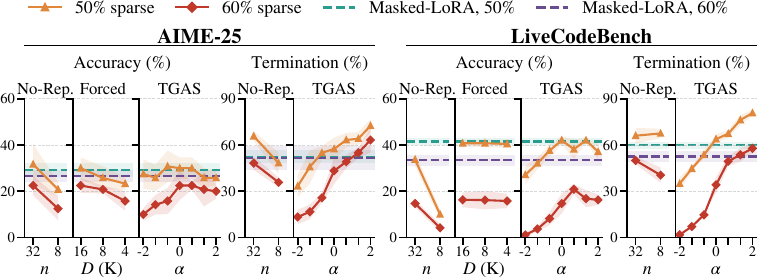}
\caption{Ablation results on DS-Qwen-7B. TGAS provides controllable termination steering via $\alpha$. Masked-LoRA exceeds training-free methods on LCB, but requires separate optimization for each model and sparsity level.}
\label{fig:ablation_7B}
\end{minipage}\hfill
\begin{minipage}[t]{0.38\textwidth}
\vspace{0pt}\centering
\scriptsize
\setlength{\tabcolsep}{3pt}
\resizebox{\linewidth}{!}{%
\begin{tabular}{l rrr r rr rr}
\toprule
 & \multicolumn{3}{c}{\thd{DS-Qwen}} & \thd{DS-Llama} & \multicolumn{2}{c}{\thd{OT3}} & & \\
\cmidrule(lr){2-4}\cmidrule(lr){5-5}\cmidrule(lr){6-7}
\thd{Sparsity} & \thd{1.5B} & \thd{7B} & \thd{14B} & \thd{8B} & \thd{1.5B} & \thd{7B} & \thd{Avg.} & \thd{Saved} \\
\midrule
Dense & 11.0 & 2.8 & 2.7 & 3.3 & 3.6 & 2.7 & 4.4 & 392 \\
10\% & 10.2 & 10.9 & 5.9 & 4.6 & 5.9 & 3.7 & 6.9 & 517 \\
50\% & 17.4 & 8.3 & 1.7 & 2.5 & 7.9 & 3.4 & 6.9 & 680 \\
60\% & 23.8 & 18.5 & 9.4 & 12.9 & 3.6 & 5.7 & \textbf{12.3} & \textbf{1{,}423} \\
\bottomrule
\end{tabular}}
\raggedright
\captionof{table}{Reduction in generated tokens (\%) from TGAS relative to the unsteered model, averaged over MATH-500, AIME-25, and LiveCodeBench. \thd{Avg.}: mean reduction over models (\%); \thd{Saved}: mean tokens saved per generation.}
\label{tab:tokens-main}
\end{minipage}
\end{figure*}

\paragraph{Setting.} We compare TGAS with three training-free post-pruning baselines: Forced-Answer, which stops after $D$ reasoning tokens; No-Repeat, which suppresses repeated $n$-grams; and SEAL, which constructs a steering vector from reflection versus execution steps to reduce redundant reasoning. TGAS constructs one direction per model from OpenR1-Math calibration rollouts of an intermediate-sparsity checkpoint (50\% for 1.5B models and 60\% for other models), and reuses it across all sparsity levels. We also compare with training-based \textit{Masked-LoRA}~\citep{kim2024shortened}, which trains rank-64 LoRA adapters restricted to the surviving weights, details in Appendix~\ref{app:detail_training_baseline}. Hyperparameters are tuned on a held-out OpenR1 split and evaluated on AIME-25, MATH-500, and LCB (Appendix~\ref{app:detail_training_baseline}). Table~\ref{tab:main-results} reports three representative models (1.5B–14B) at the two highest sparsity levels before accuracy collapses on AIME-25 and LCB; results for all six models are provided in Appendix~\ref{app:full_result_comparison}


\paragraph{Does Targeting Termination Recover Accuracy?} Table~\ref{tab:main-results} compares TGAS with training-free baselines. Consistent with our diagnosis, steering toward termination improves both termination and accuracy: averaged over benchmarks, TGAS improves over standard inference at every model-sparsity setting (by 1.3--3.7 points), and it raises termination in 16 of 18 checkpoint-benchmark pairs without lowering it in any. It also attains the best average accuracy among training-free methods in all six settings, although per-benchmark gains are often within one standard deviation and TGAS trails the strongest baseline in four of 18 pairs. Baselines support our diagnosis: Forced-Answer, which also intervenes on stopping, is competitive when its budget matches problem difficulty; No-Repeat suppresses surface repetition but often hurts accuracy, especially on code; and SEAL, which suppresses reflection in general rather than looping specifically, raises termination at the cost of accuracy in several settings.

\paragraph{Does TGAS Fix the Diagnosed Failure?} Table~\ref{tab:steer-deltacyc-main} shows that TGAS makes $\Delta_{\mathrm{cyc}}$ more negative in all settings and substantially reduces decision points, roughly halving them at 60\% sparsity ($12.3\to5.8$ and $11.5\to6.8$ per 1k tokens for 7B and 14B, detailed in Appendix~\ref{app:behavior}). TGAS also reduces generation length by 7.6\% on average across all sparsity levels, including the dense model, and by 12.3\% at 60\% sparsity (Table~\ref{tab:tokens-main}). These results support our hypothesis that TGAS reduces repetitive reconsideration (Section~\ref{subsec:behavior}). 


\paragraph{Does TGAS Generalize Across Pruning Methods?} On DS-Qwen-7B, Table~\ref{tab:ablation-pruners-main} shows that TGAS consistently improves accuracy across pruning methods, outperforming the pruned baseline in all eight settings and achieving the best accuracy in seven. It also restores termination on degraded checkpoints, improving both accuracy and termination, unlike No-Repeat, which often trades accuracy for termination. At 60\% sparsity, where RAC- and Wanda-pruned models retain little accuracy, all training-free methods remain far below the dense model (Appendix~\ref{app:different_pruning_method}, which also covers other models). Thus, TGAS transfers across pruning algorithms and calibration sets, provided the pruned model retains sufficient capability.

\paragraph{Comparison with Training-Based Recovery.} Figure~\ref{fig:ablation_7B} shows that Masked-LoRA yields limited gains at 50\% sparsity but substantially improves recovery at 60\%, particularly on LCB, raising accuracy from 14.7\% to 33.4\% for DS-Qwen-7B and from 37.6\% to 51.8\% for 14B counterpart (Appendix~\ref{app:lora}). This suggests that Masked-LoRA recovers capability lost to aggressive pruning. However, Masked-LoRA is trained on traces from the dense DS-Qwen-14B model, even for smaller models, confounding recovery with distillation. This contrast supports our diagnosis: under moderate pruning, much of the loss stems from failure to stop, which TGAS addresses without training; at higher sparsity, some capability is genuinely lost and can only be restored by updating the weights. Such recovery, however, requires a separate adapter and optimization procedure for each sparsity level and model.

\paragraph{Ablation Studies.} Figure~\ref{fig:ablation_7B} summarizes the ablations on DS-Qwen-7B (other models in Appendix~\ref{app:full_ablation}). For Forced-Answer, reducing the reasoning budget consistently hurts accuracy, with 4K performing worse than 16K on both AIME-25 and LCB. For No-Repeat, 8-gram blocking is substantially more harmful than 32-gram blocking, particularly at 60\% sparsity. For TGAS, increasing $\alpha$ consistently improves termination, especially on LCB, so $\alpha$ provides a controllable knob for balancing termination and accuracy. We also test TGAS robustness to the number of samples and checkpoint choice, finding that 50 samples per class suffice (Appendix~\ref{app:number_sample_create_direction}) and directions from different checkpoints of the same model remain effective (Appendix~\ref{app:different-direction}). We also compare TGAS with Plan-CoT~\citep{sheng2025reasoning} and report significance tests with $K=16$ on AIME-25 and LCB in Appendix~\ref{app:Plan-CoT-comparison} and ~\ref{app:significant_test}, respectively.

\section{Conclusion}
We find that under moderate pruning, a substantial and recoverable share of accuracy loss stems from disrupted termination rather than incorrect reasoning. As sparsity increases, pruned LRMs exhibit reasoning divergence: they often reach the correct answer and recognize where to stop under teacher forcing, yet fall into repetitive loops and fail to commit within the token budget. The hidden-state contrast between terminating and looping continuations is consistent across pruned checkpoints; building on it, TGAS steers pruned models toward termination without retraining, improving termination and accuracy across configurations. At higher sparsity, some capability is genuinely lost and can only be recovered by updating the weights, suggesting that future pruning methods should preserve termination behavior.

\clearpage
\section*{Limitations}
Our work focuses on post-pruning recovery and does not modify the pruning process itself. Thus, the proposed interventions recover degraded behavior after pruning rather than preventing it during pruning. Future work could incorporate these observations into pruning algorithms to preserve termination and reasoning behavior directly, alongside a more theoretical characterization of the relationship between pruning and reasoning dynamics.

Following prior work on pruning LRMs~\citep{prune-RAC,OBC-prune,beware_calibration}, we study unstructured sparsity; semi-structured patterns, which SparseGPT and Wanda also support, are a natural extension. All models are distilled or SFT-trained LRMs of at most 14B parameters, and the benchmarks cover mathematics and code only; whether the diagnosis holds for RL-trained reasoning models, larger models, or open-ended tasks remains open. Our analysis uses a 16{,}384-token budget, and TGAS requires sampling continuations from a pruned checkpoint once per model to construct the direction. Finally, TGAS targets termination; at high sparsity, where capability is genuinely lost, it recovers only part of the gap to the dense model.

\section*{Ethical Considerations}
This work studies inference-time interventions for pruned reasoning models and does not introduce new data collection involving human subjects. The proposed methods can alter model generation behavior, including its tendency to terminate or continue reasoning; therefore, they should be evaluated for task-specific reliability before deployment.

\bibliography{citation}
\appendix
\etocdepthtag.toc{appendix}
\etocsettagdepth{mainbody}{none}
\etocsettagdepth{appendix}{subsection}
\renewcommand{\contentsname}{Appendices}
\tableofcontents
\raggedbottom
\section{Related Works}
\label{app:related_work}
\subsection{Pruning Methods for Large Language Models}

Pruning has long been explored as a means of reducing the computational and memory costs of neural networks while preserving model performance \citep{NIPS1989_6c9882bb,NIPS1992_303ed4c6}. Recent work has extended pruning to large language models (LLMs), with methods broadly falling into unstructured weight pruning and structured architectural pruning \citep{cheng2024survey}. For unstructured pruning, \citet{sparsegpt} proposed SparseGPT, a post-training pruning method based on approximate second-order information that enables one-shot pruning of GPT-scale models without subsequent retraining. In contrast, Wanda \citep{wanda} estimates weight importance from the product of weight magnitude and input activation magnitude, providing a simple activation-aware pruning criterion without weight reconstruction or retraining. 

For structured pruning, LLM-Pruner \citep{llm_pruner} removes coupled structures according to gradient-based importance and subsequently recovers performance through lightweight fine-tuning. ShortGPT \citep{men2025shortgpt} instead identifies redundant transformer blocks using block influence and directly removes entire layers. Sheared-LLaMA \citep{xia2024sheared} further considers pruning layers and attention heads, followed by continued pre-training to recover the capacity of the compressed model. \citet{kim2024shortened} further shows that simple depth pruning can effectively compress LLMs while achieving comparable or superior performance to other post-pruning finetuning methods. Complementary work examines calibration data in post-training pruning \citep{bandari2024c4,williams2024impact}. While pruning often uses small calibration sets, \citet{beware_calibration} show that calibration choice substantially affects pruning quality, especially at high sparsity, and that self-generated data can be an effective alternative when pre-training data are unavailable.

\subsection{Pruning Methods for Large Reasoning Models}

Recent LRMs, such as DeepSeek-R1 \citep{deepseek_r1} and Qwen3.5 \citep{qwen35blog}, introduce a distinct challenge for model compression. Unlike conventional LLMs, whose computation is often dominated by processing the input context, LRMs can generate long chain-of-thought (CoT) trajectories during autoregressive decoding. Consequently, preserving the input-to-output behavior alone may not be sufficient to preserve reasoning capability \citep{chen2026towards}. Recent studies have therefore begun to adapt pruning to the decode-dominated nature of LRMs \citep{paser,less}. \citet{prune-RAC} show that conventional pruning can degrade reasoning performance and increase inference cost by inducing longer reasoning trajectories. They attribute this to the mismatch between input-oriented activation reconstruction and decode-time reasoning behavior, and propose Reasoning-Aware Compression (RAC), which incorporates both input tokens and on-policy reasoning trajectories.

Complementing this direction, \citet{resp} propose RESP, a self-reflective structured pruning framework that explicitly aligns pruning decisions with the reasoning dynamics of reasoning language models. Their approach uses self-generated reasoning traces as calibration data, decode-only gradient-based importance estimation, and progressive regeneration of calibration traces during pruning. Building on these findings, \citet{ssgr} argue that model-generated reasoning trajectories offer a more informative calibration signal for pruning LRMs than conventional human-written or generic language-modeling data. 

Beyond individual pruning methods, \citet{zhang2026when} systematically investigate how different compression techniques, including pruning, quantization, and distillation, affect large reasoning models, with a particular focus on the components associated with reasoning degradation. Their findings suggest that compression can have distinct effects on memorization and reasoning, and that certain components, particularly MLP modules in later layers, play a disproportionately important role in preserving reasoning capabilities. Complementary studies on quantization further show that reasoning models exhibit distinct sensitivity to reduced numerical precision compared with conventional LLMs \citep{liu2025quantization}. More recently, \citet{lotfi2026quantized} investigate the overthinking phenomenon in quantized reasoning models, highlighting another potential consequence of model compression on the reasoning behavior of LRMs.

\section{Preliminaries}
\label{app:preliminaries}

Pruning reduces an LLM's size by setting selected weights to zero while preserving its performance. Most one-shot methods formulate pruning as a layer-wise reconstruction problem~\citep{obc}, finding a sparse weight matrix $\hat{\mathbf{W}}_\ell$ that preserves the original layer outputs on representative calibration inputs $\hat{\mathbf{W}}_\ell = \operatorname*{argmin}_{\|\hat{\mathbf{W}}_\ell\|_0 \le S} \|\mathbf{W}_\ell \mathbf{X}_\ell - \hat{\mathbf{W}}_\ell \mathbf{X}_\ell\|_F^2.$ guides which weights can be removed, while $S$ specifies how many weights are retained.
 
In this work, we focus on two representative unstructured one-shot pruning methods, SparseGPT \citep{sparsegpt} and Wanda \citep{wanda}, due to their computational efficiency and ability to prune large language models without costly retraining. SparseGPT \citep{sparsegpt} scores weights via the layer Hessian $\mathbf{H}_\ell = \mathbf{X}_\ell \mathbf{X}_\ell^\top$, using $w_q^2 / (\mathbf{H}_\ell^{-1})_{qq}$ and compensating for removal-induced errors. Wanda \citep{wanda} instead uses the diagonal score $s_{ij} = |w_{ij}| \cdot \|\mathbf{x}_j\|_2$ without compensation. 

Both methods depend on calibration activations, but standard text calibration may not adequately capture the activation distribution during extended reasoning \cite{liu2025quantization}. We therefore combine one-shot pruning methods with RAC~\citep{prune-RAC} and OBC~\citep{OBC-prune} to construct calibration sets tailored to pruned LRMs. Specifically, RAC augments prompt activations with model-generated chain-of-thought activations, forming $\mathbf{X}_\ell = [\mathbf{X}_\ell^{\text{prompt}}, \mathbf{X}_\ell^{\text{CoT}}]$ to estimate parameter importance using inputs that better reflect deployment-time reasoning. Besides, OBC generates difficulty-matched correct and incorrect rollouts and assigns token-level calibration weights according to each reasoning step's causal importance. These weights rescale calibration activations before one-shot pruning, helping preserve parameters critical to correct reasoning.

\section{Detailed Experimental Setup}
\label{app:experimental_setup}

This appendix describes the experimental protocol used to investigate the behavioral changes induced by weight pruning. 

\subsection{Generation Setup}
\label{app:generation_setup}

\textbf{Setting.} We evaluate DeepSeek-R1-Distill-Qwen, DeepSeek-R1-Distill-Llama~\citep{deepseek_r1}, and OpenThinker3~\citep{guha2026openthoughts} under one-shot unstructured pruning of decoder linear layers at varying sparsity levels. We construct calibration data using OBC, RAC, or C4~\citep{bandari2024c4}, with SparseGPT and Wanda as pruning backbones. We consider one-shot unstructured pruning at nominal sparsities of $10$-$70\%$. Pruning is restricted to decoder linear layers, including the MLP projections, full-attention $q/k/v/o$ projections. Embeddings and \texttt{lm\_head} remain dense, followed \citet{sparsegpt}. 

\begin{itemize}
   \item \textbf{OBC/SparseGPT}: OBC constructs difficulty-matched correct and incorrect rollouts and assigns per-token calibration weights based on the causal importance of reasoning steps. These weights rescale calibration activations before one-shot pruning methods, preserving parameters that are more important for correct reasoning.
    \item \textbf{RAC/SparseGPT}: The dense LRM generates chain-of-thought trajectories for the
    calibration prompts, and hidden states collected during autoregressive
    decoding are concatenated with prompt activations before being passed to
    SparseGPT. This produces a calibration distribution that better reflects
    the decode-heavy reasoning workload.
    \item \textbf{C4/Wanda}: the same C4 calibration set and pruning plan, using Wanda's activation-norm criterion instead of SparseGPT's second-order statistic.
\end{itemize}

Our evaluation covers 262 Level 4--5 MATH-500 questions, 30 AIME-25 questions, and 95 easy/medium LiveCodeBench questions. We evaluate on held-out problems from three benchmarks spanning mathematical
reasoning, competition-level problems, and code generation:
\begin{itemize}
    \item \textbf{MATH-500}: 262 samples, covering two most difficult levels and seven subjects.
    \item \textbf{AIME-25}: all 30 problems, providing a challenging multi-step mathematical reasoning benchmark.
    \item \textbf{LiveCodeBench-v6}: 95 easy and medium problems from code contests held in 2025. These problems post-date the models' training data and therefore provide a test of generalization rather than memorization. Since steering vectors are constructed from mathematical trajectories, LiveCodeBench additionally evaluates cross-domain transfer.
\end{itemize}

Unless otherwise specified, generation uses a 16,384-token budget, \(T=0.6\) and top-\(p=0.95\). For each question, we generate \(k=4\) responses for Math-500 and LCB, and AIME-25, reporting the average for more stable evaluation. We use vLLM~\citep{vllm} for inference and Lighteval~\citep{lighteval} for evaluation. 

\paragraph{Computing infrastructure.} All experiments were run on 1 NVIDIA H200 144GB GPU. Pruning a checkpoint with SparseGPT/Wanda takes about 1 GPU hour; constructing a TGAS direction (sampling continuations and computing hidden states) takes about 1 GPU hour per model; evaluation across all benchmarks takes about 2 GPU hours per checkpoint and method. Masked-LoRA training takes about 6 GPU hours per checkpoint.

\subsection{Measuring Answer Presence in Reasoning Divergence}
\label{app:presence}

Final-answer correctness alone does not determine whether a model has successfully reasoned to the correct solution. In particular, a non-termination trace may contain the correct answer in its reasoning segment even though the generation never reaches the final-answer stage. We therefore introduce a separate \emph{answer-presence} measurement that asks whether the reasoning process itself reaches or asserts the gold answer.

We therefore modify the extraction procedure to identify candidate values asserted throughout the reasoning trace, rather than extracting a single final answer, and evaluate each candidate using the same grading procedure. To account for the many intermediate numerical values in long traces, we define a hierarchy of increasingly permissive answer-presence criteria:
\begin{itemize}
    \item \textbf{Explicit}: an explicit answer in the think segment, identified by either a \verb|\boxed{}| or \verb|\fbox{}| expression matching the gold answer, or a statement of the form \emph{"the (final) answer is $X$"} or \emph{"Answer: $X$"}, where the remainder of the line is taken as $X$.
    \item \textbf{Conclusive}: an \emph{Explicit} answer or an \emph{Assertion}, where an \emph{Assertion} is a sentence containing a conclusion cue, detailed \emph{"so", "thus", "therefore", "hence", "the answer", "which is/gives/means/equals", "we get", or "in total")} whose text after the last copula with \emph{"is", "equals", "gives", "yields", "="}, or \emph{"$\approx$"} matches the gold answer, following \citet{lotfi2026quantized}.
\end{itemize}

\begin{figure*}[ht]
    \centering
 \includegraphics[width=0.85\linewidth]{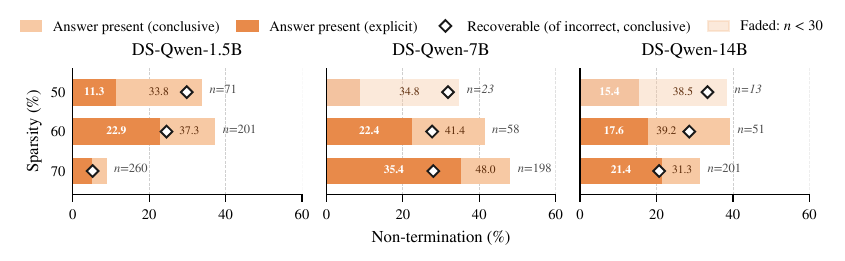}
\caption{Gold answers appearing in reasoning traces graded incorrect for DeepSeek-R1-Distill-Qwen models. \emph{Explicit} and \emph{Conclusive} denote gold-answer presence under nested detection rules.}
    \label{fig:answer_in_think_full}
\end{figure*}

The nested design distinguishes explicit commitment from incidental occurrences. In particular, \emph{Conclusive} is our primary answer-presence criterion, as it captures cases where the model reaches the correct result but fails to transition to a final answer. Figure~\ref{fig:answer_in_think_full} shows that many incorrectly graded traces nevertheless contain the correct answer with Explicit or Conclusive mentioned. At 60\% sparsity, this occurs in $41.4\%$ and $39.2\%$ of non-terminating traces for DeepSeek-R1-Distill-Qwen 7B and 14B, respectively. These results show that correct answers are often already present in non-termination reasoning traces, rather than appearing only as incidental intermediate values.

\begin{table}[t]
\centering
\caption{Agreement between our automated measurements and the independent Gemini-2.5-Flash judge.}
\label{tab:judge}
\footnotesize
\begin{tabular}{@{}lrrr@{}}
\toprule
comparison & $n$ & agreement & $\kappa$ \\
\midrule
\addlinespace[2pt]
\emph{Conclusive} vs judge & 250 & 96.0\% & 0.898 \\
\emph{Explicit} vs judge & 250 & 84.8\% & 0.782 \\
\bottomrule
\end{tabular}
\end{table}

\subsection{Independent Validation with an LLM Judge}
\label{app:judge}

\begin{figure*}[t]
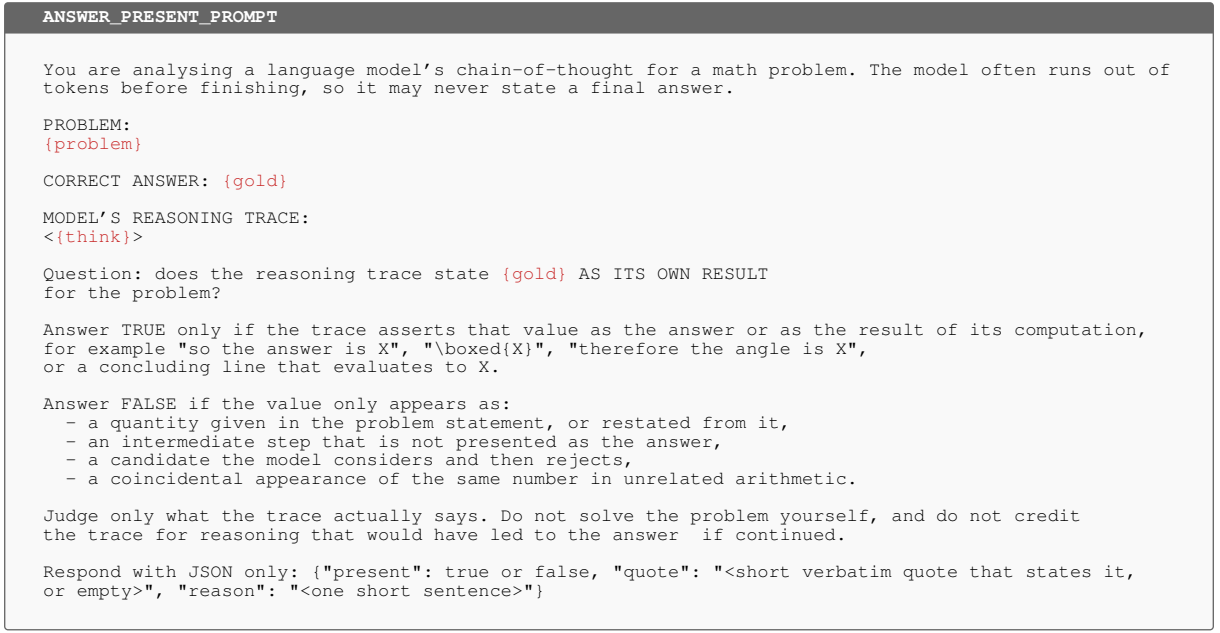

\centering
\begin{tcolorbox}[
    colback=gray!5,
    colframe=black!60,
title=\texttt{ANSWER\_PRESENT\_PROMPT},
    fonttitle=\bfseries\scriptsize,
    fontupper=\fontsize{8pt}{11pt}\selectfont,
    boxrule=0.5pt,
    arc=1pt
]
\begin{lstlisting}[
    style=promptstyle,
    basicstyle=\ttfamily\scriptsize,
    aboveskip=2pt, belowskip=2pt,
    lineskip=-1pt
]
You are analysing a language model's chain-of-thought for a math problem. The model often runs out of 
tokens before finishing, so it may never state a final answer.

PROBLEM:
(*\textcolor{promptplaceholder}{\{problem\}}*)

CORRECT ANSWER: (*\textcolor{promptplaceholder}{\{gold\}}*)

MODEL'S REASONING TRACE:
<(*\textcolor{promptplaceholder}{\{think\}}*)>

Question: does the reasoning trace state (*\textcolor{promptplaceholder}{\{gold\}}*) AS ITS OWN RESULT 
for the problem?

Answer TRUE only if the trace asserts that value as the answer or as the result of its computation, 
for example "so the answer is X", "\boxed{X}", "therefore the angle is X", 
or a concluding line that evaluates to X.

Answer FALSE if the value only appears as:
  - a quantity given in the problem statement, or restated from it,
  - an intermediate step that is not presented as the answer,
  - a candidate the model considers and then rejects,
  - a coincidental appearance of the same number in unrelated arithmetic.

Judge only what the trace actually says. Do not solve the problem yourself, and do not credit 
the trace for reasoning that would have led to the answer  if continued.

Respond with JSON only: {"present": true or false, "quote": "<short verbatim quote that states it, 
or empty>", "reason": "<one short sentence>"}
\end{lstlisting}
\end{tcolorbox}
\caption{Prompt used to independently validate whether a reasoning trace asserts the reference answer as its own result.}
\label{fig:present_prompt}
\end{figure*}

To independently validate both our deterministic answer-grading procedure and the answer-presence measurement, we use Gemini-2.5-Flash\footnote{\url{ai.google.dev/gemini-api/docs/models/gemini-2.5-flash}} as an external judge on the responses generated by the DeepSeek-R1-Distill-Qwen-7B model. The judge is used strictly for validation: its outputs are never written back to the dataset, used to modify the correct labels, or included in the main analysis. We validate answer presence in the reasoning trace with detailed prompts in Figure \ref{fig:present_prompt}.

\paragraph{LLM-as-a-judge setup.} For independent validation, we use Gemini-2.5-Flash at temperature $0$. The judge evaluates answer-presence agreement whether the reasoning trace explicitly asserts the gold result. We sample 50 traces from each of five sparsity levels ($10\%,40\%-70\%$), 250 in total.

\paragraph{Answer-presence validation.}
Given the problem, gold answer, and full reasoning trace, the judge determines whether the model explicitly arrives at the gold answer, even if the trace does not terminate. The judge must not solve the problem independently or infer an answer from incomplete reasoning. It must also distinguish the final answer from intermediate calculations, rejected candidates, copied values, and coincidental matches.

\paragraph{Agreement with the independent judge.}
The independent evaluation shows strong agreement with our automated measurements. On the same random sample, \emph{Conclusive} achieves \(96.0\%\) agreement with the judge and \(\kappa=0.898\), while the more conservative \emph{Explicit} criterion achieves \(84.8\%\) agreement and \(\kappa=0.782\). These results support \emph{Conclusive} as the primary measure of answer presence and confirm the reliability of our automated evaluation, detailed in Table~\ref{tab:judge}.

\subsection{Training Baseline}
\label{app:detail_training_baseline}

\subsubsection{Masked LoRA Recovery}
To assess whether pruning degradation can be recovered without restoring zeroed weights, we extend masked fine-tuning \citep{llm_pruner,xia2024sheared,kim2024shortened} to LRMs by restricting LoRA updates strictly to the surviving support. For a pruned weight $\widehat{W}_\ell$ and its binary support $\Omega_\ell$, the update is defined as:
\begin{equation}
\begin{split}
& \widetilde{W}_\ell = \widehat{W}_\ell + s \cdot \bigl(P_\ell Q_\ell \odot \Omega_\ell\bigr), \\
& [\Omega_\ell]_{ij} = \mathbf{1}\left[[\widehat{W}_\ell]_{ij} \neq 0\right]
\end{split}
\label{eq:masked-lora}
\end{equation}
where $P_\ell$ and $Q_\ell$ are trainable low-rank adapters with rank $\rho=64$, and $\Omega_\ell$ masks pruned positions to preserve sparsity. We apply LoRA to 200 decoder linear layers, excluding gate matrices, embeddings, and tied heads, and train for one epoch. After training, we merge the adapters and verify sparsity. We evaluate accuracy and termination rate to assess recovery from non-termination.

\paragraph{Supervision signal.}
We train on verified-correct reasoning traces from OpenR1-Math, generated by dense DeepSeek-R1-Distill-Qwen 14B. Target traces are fixed across all checkpoints so that every model is trained on identical supervision. Prompts are masked, supervising only the response span $\mathcal{T}(x)$ over a gradient-accumulation group $\mathcal{G}$:

\begin{equation*}
\begin{split}
\mathcal{L}_{\mathcal{G}}(\theta) &= -\frac{1}{N_{\mathcal{G}}} \sum_{x \in \mathcal{G}} \sum_{t \in \mathcal{T}(x)} \log p_\theta\bigl(x_{t+1} \mid x_{\leq t}\bigr), \\
N_{\mathcal{G}} &= \sum_{x \in \mathcal{G}} \left|\mathcal{T}(x)\right|.
\end{split}
\label{eq:token-weighted-ce}
\end{equation*}

Normalizing by the total number of supervised tokens avoids bias toward shorter sequences. Since Masked-LoRA uses verified reasoning traces from a dense 14B model, its performance reflects both recovery and distillation, particularly for smaller models. We therefore treat it as a reference for training-free methods, serving as an upper bound only at high sparsity (60\%--70\% sparse ratios).

\subsubsection{Forced-Termination Intervention.}
We investigate whether non-termination in pruned models can be mitigated by explicitly enforcing termination after sufficiently long reasoning traces. Our analysis shows that pruned models largely preserve both reasoning traces and the stopping mechanism of dense models, particularly at moderate sparsity, motivating this intervention. We therefore introduce a simple forced-termination intervention: given a reasoning trace, we append the \texttt{</think>} token to terminate the reasoning phase and then prompt the model to produce the final answer using an explicit pattern such as "Answer:" after $D$ generation budgets. This intervention prevents further free-running generation and directly tests whether the existing reasoning trace is sufficient to recover the final answer without modifying the model parameters. 

In practice, the difficulty of a user query is unknown, making it difficult to select an appropriate $D$. Moreover, tuning $D$ on a development set may not generalize across benchmarks with different difficulty distributions. We therefore use a fixed budget of $D=16{,}384$ tokens for AIME-25 and $D=8,192$ for MATH-500 and LCB and evaluate alternative budgets in Appendix~\ref{app:ablation-force-answer}.

\subsubsection{No-Repeat}
Our observations in Section \ref{sec:observation} show that unconstrained decoding of pruned LRMs can fall into cyclic loops. To prevent this, we forbid the model from repeating any $n$-gram already present in its context. Let $y_{<t} = (y_1, \dots, y_{t-1})$ be the tokens generated so far and $\mathbf{z}_t \in \mathbb{R}^{|\mathcal{V}|}$ the logits at step $t$. We define the set of banned tokens as those that would complete an $n$-gram whose $(n-1)$-token prefix matches the current suffix: $\mathcal{B}_t = \{ y_{i+n-1} \mid 1 \le i \le t-n,\ (y_i, \dots, y_{i+n-2}) = (y_{t-n+1}, \dots, y_{t-1}) \}$. The adjusted logits are:
\begin{equation}
\tilde{z}_{t,v} =
\begin{cases}
-\infty, & v \in \mathcal{B}_t, \\
z_{t,v}, & \text{otherwise},
\end{cases}
\end{equation}
and the next token is drawn from $\operatorname{softmax}(\tilde{\mathbf{z}}_t)$\footnote{Implemented with \texttt{no\_repeat\_ngram\_size} in Hugging Face \texttt{transformers}.}.

We evaluate $n$-gram sizes in $\{8,32\}$ on the held-out OpenR1-Math set and select $n=32$ as the default based on its performance. Ablation results are reported in Appendix~\ref{app:ablation-no-repeat}.

\subsubsection{SEAL: Steerable Reasoning Calibration of LRMs}
SEAL~\citep{chen2025seal} is a training-free activation-steering method that
reduces redundant reasoning in LRMs. It segments a reasoning trace into
thoughts separated by \texttt{\textbackslash n\textbackslash n} and labels each
thought by keyword rules: a thought is a \emph{transition} if it starts with
\emph{"Alternatively"} or contains phrases such as \emph{"another approach"}, a
\emph{reflection} if it starts with \emph{"Wait"} or contains phrases such as
\emph{"let me check"} or \emph{"make sure"}, and an \emph{execution} thought otherwise. SEAL finds that, for problems of comparable difficulty, incorrect responses contain more reflection and transition thoughts than correct ones. Since these thought types are separable in the representation space, SEAL constructs activation steering vectors to encourage desired reasoning behaviors, steering LRMs toward more \emph{execution} thoughts.

To extract a steering vector, SEAL collects the layer-$\ell$ hidden states at the \texttt{\textbackslash n\textbackslash n} token closing each thought on a
calibration set, averages them per category, and takes the difference
between execution thoughts and the union of reflection and transition thoughts.
\begin{equation*}
    \mathbf{S} = \bar{\mathbf{H}}_{E} - \bar{\mathbf{H}}_{RT},
\end{equation*}
with $E$, $R$, and $T$ denote \emph{execution}, \emph{reflection}, and \emph{transition} thoughts, respectively. During decoding, SEAL adds $\alpha\,\mathbf{S}$ to the hidden state of every
\texttt{\textbackslash n\textbackslash n} token at layer $\ell$, so the
intervention is applied only at step boundaries. We construct $\mathbf{S}$ from calibration rollouts on OpenR1-Math, the same source used for TGAS, apply it at the layers listed in Table~\ref{tab:hparams}, and select its strength from
$\{0.25, 0.5, \ldots, 1.25, 1.5\}$ per checkpoint.

\begin{table*}[t]
\begin{center}
\scriptsize
\setlength{\tabcolsep}{3pt}
\resizebox{0.8\textwidth}{!}{%
\begin{tabular}{ll rrrrr rrrrr rrrrr}
\toprule
 & & \multicolumn{5}{c}{MATH-500 L4--5} & \multicolumn{5}{c}{AIME-25} & \multicolumn{5}{c}{OpenR1-Math} \\
\cmidrule(lr){3-7}\cmidrule(lr){8-12}\cmidrule(lr){13-17}
\thd{Model} & \thd{Ckpt} & \thd{Prob.} & \thd{BP} & \thd{POS} & \thd{NEG} & \thd{Len} & \thd{Prob.} & \thd{BP} & \thd{POS} & \thd{NEG} & \thd{Len} & \thd{Prob.} & \thd{BP} & \thd{POS} & \thd{NEG} & \thd{Len} \\
\midrule
\multirow{4}{*}{\makecell{Deepseek-R1-Distill\\-Qwen-1.5B}} & \ckpt{40} & \multirow{4}{*}{\num{91}} & \multirow{4}{*}{\num{91}} & \num{2780} & \num{132} & \num{1742} & \multirow{4}{*}{\num{8}} & \multirow{4}{*}{\num{14}} & \num{439} & \num{9} & \num{1515} & \multirow{4}{*}{\num{112}} & \multirow{4}{*}{\num{120}} & \num{3433} & \num{407} & \num{2585} \\
 & \ckpt{50} &  &  & \num{2545} & \num{367} & \num{1575} &  &  & \num{410} & \num{38} & \num{1347} &  &  & \num{2948} & \num{892} & \num{2484} \\
 & \ckpt{60} &  &  & \num{953} & \num{1959} & \num{2695} &  &  & \num{260} & \num{188} & \num{2076} &  &  & \num{1034} & \num{2806} & \num{3392} \\
 & \ckpt{70} &  &  & \num{0} & \num{2912} & --- &  &  & \num{0} & \num{448} & --- &  &  & \num{0} & \num{960} & --- \\
\midrule
\multirow{4}{*}{\makecell{Deepseek-R1-Distill\\-Qwen-7B}} & \ckpt{40} & \multirow{4}{*}{\num{80}} & \multirow{4}{*}{\num{80}} & \num{2554} & \num{6} & \num{1577} & \multirow{4}{*}{\num{8}} & \multirow{4}{*}{\num{13}} & \num{415} & \num{1} & \num{1822} & \multirow{4}{*}{\num{72}} & \multirow{4}{*}{\num{103}} & \num{3229} & \num{67} & \num{2729} \\
 & \ckpt{50} &  &  & \num{2504} & \num{56} & \num{1630} &  &  & \num{407} & \num{9} & \num{1870} &  &  & \num{3012} & \num{284} & \num{2697} \\
 & \ckpt{60} &  &  & \num{2172} & \num{388} & \num{2353} &  &  & \num{355} & \num{61} & \num{2350} &  &  & \num{2300} & \num{996} & \num{3182} \\
 & \ckpt{70} &  &  & \num{893} & \num{1667} & \num{3203} &  &  & \num{161} & \num{255} & \num{2391} &  &  & \num{226} & \num{598} & \num{3360} \\
\midrule
\multirow{4}{*}{\makecell{Deepseek-R1-Distill\\-Qwen-14B}} & \ckpt{40} & \multirow{4}{*}{\num{72}} & \multirow{4}{*}{\num{72}} & \num{2303} & \num{1} & \num{1431} & \multirow{4}{*}{\num{11}} & \multirow{4}{*}{\num{17}} & \num{544} & \num{0} & \num{1368} & \multirow{4}{*}{\num{63}} & \multirow{4}{*}{\num{100}} & \num{6382} & \num{18} & \num{2060} \\
 & \ckpt{50} &  &  & \num{2302} & \num{2} & \num{1558} &  &  & \num{544} & \num{0} & \num{1566} &  &  & \num{6293} & \num{107} & \num{2242} \\
 & \ckpt{60} &  &  & \num{2127} & \num{177} & \num{1966} &  &  & \num{532} & \num{12} & \num{2060} &  &  & \num{5253} & \num{1147} & \num{2775} \\
 & \ckpt{70} &  &  & \num{1050} & \num{1254} & \num{3359} &  &  & \num{340} & \num{204} & \num{3844} &  &  & \num{333} & \num{467} & \num{3667} \\
\bottomrule
\end{tabular}
}
\end{center}
\caption{Branch points and continuation outcomes across pruning ratios. \thd{Prob.} denotes problems with at least one branch point; \thd{BP}, the number of branch points; \thd{POS}/\thd{NEG}, self-terminating/non-terminating continuations; and \thd{Len}, median positive-continuation length. Despite sharing prefixes from dense traces sufficient for correct answers, higher-pruned checkpoints produce more negative continuations and longer median continuation lengths.}
\label{tab:exp5-continuation-stats}
\end{table*}

\begin{table*}[t]
\begin{minipage}[t]{0.42\textwidth}
\begin{center}
\tabsmall
\setlength{\tabcolsep}{4pt}
\resizebox{\linewidth}{!}{%
\begin{tabular}{llr r rrr}
\toprule
\thd{Model} & \thd{Ckpt} & \thd{N} & \thd{Branch pts} & \thd{POS} & \thd{NEG} & \thd{Len} \\
\midrule
\multirow{2}{*}{\makecell[l]{Deepseek-\\Distill-Llama-8B}}
  & \ckpt{50} & 8 & \num{103} & \num{810}  & \num{14}  & \num{1034} \\
  & \ckpt{60} & 8 & \num{64}  & \num{372}  & \num{140} & \num{1743} \\
\midrule
\makecell[l]{OpenThinker3-1.5B}
  & \ckpt{50} & 8 & \num{177} & \num{1289} & \num{127} & \num{3398} \\
\midrule
\multirow{2}{*}{\makecell[l]{OpenThinker3-7B}}
  & \ckpt{50} & 8  & \num{179} & \num{1381} & \num{51}  & \num{3398} \\
  & \ckpt{60} & 24 & \num{179} & \num{4132} & \num{164} & \num{3394} \\
\bottomrule
\end{tabular}}
\end{center}
\caption{Continuation outcomes for DeepSeek-R1-Distill-Llama-8B and OpenThinker3 from their steering calibration sets on OpenR1-Math. \thd{Branch pts}, \thd{POS}, \thd{NEG}, and \thd{Len} denote the number of branch points, positive continuations, negative continuations, and median continuation length, respectively.}
\label{tab:exp5-continuation-stats-v5}
\end{minipage}
\hfill
\begin{minipage}[t]{0.56\textwidth}
\scriptsize
\centering
\setlength{\tabcolsep}{3pt}
\resizebox{0.8\linewidth}{!}{%
\begin{tabular}{ll cccccc}
\toprule
& & \multicolumn{3}{c}{\textbf{DeepSeek-R1-Distill-Qwen}} & \textbf{\makecell{DeepSeek-R1-\\Distill-Llama}} & \multicolumn{2}{c}{\textbf{OpenThinker3}} \\
\cmidrule(lr){3-5} \cmidrule(lr){6-6} \cmidrule(lr){7-8}
& & 1.5B & 7B & 14B & 8B & 1.5B & 7B \\
\midrule
\multirow{6}{*}{\textbf{TGAS}}
 & Layer $\ell^\star$ & 18 & 20 & 29 & 9 & 20 & 26 \\
\cmidrule(lr){2-8}
 & Dense & $1$ & $1$ & $1$ & $2$ & $0.5$ & $1.5$ \\
 & 10\%  & $1.5$ & $1.25$ & $1.25$ & $2$ & $1.5$ & $1.5$ \\
 & 50\%  & $1.5$ & $0.5$ & $0.5$ & $1.25$ & $1$ & $1.25$ \\
 & 60\%  & $1.75$ & $1.25$ & $0.75$ & $1.5$ & $1$ & $1.5$ \\
 & 70\%  & $2$ & $1.5$ & $1$ & $2$ & $0.5$ & $1.25$ \\
\midrule
\multirow{6}{*}{\textbf{SEAL}}
 & Layer $\ell^*$ & 20 & 20 & 34 & 23 & 20 & 20 \\
\cmidrule(lr){2-8}
 & Dense & $1$ & $1$ & $0.5$ & $1$ & $1$ & $0.5$ \\
 & 10\%  & $1$ & $1$ & $0.5$ & $0.5$ & $1$ & $0.25$ \\
 & 50\%  & $1$ & $0.5$ & $0.5$ & $0.5$ & $1$ & $1$ \\
 & 60\%  & $0.5$ & $1$ & $0.25$ & $1$ & $1$ & $0.5$ \\
 & 70\%  & $1$ & $0.5$ & $1$ & $0.5$ & $1.25$ & $0.5$ \\
\bottomrule
\end{tabular}}
\caption{Hyperparameters for TGAS and SEAL. $\ell^*$ denotes the decoder layers. TGAS strength $\alpha$ is selected from $0.5\rightarrow2.0$ per checkpoint, and SEAL strength from $0.25\rightarrow1.5$ based on performance on the held-out OpenR1-Math set.}
\label{tab:hparams}
\end{minipage}
\end{table*}

\subsubsection{Termination-guided Activation Steering}
\label{app:construct_steer_vector}
\begin{figure}[t]
    \centering
    \includegraphics[width=\linewidth]{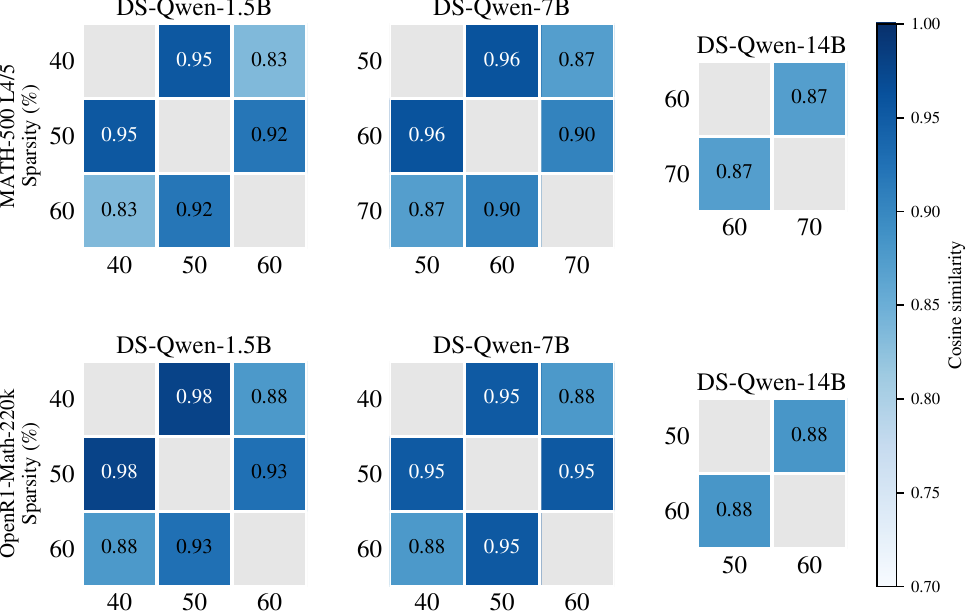}
    \caption{Cosine similarity between termination-guided directions from DeepSeek-R1-Distill-Qwen checkpoints. We include only checkpoints with sufficient positive and negative samples.}
    \label{fig:full_consine_directions}
\end{figure}
Figure \ref{fig:full_consine_directions} provides an additional observation that directions constructed from different pruned checkpoints can remain highly similar, suggesting a shared behavioral direction across pruning levels. A detailed analysis is provided in Appendix~\ref{app:commitment-direction-analysis}.

To test whether this direction generalizes beyond the calibration data, our benchmarking uses an out-of-distribution (OOD) setting: we construct one steering vector per model, from its medium-sparsity checkpoint and calibration rollouts on \texttt{open-r1/OpenR1-Math}, using at most six rollouts per problem per class, and evaluate it on separate downstream benchmarks. We evaluate on MATH-500, AIME-25, and LiveCodeBench-v6 to test generalization across unseen mathematical problems and a different domain. Each vector is constructed once and fixed across all pruning conditions, with calibration problems disjoint from evaluation sets. We provide the statistic of number of positive and negative continuations in Table \ref{tab:exp5-continuation-stats} and Table \ref{tab:exp5-continuation-stats-v5}. 

To select the steering layer, we split the branch points $\mathcal{B}$ into a fitting set $\mathcal{B}_{\mathrm{fit}}$ and a held-out set $\mathcal{B}_{\mathrm{val}}$, keeping all continuations of a branch point in the same set since they share a prefix. For each layer $\ell$, we compute $\mathbf{r}_\ell$ with Equation ~\eqref{eq:classmeans} on $\mathcal{B}_{\mathrm{fit}}$, project every held-out continuation onto it, $\langle \mathbf{s}_\ell(c), \mathbf{r}_\ell \rangle$, and measure the ROC AUC for separating terminating from non-terminating continuations. We select $\ell^\star=\arg\max_\ell \mathrm{AUC}(\ell)$ and build the steering vector $\mathbf{v}$ from $\mathbf{r}_{\ell^\star}$ as described in Section~\ref{sec:method_steer}. Algorithm~\ref{alg:steer} summarizes the full procedure.
\FloatBarrier
\subsubsection{Hyperparameter Tuning} 

TGAS and SEAL both use activation steering and are calibrated on the OpenR1-Math set. TGAS has two hyperparameters: the steering layer $\ell^\star$ and strength $\alpha$. We select $\ell^\star$ using calibration rollouts on the medium pruning checkpoints with TGAS and on dense model with SEAL and reuse it across all pruned checkpoints, while $\alpha$ is selected per checkpoint based on accuracy on the held-out OpenR1-Math split. We also compare with directions created from medium pruning checkpoints of SEAL at Appendix \ref{app:seal-comparison}. The resulting hyperparameters for all models and sparsity levels are reported in Table~\ref{tab:hparams}.

\clearpage
\newpage

\begin{algorithm*}[t]
\makeatletter
\renewcommand{\fnum@algorithm}{\small\fname@algorithm~\thealgorithm}
\makeatother
\caption{\small Training-free Post-pruning via Termination-Guided Activation Steering (TGAS)}
\label{alg:steer}
\small
\begin{algorithmic}[1]
   \INPUT Calibration checkpoint $M_c$ (intermediate sparsity) and target checkpoint $M_p$ of the same model; dense correct traces $\mathcal{D}_{\text{cov}}$; number of continuations $N$; layers $\{1,\dots,L\}$; \\ candidate strengths $\mathcal{A}$; held-out OpenR1-Math split $\mathcal{D}_{\mathrm{val}}$.
   \OUTPUT Steered generation of $M_p$ with steering vector $\mathbf{v}$ at layer $\ell^\star$.

   \STATE {\bf Step 1: Construct branch-point continuation sets} \COMMENT{once per model, with $M_c$}
   \FOR{each trace in $\mathcal{D}_{\text{cov}}$}
      \STATE Identify branch point $b$: first paragraph start after \emph{Conclusive} s.t.\ $b < t_{\mathrm{end}}$
      \STATE Teacher-force prefix $\mathbf{x}_{<b}$ through $M_c$; sample $N$ continuations
      \STATE Assign continuations to $\mathcal{C}^{+}_{b}$ (terminating) or $\mathcal{C}^{-}_{b}$ (non-terminating)
   \ENDFOR

   \STATE {\bf Step 2: Compute difference-in-means direction}
   \FOR{$\ell = 1, \dots, L$}
      \FOR{each branch point $b \in \mathcal{B}$}
         \FOR{each continuation $c \in \mathcal{C}^{+}_{b} \cup \mathcal{C}^{-}_{b}$}
            \STATE $\mathbf{s}_\ell(c) \leftarrow \big(\sum_{t=b}^{t_{\mathrm{end}}-1}\mathbf{h}_{\ell,t}(c)\big)/(t_{\mathrm{end}}-b)$ \COMMENT{up to the continuation's end if it is shorter}
         \ENDFOR
         \STATE $\boldsymbol{\mu}^{+}_{\ell,b} \leftarrow \big(\sum_{c \in \mathcal{C}^{+}_{b}} \mathbf{s}_\ell(c)\big)/|\mathcal{C}^{+}_{b}|$, \quad
                $\boldsymbol{\mu}^{-}_{\ell,b} \leftarrow \big(\sum_{c \in \mathcal{C}^{-}_{b}} \mathbf{s}_\ell(c)\big)/|\mathcal{C}^{-}_{b}|$
      \ENDFOR
      \STATE $\mathbf{r}_\ell \leftarrow \frac{1}{|\mathcal{B}|}\sum_{b \in \mathcal{B}} \bigl(\boldsymbol{\mu}^{+}_{\ell,b}-\boldsymbol{\mu}^{-}_{\ell,b}\bigr)$
   \ENDFOR

   \STATE {\bf Step 3: Build the steering vector}
   \STATE $\ell^\star \leftarrow \arg\max_\ell \; \mathrm{AUC}(\mathbf{r}_\ell)$ on held-out branch points \COMMENT{ROC AUC of the projection}
   \STATE $\bar{n}_{\ell^\star} \leftarrow$ mean $\ell_2$ norm of $\mathbf{h}_{\ell^\star,t}$ over calibration trajectories
   \STATE $\mathbf{v} \leftarrow \bar{n}_{\ell^\star}\,\mathbf{r}_{\ell^\star} / \|\mathbf{r}_{\ell^\star}\|_2$ \COMMENT{reused for all sparsity levels of the model}
   \STATE $\alpha \leftarrow \arg\max_{\alpha' \in \mathcal{A}} \mathrm{Acc}(M_p, \alpha'\mathbf{v})$ on $\mathcal{D}_{\mathrm{val}}$ \COMMENT{per checkpoint}

   \STATE {\bf Step 4: Apply steering at inference}
   \FOR{each token position $t$ during autoregressive generation}
      \STATE $\tilde{\mathbf{h}}_{\ell^\star, t} \leftarrow \mathbf{h}_{\ell^\star, t} + \alpha\, \mathbf{v}$
   \ENDFOR
   \STATE \textbf{return} steered outputs of $M_p$
\end{algorithmic}
\end{algorithm*}

\clearpage
\newpage

\section{Complete Observation Results}
\label{app:full_observation}

\subsection{Answer Retention and Termination in Pruned Models on Dense-Model-Correct Samples}
\label{app:corrected_dense_acc_full}

\begin{table*}[t]
\centering
\scriptsize
\setlength{\tabcolsep}{3pt}
\resizebox{0.7\textwidth}{!}{%
\begin{tabular}{ll rrr rrr rrr}
\toprule
 & & \multicolumn{3}{c}{AIME25} & \multicolumn{3}{c}{LCB} & \multicolumn{3}{c}{MATH-500 L4--5} \\
\cmidrule(lr){3-5}\cmidrule(lr){6-8}\cmidrule(lr){9-11}
family & level & $n$ kept & kept & & $n$ kept & kept & & $n$ kept & kept & \\
\midrule
\multirow{5}{2.0cm}{DeepSeek-R1-Distill-Qwen-1.5B (OBC)}
 & dense            & 27.0 $\pm$ 0.0 & 100 & & 133.0 $\pm$ 2.8 & 100 & & 197.0 $\pm$ 1.4 & 100 & \\
 & sparse10         & 19.5 $\pm$ 0.7 & 72.2 $\pm$ 2.6\% & & 107.5 $\pm$ 3.5 & 80.8 $\pm$ 0.9\% & & 175.0 $\pm$ 2.8 & 88.8 $\pm$ 2.1\% & \\
 & sparse50         & 15.0 $\pm$ 1.4 & 55.6 $\pm$ 5.2\% & & 51.0 $\pm$ 4.2 & 38.3 $\pm$ 2.4\% & & 133.5 $\pm$ 6.4 & 67.8 $\pm$ 2.7\% & \\
 & sparse60         & 1.5 $\pm$ 0.7 & 5.6 $\pm$ 2.6\% & & 0.0 $\pm$ 0.0 & 0.0 $\pm$ 0.0\% & & 39.5 $\pm$ 9.2 & 20.0 $\pm$ 4.5\% & \\
\midrule
\multirow{5}{2.0cm}{DeepSeek-R1-Distill-Qwen-7B (OBC)}
 & dense            & 47.0 $\pm$ 5.7 & 100 & & 223.5 $\pm$ 2.1 & 100 & & 223.0 $\pm$ 2.8 & 100 & \\
 & sparse10         & 33.0 $\pm$ 1.4 & 70.5 $\pm$ 5.5\% & & 198.0 $\pm$ 2.8 & 88.6 $\pm$ 2.1\% & & 208.0 $\pm$ 4.2 & 93.3 $\pm$ 0.7\% & \\
 & sparse50         & 30.0 $\pm$ 1.4 & 64.1 $\pm$ 4.7\% & & 145.0 $\pm$ 7.1 & 64.9 $\pm$ 2.5\% & & 204.5 $\pm$ 0.7 & 91.7 $\pm$ 0.8\% & \\
 & sparse60         & 21.0 $\pm$ 4.2 & 44.5 $\pm$ 3.7\% & & 55.5 $\pm$ 0.7 & 24.8 $\pm$ 0.6\% & & 173.5 $\pm$ 0.7 & 77.8 $\pm$ 0.7\% & \\
 & sparse70         & 6.0 $\pm$ 0.0 & 12.9 $\pm$ 1.5\% & & 0.0 $\pm$ 0.0 & 0.0 $\pm$ 0.0\% & & 57.0 $\pm$ 1.4 & 25.6 $\pm$ 1.0\% & \\
\midrule
\multirow{5}{2.0cm}{DeepSeek-R1-Distill-Qwen-14B (OBC)}
 & dense            & 64.0 $\pm$ 1.4 & 100 & & 278.0 $\pm$ 8.6 & 100 & & 231.5 $\pm$ 0.7 & 100 & \\
 & sparse10         & 47.5 $\pm$ 0.7 & 74.2 $\pm$ 0.5\% & & 254.0 $\pm$ 4.7 & 91.4 $\pm$ 1.7\% & & 220.0 $\pm$ 4.2 & 95.0 $\pm$ 1.5\% & \\
 & sparse50         & 37.5 $\pm$ 0.7 & 58.6 $\pm$ 0.2\% & & 225.0 $\pm$ 6.5 & 80.9 $\pm$ 2.4\% & & 206.5 $\pm$ 2.1 & 89.2 $\pm$ 0.6\% & \\
 & sparse60         & 29.5 $\pm$ 0.7 & 46.1 $\pm$ 0.1\% & & 140.0 $\pm$ 8.3 & 50.4 $\pm$ 3.0\% & & 172.0 $\pm$ 7.1 & 74.3 $\pm$ 2.8\% & \\
 & sparse70         & 3.0 $\pm$ 1.4 & 4.7 $\pm$ 2.1\% & & 0.0 $\pm$ 0.0 & 0.0 $\pm$ 0.0\% & & 40.5 $\pm$ 2.1 & 17.5 $\pm$ 1.0\% & \\
\midrule
\multirow{6}{2.0cm}{DeepSeek-R1-Distill-Llama-8B (OBC)}
 & dense            & 10.2 $\pm$ 1.7 & 100 & & 55.8 $\pm$ 2.4 & 100 & & 193.0 $\pm$ 7.1 & 100 & \\
 & sparse10         & 7.8 $\pm$ 1.7 & 76.5 $\pm$ 17.3\% & & 49.8 $\pm$ 2.9 & 89.3 $\pm$ 5.9\% & & 178.0 $\pm$ 3.7 & 92.2 $\pm$ 1.9\% & \\
 & sparse40         & 7.2 $\pm$ 1.0 & 73.2 $\pm$ 20.6\% & & 47.0 $\pm$ 2.2 & 84.3 $\pm$ 0.8\% & & 155.0 $\pm$ 5.5 & 80.3 $\pm$ 2.9\% & \\
 & sparse50         & 5.5 $\pm$ 1.7 & 54.2 $\pm$ 16.2\% & & 38.8 $\pm$ 1.0 & 69.6 $\pm$ 2.1\% & & 119.0 $\pm$ 6.8 & 61.7 $\pm$ 3.5\% & \\
 & sparse60         & 1.0 $\pm$ 0.8 & 9.6 $\pm$ 7.1\% & & 10.2 $\pm$ 1.5 & 18.4 $\pm$ 3.1\% & & 31.0 $\pm$ 5.1 & 16.1 $\pm$ 2.6\% & \\
 & sparse70         & 0.0 $\pm$ 0.0 & 0.0 $\pm$ 0.0\% & & 0.0 $\pm$ 0.0 & 0.0 $\pm$ 0.0\% & & 0.0 $\pm$ 0.0 & 0.0 $\pm$ 0.0\% & \\
\midrule
\multirow{6}{2.0cm}{OpenThinker3-1.5B (OBC)}
 & dense            & 9.8 $\pm$ 0.5 & 100 & & 41.0 $\pm$ 2.7 & 100 & & 199.0 $\pm$ 6.9 & 100 & \\
 & sparse10         & 7.8 $\pm$ 1.0 & 79.4 $\pm$ 8.2\% & & 35.2 $\pm$ 2.2 & 86.0 $\pm$ 1.8\% & & 172.0 $\pm$ 4.8 & 86.4 $\pm$ 2.4\% & \\
 & sparse40         & 6.2 $\pm$ 1.0 & 64.4 $\pm$ 12.1\% & & 25.5 $\pm$ 2.5 & 62.2 $\pm$ 3.7\% & & 139.0 $\pm$ 6.5 & 69.8 $\pm$ 3.3\% & \\
 & sparse50         & 5.2 $\pm$ 0.5 & 53.9 $\pm$ 4.8\% & & 8.5 $\pm$ 0.6 & 20.8 $\pm$ 1.2\% & & 69.0 $\pm$ 6.7 & 34.7 $\pm$ 3.4\% & \\
 & sparse60         & 0.2 $\pm$ 0.5 & 2.8 $\pm$ 5.6\% & & 0.0 $\pm$ 0.0 & 0.0 $\pm$ 0.0\% & & 7.0 $\pm$ 2.6 & 3.5 $\pm$ 1.3\% & \\
\midrule
\multirow{6}{2.0cm}{OpenThinker3-7B (OBC)}
 & dense            & 13.2 $\pm$ 1.5 & 100 & & 60.2 $\pm$ 1.9 & 100 & & 198.0 $\pm$ 7.0 & 100 & \\
 & sparse10         & 11.5 $\pm$ 1.3 & 87.1 $\pm$ 9.2\% & & 53.8 $\pm$ 1.7 & 89.3 $\pm$ 4.8\% & & 177.0 $\pm$ 4.3 & 89.4 $\pm$ 2.2\% & \\
 & sparse40         & 11.0 $\pm$ 1.8 & 82.7 $\pm$ 5.3\% & & 53.2 $\pm$ 1.0 & 88.4 $\pm$ 2.4\% & & 163.0 $\pm$ 5.4 & 82.3 $\pm$ 2.7\% & \\
 & sparse50         & 9.8 $\pm$ 0.5 & 74.4 $\pm$ 10.4\% & & 46.0 $\pm$ 1.6 & 76.3 $\pm$ 1.4\% & & 157.0 $\pm$ 5.7 & 79.3 $\pm$ 2.9\% & \\
 & sparse60         & 7.0 $\pm$ 1.6 & 52.5 $\pm$ 8.4\% & & 28.8 $\pm$ 1.7 & 47.7 $\pm$ 1.8\% & & 126.0 $\pm$ 6.8 & 63.6 $\pm$ 3.4\% & \\
 & sparse70         & 0.8 $\pm$ 0.5 & 6.0 $\pm$ 4.0\% & & 0.0 $\pm$ 0.0 & 0.0 $\pm$ 0.0\% & & 1.0 $\pm$ 1.0 & 0.5 $\pm$ 0.5\% & \\
\bottomrule
\end{tabular}
}
\caption{Dense-correct generations retained by the pruned model. \textit{$n$ kept}: correct for both models; \textit{kept}: fraction of dense-correct generations also solved by the pruned model. Values are mean $\pm$ std.}
\label{tab:acc-given-dense-correct}
\end{table*}

\begin{table*}[t]
\centering
\footnotesize
\setlength{\tabcolsep}{2.5pt}
\resizebox{0.9\textwidth}{!}{%
\begin{tabular}{ll rrrr rrrr rrrr}
\toprule
 & & \multicolumn{4}{c}{AIME25} & \multicolumn{4}{c}{LCB} & \multicolumn{4}{c}{MATH-500 L4--5} \\
\cmidrule(lr){3-6}\cmidrule(lr){7-10}\cmidrule(lr){11-14}
family & level & (a) & (b) & (c) & kept $\mid$ term. & (a) & (b) & (c) & kept $\mid$ term. & (a) & (b) & (c) & kept $\mid$ term. \\
\midrule
\multirow{3}{2.0cm}{DeepSeek-R1-Distill-Qwen-1.5B (OBC)}
 & sparse10         & 19.5 (72) & 2.5 (9) & 5.0 (19) & 79.7 $\pm$ 5.2\% & 107.5 (81) & 4.5 (3) & 21.0 (16) & 83.7 $\pm$ 0.5\% & 175.0 (89) & 5.5 (3) & 16.5 (8) & 91.4 $\pm$ 0.5\% \\
 & sparse50         & 15.0 (56) & 6.5 (24) & 5.5 (20) & 73.1 $\pm$ 4.4\% & 51.0 (38) & 50.5 (38) & 31.5 (24) & 61.8 $\pm$ 1.4\% & 133.5 (68) & 38.0 (19) & 25.5 (13) & 83.9 $\pm$ 3.3\% \\
 & sparse60         & 1.5 (6) & 22.5 (83) & 3.0 (11) & \textit{35.0 $\pm$ 21.2\%} & 0.0 (0) & 132.5 (100) & 0.5 (0) & \textit{0.0\%} & 39.5 (20) & 142.5 (72) & 15.0 (8) & 72.0 $\pm$ 8.5\% \\
\midrule
\multirow{4}{2.0cm}{DeepSeek-R1-Distill-Qwen-7B (OBC)}
 & sparse10         & 33.0 (71) & 4.0 (9) & 10.0 (21) & 77.5 $\pm$ 9.5\% & 198.0 (89) & 4.5 (2) & 21.0 (9) & 90.4 $\pm$ 1.3\% & 208.0 (93) & 3.0 (1) & 12.0 (5) & 94.5 $\pm$ 1.3\% \\
 & sparse50         & 30.0 (64) & 8.5 (18) & 8.5 (18) & 78.1 $\pm$ 3.5\% & 145.0 (65) & 25.5 (11) & 53.0 (24) & 73.2 $\pm$ 1.0\% & 204.5 (92) & 8.5 (4) & 10.0 (4) & 95.3 $\pm$ 0.6\% \\
 & sparse60         & 21.0 (44) & 13.0 (28) & 13.0 (28) & 61.6 $\pm$ 2.2\% & 55.5 (25) & 108.0 (48) & 60.0 (27) & 48.1 $\pm$ 0.3\% & 173.5 (78) & 37.5 (17) & 12.0 (5) & 93.5 $\pm$ 0.7\% \\
 & sparse70         & 6.0 (13) & 34.5 (73) & 6.5 (14) & \textit{48.1 $\pm$ 2.7\%} & 0.0 (0) & 225.0 (100) & 0.0 (0) & -- & 57.0 (26) & 150.5 (67) & 15.5 (7) & 78.6 $\pm$ 2.7\% \\
\midrule
\multirow{4}{2.0cm}{DeepSeek-R1-Distill-Qwen-14B (OBC)}
 & sparse10         & 47.5 (74) & 5.5 (9) & 11.0 (17) & 81.2 $\pm$ 1.7\% & 254.0 (91) & 5.0 (2) & 19.0 (7) & 93.0 $\pm$ 1.5\% & 220.0 (95) & 1.5 (1) & 10.0 (4) & 95.6 $\pm$ 0.7\% \\
 & sparse50         & 37.5 (59) & 11.5 (18) & 15.0 (23) & 71.4 $\pm$ 0.4\% & 225.0 (81) & 22.0 (8) & 31.0 (11) & 87.9 $\pm$ 2.0\% & 206.5 (89) & 5.5 (2) & 19.5 (8) & 91.4 $\pm$ 0.8\% \\
 & sparse60         & 29.5 (46) & 24.0 (37) & 10.5 (16) & 73.8 $\pm$ 4.4\% & 140.0 (50) & 110.0 (40) & 28.0 (10) & 83.3 $\pm$ 2.9\% & 172.0 (74) & 38.0 (16) & 21.5 (9) & 88.9 $\pm$ 1.4\% \\
 & sparse70         & 3.0 (5) & 53.5 (84) & 7.5 (12) & \textit{27.9 $\pm$ 4.1\%} & 0.0 (0) & 278.0 (100) & 0.0 (0) & -- & 40.5 (17) & 172.5 (75) & 18.5 (8) & 68.6 $\pm$ 2.0\% \\
\midrule
\multirow{5}{2.0cm}{DeepSeek-R1-Distill-Llama-8B (OBC)}
 & sparse10         & 7.8 (77) & 0.8 (6) & 1.8 (17) & 81.4 $\pm$ 12.6\% & 49.8 (89) & 1.2 (2) & 4.8 (8) & 91.3 $\pm$ 5.0\% & 178.0 (92) & 7.0 (4) & 8.0 (4) & 95.7 $\pm$ 1.5\% \\
 & sparse40         & 7.2 (73) & 0.5 (4) & 2.5 (22) & 76.1 $\pm$ 19.0\% & 47.0 (84) & 1.8 (3) & 7.0 (13) & 87.1 $\pm$ 1.2\% & 155.0 (80) & 28.0 (15) & 10.0 (5) & 93.9 $\pm$ 1.9\% \\
 & sparse50         & 5.5 (54) & 0.8 (6) & 4.0 (39) & 57.5 $\pm$ 15.0\% & 38.8 (70) & 7.2 (13) & 9.8 (17) & 80.2 $\pm$ 5.4\% & 119.0 (62) & 63.0 (33) & 11.0 (6) & 91.5 $\pm$ 2.4\% \\
 & sparse60         & 1.0 (10) & 6.2 (60) & 3.0 (30) & \textit{28.8 $\pm$ 21.7\%} & 10.2 (18) & 33.0 (59) & 12.5 (22) & 44.9 $\pm$ 4.7\% & 31.0 (16) & 150.0 (78) & 12.0 (6) & 72.1 $\pm$ 6.8\% \\
 & sparse70         & 0.0 (0) & 10.2 (100) & 0.0 (0) & -- & 0.0 (0) & 55.8 (100) & 0.0 (0) & -- & 0.0 (0) & 193.0 (100) & 0.0 (0) & -- \\
\midrule
\multirow{4}{2.0cm}{OpenThinker3-1.5B (OBC)}
 & sparse10         & 7.8 (79) & 1.0 (10) & 1.0 (10) & 88.5 $\pm$ 9.1\% & 35.2 (86) & 2.5 (6) & 3.2 (8) & 91.6 $\pm$ 3.8\% & 172.0 (86) & 21.0 (11) & 6.0 (3) & 96.6 $\pm$ 1.4\% \\
 & sparse40         & 6.2 (64) & 3.0 (31) & 0.5 (5) & \textit{92.9 $\pm$ 14.3\%} & 25.5 (62) & 7.8 (19) & 7.8 (19) & 76.6 $\pm$ 5.8\% & 139.0 (70) & 51.0 (26) & 9.0 (5) & 93.9 $\pm$ 2.0\% \\
 & sparse50         & 5.2 (54) & 3.2 (34) & 1.2 (12) & \textit{82.1 $\pm$ 13.7\%} & 8.5 (21) & 21.8 (53) & 10.8 (26) & 44.4 $\pm$ 2.6\% & 69.0 (35) & 122.0 (61) & 8.0 (4) & 89.6 $\pm$ 3.5\% \\
 & sparse60         & 0.2 (3) & 8.5 (87) & 1.0 (10) & \textit{12.5 $\pm$ 25.0\%} & 0.0 (0) & 40.5 (99) & 0.5 (1) & \textit{0.0 $\pm$ 0.0\%} & 7.0 (4) & 189.0 (95) & 3.0 (2) & \textit{70.0 $\pm$ 14.5\%} \\
\midrule
\multirow{5}{2.0cm}{OpenThinker3-7B (OBC)}
 & sparse10         & 11.5 (87) & 1.0 (7) & 0.8 (6) & 93.9 $\pm$ 4.2\% & 53.8 (89) & 3.0 (5) & 3.5 (6) & 93.9 $\pm$ 3.5\% & 177.0 (89) & 3.0 (2) & 18.0 (9) & 90.8 $\pm$ 2.1\% \\
 & sparse40         & 11.0 (83) & 2.0 (16) & 0.2 (2) & 98.2 $\pm$ 3.6\% & 53.2 (88) & 3.8 (6) & 3.2 (5) & 94.3 $\pm$ 2.4\% & 163.0 (82) & 11.0 (6) & 24.0 (12) & 87.2 $\pm$ 2.4\% \\
 & sparse50         & 9.8 (74) & 2.0 (15) & 1.5 (10) & 88.0 $\pm$ 13.9\% & 46.0 (76) & 8.8 (15) & 5.5 (9) & 89.4 $\pm$ 3.2\% & 157.0 (79) & 26.0 (13) & 15.0 (8) & 91.3 $\pm$ 2.2\% \\
 & sparse60         & 7.0 (52) & 3.8 (28) & 2.5 (20) & 74.4 $\pm$ 13.2\% & 28.8 (48) & 15.2 (25) & 16.2 (27) & 63.9 $\pm$ 2.1\% & 126.0 (64) & 59.0 (30) & 13.0 (7) & 90.6 $\pm$ 2.5\% \\
 & sparse70         & 0.8 (6) & 10.5 (79) & 2.0 (15) & \textit{43.8 $\pm$ 42.7\%} & 0.0 (0) & 59.8 (99) & 0.5 (1) & \textit{0.0 $\pm$ 0.0\%} & 1.0 (1) & 196.0 (99) & 1.0 (1) & \textit{50.0 $\pm$ 35.4\%} \\
\bottomrule
\end{tabular}}
\caption{Breakdown of dense-correct generations under pruning. Each generation is (a) \textit{correct}, (b) \textit{reasoning divergence} (no \texttt{</think>}), or (c) \textit{terminated-wrong} (\texttt{</think>} emitted but the final answer is incorrect). Values show mean count per run with the corresponding percentage in parentheses. \textit{Kept $\mid$ term.} denotes accuracy among terminated generations. Italics indicate fewer than 30 terminated dense-correct items.}
\label{tab:acc-given-dense-correct-breakdown}
\end{table*}


To separate loss of reasoning capability from loss of termination, we restrict evaluation to generations that the dense model answers correctly and track what happens to them after pruning. Table~\ref{tab:acc-given-dense-correct} measures how many of these generations the pruned model still solves.
Table~\ref{tab:acc-given-dense-correct-breakdown} then asks why the remaining generations fail, classifying each as correct, non-termination, or terminated-wrong, and reports accuracy among terminated generations.


As sparsity increases, pruned models retain fewer problems solved by their dense counterparts, detailed in Table~\ref{tab:acc-given-dense-correct}, primarily due to increased non-termination rather than incorrect terminated generations, detailed in Table~\ref{tab:acc-given-dense-correct-breakdown}. For DeepSeek-R1-Distill-Qwen-7B on AIME-25, non-termination rises from 9\% at 10\% sparsity to 73\% at 70\%, while terminated-wrong remains between 14\% and 28\%. This pattern generalizes across models and benchmarks with non-termination exceeding terminated-wrong rates in 10 of 12 pairs at 60\% sparsity. Moreover, conditional on termination, larger models retain relatively high accuracy. For example, DeepSeek-R1-Distill-Qwen-14B achieves 73.8\% and 83.3\%, while OpenThinker3-7B achieves 74.4\% and 63.9\% on AIME-25 and LiveCodeBench, respectively. These results suggest that pruning disproportionately impairs termination behavior, while the ability to reason correctly remains comparatively robust when generation terminates.

\subsection{Comprehensive Evaluation of Larger Budget Generation}
\label{app:larger_budget}
Increasing the generation limit from 16,384 to 65,536 tokens provides limited benefit for non-terminating traces, detailed in Table \ref{tab:budget}. While the extended budget substantially reduces non-termination for dense models, $20/24$ of the 7B and $15/15$ of the 14B non-terminating traces terminate under the larger budget, this yields only $0$ and $2$ additional correct answers, respectively. Recovery becomes increasingly rare under stronger pruning: for 60\% pruned by OBC, only $10/68$ of 7B model and $4/73$ of 14B model traces are recovered, while 70\% pruned recovers none. This suggests that insufficient generation length is not the primary cause of failure; rather, pruning induces a persistent tendency to continue generating instead of committing to a final answer.

\begin{table*}[t]
\centering
\scriptsize
\resizebox{0.6\textwidth}{!}{%
\begin{tabular}{@{}lllrrrrr@{}}
\toprule
\thead{Model} & \thead{Level} & \thead{Token\\Limit} & \thead{Non-termination\\Rate} & \thead{Accuracy} & \thead{Accuracy $\mid$\\Terminated} & \thead{Recovered\\Traces} & \thead{Recovered\\Correct} \\
\midrule
7B & dense & 16,384 & 20.0\% & 42.5\% & 53.1\% & --- & --- \\
7B & dense & 65,536 & 3.3\% & 42.5\% & 44.0\% & 20 / 24 (83.3\%) & 0 \\
7B & \ckpt{10} & 16,384 & 20.0\% & 31.7\% & 39.6\% & --- & --- \\
7B & \ckpt{10} & 65,536 & 2.5\% & 31.7\% & 32.5\% & 21 / 24 (87.5\%) & 0 \\
7B & \ckpt{50} & 16,384 & 42.5\% & 30.0\% & 52.2\% & --- & --- \\
7B & \ckpt{50} & 65,536 & 12.5\% & 30.8\% & 35.2\% & 36 / 51 (70.6\%) & 2 \\
7B & \ckpt{60} & 16,384 & 56.7\% & 22.5\% & 51.9\% & --- & --- \\
7B & \ckpt{60} & 65,536 & 48.3\% & 22.5\% & 43.5\% & 10 / 68 (14.7\%) & 0 \\
7B & \ckpt{70} & 16,384 & 80.8\% & 5.0\% & 26.1\% & --- & --- \\
7B & \ckpt{70} & 65,536 & 80.8\% & 5.0\% & 26.1\% & 0 / 97 (0.0\%) & 0 \\
\midrule
14B & dense & 16,384 & 12.5\% & 54.2\% & 61.9\% & --- & --- \\
14B & dense & 65,536 & 0.0\% & 55.8\% & 55.8\% & 15 / 15 (100.0\%) & 2 \\
14B & \ckpt{10} & 16,384 & 15.8\% & 47.5\% & 56.4\% & --- & --- \\
14B & \ckpt{10} & 65,536 & 0.0\% & 49.2\% & 49.2\% & 19 / 19 (100.0\%) & 2 \\
14B & \ckpt{50} & 16,384 & 36.7\% & 35.8\% & 56.6\% & --- & --- \\
14B & \ckpt{50} & 65,536 & 10.8\% & 38.3\% & 43.0\% & 31 / 44 (70.5\%) & 3 \\
14B & \ckpt{60} & 16,384 & 60.8\% & 26.7\% & 68.1\% & --- & --- \\
14B & \ckpt{60} & 65,536 & 57.5\% & 27.5\% & 64.7\% & 4 / 73 (5.5\%) & 1 \\
14B & \ckpt{70} & 16,384 & 89.2\% & 3.3\% & 30.8\% & --- & --- \\
14B & \ckpt{70} & 65,536 & 89.2\% & 3.3\% & 30.8\% & 0 / 107 (0.0\%) & 0 \\
\bottomrule
\end{tabular}}
\caption{Effect of increasing the generation budget from 16,384 to 65,536 tokens on the DeepSeek-R1-Distill-Qwen 7B and 14B models on AIME-25. A 65,536 row continues each generation non-termination at 16,384 from its own tokens; \emph{recovered} = non-termination at 16,384 and terminated by 65,536.}
\label{tab:budget}
\end{table*}

\subsection{Comprehensive Evaluation of the Stop-Token Mechanism}
\label{app:stopping_mechanism}

We evaluate the pruned checkpoints using forward passes with teacher forcing along the dense model's trajectories. We focus on the $\mathcal{D}_{\mathrm{cov}}$ set, where the dense models terminate and answer correctly. Each trace is replayed under its original generation prompt. We focus on two position sets: \textbf{STOP}, the position where the dense model emits \ethink{}, and \textbf{OTHER}, positions more than 50 tokens before STOP. At each position, we measure the probability and rank assigned to \ethink{}. We define $\Delta_{\mathrm{close}}$ as the log-probability of \ethink{} at STOP minus its mean log-probability over OTHER positions, and $\Delta_{\mathrm{stop}}$ as the log-probability of \ethink{} at STOP minus the mean log-probability of the token the dense model actually generated at OTHER positions. The full results are reported in Table \ref{tab:full_prob_think}.

Across models and tasks, the stopping signal remains highly localized at the dense model's STOP position. For the 7B and 14B models, the median rank of \ethink{} is 0 at 10\%--60\% pruning ratios, meaning that \ethink{} is consistently the top prediction at the correct stopping position. Even at 70\% sparsity, where this signal weakens, \ethink{} remains near the top: its median rank is 5.5 and 2 on AIME-25 for the 7B and 14B models, respectively. The same pattern holds on MATH-500, with median ranks of 7 and 3. At OTHER positions, in contrast, \ethink{} receives almost no probability at any checkpoint: its mean log-probability lies between $-23.3$ and $-30.4$ for the 7B and 14B models.

The log-probability of \ethink{} at STOP is substantially higher than at OTHER positions. For the 7B model, $\log p(\text{\ethink{}})$ at STOP rises from $-4.83$ at 70\% pruning ratio to near zero at pruned checkpoints 10\%--60\%, while remaining between $-23$ and $-28$ at OTHER positions. Consequently, $\Delta_{\mathrm{close}}$ remains at least $+22$ for both 7B and 14B models at 70\% sparsity. Meanwhile, $\Delta_{\mathrm{stop}}$ is positive below 70\% sparsity but drops to $-2.6$ to $-4.3$ at 70\%, indicating that aggressive pruning weakens the stopping signal relative to ordinary next-token prediction. This pattern holds across AIME-25, MATH-500, and both model sizes, suggesting that pruning preserves a localized stopping signal even at high sparsity.

\begin{table*}[t]
\begin{center}
\scriptsize
\setlength{\tabcolsep}{3.5pt}
\resizebox{\textwidth}{!}{%
\begin{tabular}{ll rrrrrr rrrrrr}
\toprule
 & & \multicolumn{6}{c}{AIME-25} & \multicolumn{6}{c}{MATH-500 L4--5} \\
\cmidrule(lr){3-8}\cmidrule(lr){9-14}
\thd{Model} & \thd{Ckpt} & \thd{p(\ethink{})} & \thd{Med.\ rank} & \thd{logp STOP} & \thd{\makecell{logp \ethink{}\\OTHER}} & \thd{$\Delta_{\text{close}}$} & \thd{$\Delta_{\text{stop}}$} & \thd{p(\ethink{})} & \thd{Med.\ rank} & \thd{logp STOP} & \thd{\makecell{logp \ethink{}\\OTHER}} & \thd{$\Delta_{\text{close}}$} & \thd{$\Delta_{\text{stop}}$} \\
\midrule
\multirow{4}{*}{\makecell{Deepseek-R1-Distill\\-Qwen-1.5B}}
 & \ckpt{10} & \cellcolor{keybg}\textit{0.9999} & 0 & \num{-0.000} & \num{-22.71} & \num{+22.71} & \num{+0.377} & \cellcolor{keybg}0.9999 & 0 & \num{-0.000} & \num{-23.11} & \num{+23.11} & \num{+0.374} \\
 & \ckpt{50} & \cellcolor{keybg}\textit{0.9804} & 0 & \num{-0.020} & \num{-27.28} & \num{+27.26} & \num{+0.400} & \cellcolor{keybg}0.9918 & 0 & \num{-0.008} & \num{-26.51} & \num{+26.50} & \num{+0.428} \\
 & \ckpt{60} & \cellcolor{keybg}\textit{0.7273} & 0 & \num{-0.379} & \num{-26.56} & \num{+26.18} & \num{+0.134} & \cellcolor{keybg}0.6301 & 0 & \num{-0.771} & \num{-26.06} & \num{+25.29} & \num{-0.226} \\
 & \ckpt{70} & \cellcolor{keybg}\textit{0.0351} & 44 & \num{-9.192} & \num{-24.19} & \num{+14.99} & \num{-8.431} & \cellcolor{keybg}0.0154 & 157.5 & \num{-11.076} & \num{-23.84} & \num{+12.76} & \num{-10.229} \\
\midrule
\multirow{4}{*}{\makecell{Deepseek-R1-Distill\\-Qwen-7B}}
 & \ckpt{10} & \cellcolor{keybg}\textit{1.0000} & 0 & \num{-0.000} & \num{-23.30} & \num{+23.30} & \num{+0.361} & \cellcolor{keybg}0.9999 & 0 & \num{-0.000} & \num{-24.15} & \num{+24.15} & \num{+0.349} \\
 & \ckpt{50} & \cellcolor{keybg}\textit{0.9978} & 0 & \num{-0.002} & \num{-25.23} & \num{+25.23} & \num{+0.374} & \cellcolor{keybg}0.9963 & 0 & \num{-0.004} & \num{-25.50} & \num{+25.50} & \num{+0.375} \\
 & \ckpt{60} & \cellcolor{keybg}\textit{0.9570} & 0 & \num{-0.045} & \num{-27.75} & \num{+27.71} & \num{+0.378} & \cellcolor{keybg}0.9240 & 0 & \num{-0.097} & \num{-27.03} & \num{+26.93} & \num{+0.345} \\
 & \ckpt{70} & \cellcolor{keybg}\textit{0.0360} & 5.5 & \num{-4.826} & \num{-27.43} & \num{+22.61} & \num{-4.274} & \cellcolor{keybg}0.0400 & 7 & \num{-4.747} & \num{-26.75} & \num{+22.00} & \num{-4.133} \\
\midrule
\multirow{4}{*}{\makecell{Deepseek-R1-Distill\\-Qwen-14B}}
 & \ckpt{10} & \cellcolor{keybg}\textit{1.0000} & 0 & \num{-0.000} & \num{-24.68} & \num{+24.68} & \num{+0.353} & \cellcolor{keybg}1.0000 & 0 & \num{-0.000} & \num{-25.47} & \num{+25.47} & \num{+0.340} \\
 & \ckpt{50} & \cellcolor{keybg}\textit{0.9945} & 0 & \num{-0.006} & \num{-28.45} & \num{+28.44} & \num{+0.365} & \cellcolor{keybg}0.9976 & 0 & \num{-0.002} & \num{-27.61} & \num{+27.61} & \num{+0.376} \\
 & \ckpt{60} & \cellcolor{keybg}\textit{0.8539} & 0 & \num{-0.223} & \num{-30.35} & \num{+30.13} & \num{+0.212} & \cellcolor{keybg}0.9069 & 0 & \num{-0.130} & \num{-28.96} & \num{+28.83} & \num{+0.331} \\
 & \ckpt{70} & \cellcolor{keybg}\textit{0.0725} & 2 & \num{-3.748} & \num{-28.06} & \num{+24.31} & \num{-3.158} & \cellcolor{keybg}0.1304 & 3 & \num{-3.298} & \num{-26.95} & \num{+23.65} & \num{-2.642} \\
\midrule
\multirow{4}{*}{\makecell{OpenThinker3\\-7B}}
 & \ckpt{10} & \cellcolor{keybg}\textit{1.0000} & 0 & \num{-0.000} & \num{-31.12} & \num{+31.12} & \num{+0.319} & \cellcolor{keybg}0.9972 & 0 & \num{-0.004} & \num{-31.89} & \num{+31.89} & \num{+0.291} \\
 & \ckpt{50} & \cellcolor{keybg}\textit{0.9994} & 0 & \num{-0.001} & \num{-29.71} & \num{+29.71} & \num{+0.393} & \cellcolor{keybg}0.9924 & 0 & \num{-0.021} & \num{-30.15} & \num{+30.13} & \num{+0.368} \\
 & \ckpt{60} & \cellcolor{keybg}\textit{0.9978} & 0 & \num{-0.002} & \num{-28.81} & \num{+28.80} & \num{+0.467} & \cellcolor{keybg}0.9826 & 0 & \num{-0.039} & \num{-29.12} & \num{+29.08} & \num{+0.442} \\
 & \ckpt{70} & \cellcolor{keybg}\textit{0.8391} & 0 & \num{-0.279} & \num{-26.96} & \num{+26.68} & \num{+0.371} & \cellcolor{keybg}0.5784 & 0 & \num{-0.912} & \num{-27.05} & \num{+26.14} & \num{-0.210} \\
\bottomrule
\end{tabular}
}
\end{center}
\caption{Teacher-forced evaluation of the stopping signal at the dense model's \ethink{} position. STOP denotes the position whose next token is the dense \ethink{}, while OTHER includes all trace positions more than 50 tokens before STOP. Logp STOP is the mean log-probability of \ethink{} at STOP. Logp \ethink{} OTHER is the mean log-probability of \ethink{} over OTHER positions, and $\Delta_{\text{close}}$ = logp STOP $-$ logp \ethink{} OTHER. $\Delta_{\text{stop}}$ = logp STOP $-$ the mean log-probability of the token the dense model actually generated at OTHER positions. Rank 0 indicates that \ethink{} is the model's top prediction at STOP. All quantities are computed by teacher-forcing dense-model traces through each checkpoint.}
\label{tab:full_prob_think}
\end{table*}

\begin{table*}[t]
\begin{center}
\tabsmall
\setlength{\tabcolsep}{3.5pt}
\resizebox{0.9\textwidth}{!}{%
\begin{tabular}{ll rrrrr rrrrr rrrrr}
\toprule
 & & \multicolumn{5}{c}{AIME-25} & \multicolumn{5}{c}{LCB} & \multicolumn{5}{c}{MATH-500 L4--5} \\
\cmidrule(lr){3-7}\cmidrule(lr){8-12}\cmidrule(lr){13-17}
\thd{Model} & \thd{Ckpt} & \thd{$n_{\mathrm{dec}}$} & \thd{$n_{\mathrm{ctrl}}$} & \thd{Dec.} & \thd{Ctrl.} & \thd{Ent.} & \thd{$n_{\mathrm{dec}}$} & \thd{$n_{\mathrm{ctrl}}$} & \thd{Dec.} & \thd{Ctrl.} & \thd{Ent.} & \thd{$n_{\mathrm{dec}}$} & \thd{$n_{\mathrm{ctrl}}$} & \thd{Dec.} & \thd{Ctrl.} & \thd{Ent.} \\
\midrule
\multirow{5}{*}{DS-Qwen 1.5B}
  & \dns      & 796 & 200 & \num{-13.0} & \num{-13.2} & 1.60 & 720 & 200 & \num{-16.7} & \num{-15.5} & 1.46 & 684 & 200 & \num{-14.9} & \num{-21.9} & 1.55 \\
  & \ckpt{10} & 794 & 200 & \num{-13.4} & \num{-13.4} & 1.62 & 734 & 200 & \num{-16.0} & \num{-12.5} & 1.44 & 690 & 200 & \num{-15.3} & \num{-23.3} & 1.57 \\
  & \ckpt{50} & 786 & 194 & \num{-8.4}  & \num{-11.7} & 1.32 & 788 & 192 & \num{-8.2}  & \num{-5.3}  & 1.48 & 686 & 200 & \num{-10.2} & \num{-15.3} & 1.35 \\
  & \ckpt{60} & 762 & 182 & \num{-3.8}  & \num{-4.5}  & 1.33 & 756 & 172 & \num{-2.7}  & \num{-3.2}  & 1.33 & 762 & 198 & \num{-5.6}  & \num{-6.8}  & 1.32 \\
  & \ckpt{70} & 524 & 154 & \num{1.6}   & \num{-1.2}  & 1.48 & 556 & 158 & \num{2.2}   & \num{-0.9}  & 1.60 & 506 & 160 & \num{1.0}   & \num{-0.8}  & 1.54 \\
\midrule
\multirow{5}{*}{DS-Qwen 7B}
  & \dns      & 794 & 200 & \num{-15.3} & \num{-16.6} & 1.55 & 692 & 200 & \num{-19.5} & \num{-17.0} & 1.33 & 624 & 200 & \num{-18.2} & \num{-24.4} & 1.49 \\
  & \ckpt{10} & 792 & 200 & \num{-14.2} & \num{-15.6} & 1.59 & 708 & 200 & \num{-18.8} & \num{-14.7} & 1.33 & 638 & 200 & \num{-18.2} & \num{-24.7} & 1.48 \\
  & \ckpt{50} & 788 & 200 & \num{-11.9} & \num{-15.0} & 1.51 & 742 & 200 & \num{-14.8} & \num{-11.8} & 1.37 & 680 & 200 & \num{-15.3} & \num{-21.0} & 1.52 \\
  & \ckpt{60} & 796 & 200 & \num{-9.6}  & \num{-10.1} & 1.41 & 794 & 200 & \num{-10.4} & \num{-6.5}  & 1.45 & 710 & 200 & \num{-12.0} & \num{-14.9} & 1.42 \\
  & \ckpt{70} & 782 & 188 & \num{-5.0}  & \num{-4.6}  & 1.44 & 742 & 182 & \num{-4.1}  & \num{-1.5}  & 1.33 & 782 & 194 & \num{-6.7}  & \num{-4.6}  & 1.45 \\
\midrule
\multirow{5}{*}{DS-Qwen 14B}
  & \dns      & 790 & 200 & \num{-13.8} & \num{-16.0} & 1.54 & 680 & 200 & \num{-17.1} & \num{-18.1} & 1.33 & 622 & 200 & \num{-16.1} & \num{-23.7} & 1.41 \\
  & \ckpt{10} & 788 & 200 & \num{-13.2} & \num{-15.8} & 1.49 & 670 & 200 & \num{-17.5} & \num{-17.6} & 1.26 & 630 & 200 & \num{-16.0} & \num{-23.2} & 1.47 \\
  & \ckpt{50} & 768 & 198 & \num{-12.4} & \num{-15.9} & 1.40 & 680 & 200 & \num{-13.5} & \num{-13.5} & 1.30 & 650 & 200 & \num{-14.7} & \num{-21.6} & 1.34 \\
  & \ckpt{60} & 782 & 196 & \num{-10.7} & \num{-10.5} & 1.23 & 710 & 199 & \num{-9.2}  & \num{-8.7}  & 1.33 & 690 & 196 & \num{-12.7} & \num{-12.5} & 1.28 \\
  & \ckpt{70} & 762 & 180 & \num{-4.7}  & \num{-4.3}  & 1.24 & 720 & 174 & \num{-3.2}  & \num{-2.7}  & 1.39 & 782 & 196 & \num{-5.9}  & \num{-5.1}  & 1.32 \\
\midrule
\multirow{6}{*}{OT3-7B}
  & \dns      & 800 & 200 & \num{-23.1} & \num{-23.8} & 1.03 & 704 & 198 & \num{-28.7} & \num{-21.6} & 1.17 & 710 & 200 & \num{-29.3} & \num{-37.0} & 0.88 \\
  & \ckpt{10} & 796 & 200 & \num{-22.6} & \num{-24.3} & 1.09 & 716 & 200 & \num{-26.4} & \num{-22.7} & 1.17 & 724 & 200 & \num{-29.5} & \num{-30.6} & 0.88 \\
  & \ckpt{40} & 800 & 200 & \num{-19.6} & \num{-18.4} & 1.14 & 736 & 200 & \num{-21.6} & \num{-20.1} & 1.30 & 730 & 200 & \num{-25.5} & \num{-27.2} & 0.96 \\
  & \ckpt{50} & 800 & 200 & \num{-15.8} & \num{-17.4} & 1.22 & 772 & 200 & \num{-21.9} & \num{-15.3} & 1.33 & 748 & 200 & \num{-20.1} & \num{-25.8} & 0.99 \\
  & \ckpt{60} & 796 & 200 & \num{-13.2} & \num{-12.0} & 1.33 & 794 & 200 & \num{-13.1} & \num{-10.5} & 1.51 & 778 & 196 & \num{-14.6} & \num{-17.0} & 1.11 \\
  & \ckpt{70} & 790 & 194 & \num{-5.6}  & \num{-5.6}  & 1.60 & 696 & 192 & \num{-1.4}  & \num{-3.6}  & 1.81 & 538 & 120 & \num{-6.7}  & \num{-12.3} & 1.47 \\
\bottomrule
    \end{tabular}}
\end{center}
\caption{Log-probability ratio between cyclic and progressive continuations on AIME-25, LCB and MATH-500. $\Delta_{\mathrm{cyc}}=\log P(\mathrm{CYCLIC})-\log P(\mathrm{PROGRESS})$, scored over the next 32 tokens. \thd{Dec.} is measured at decision points, \thd{Ctrl.} at randomly sampled paragraph in the same traces; \thd{Ent.} is the predictive entropy at the decision points. $n_{\mathrm{dec}}$ is the number of decision points and $n_{\mathrm{ctrl}}$ the number of control points. More negative means the model prefers the progressive continuation more strongly.}
\label{tab:full_transition}
\end{table*}

\begin{table*}[t]
\begin{center}
\tabsmall
\setlength{\tabcolsep}{4pt}
\resizebox{0.7\textwidth}{!}{%
\begin{tabular}{ll rrr rrr}
\toprule
 & & \multicolumn{3}{c}{MATH-500 L4--5} & \multicolumn{3}{c}{AIME-25} \\
\cmidrule(lr){3-5}\cmidrule(lr){6-8}
\thd{Model} & \thd{Ckpt} & \thd{$P$(cyc $\mid$ cyc)} & \thd{$P$(cyc $\mid$ prog)} & \thd{Count} & \thd{$P$(cyc $\mid$ cyc)} & \thd{$P$(cyc $\mid$ prog)} & \thd{Count} \\
\midrule
\multirow{5}{*}{DeepSeek-R1-Distill-Qwen 7B}
  & \dns      & 0.157 & 0.025 & \num{4094}   & 0.029 & 0.026 & \num{1296} \\
  & \ckpt{40} & 0.206 & 0.027 & \num{155830} & 0.230 & 0.031 & \num{38947} \\
  & \ckpt{50} & 0.257 & 0.047 & \num{207714} & 0.283 & 0.043 & \num{51479} \\
  & \ckpt{60} & 0.325 & 0.086 & \num{434020} & 0.327 & 0.064 & \num{85675} \\
  & \ckpt{70} & 0.307 & 0.105 & \num{596073} & 0.307 & 0.085 & \num{113353} \\
\midrule
\multirow{5}{*}{DeepSeek-R1-Distill-Qwen 14B}
  & \dns      & 0.062 & 0.017 & \num{2714}   & 0.000 & 0.029 & \num{770} \\
  & \ckpt{40} & 0.142 & 0.028 & \num{122288} & 0.164 & 0.025 & \num{30508} \\
  & \ckpt{50} & 0.212 & 0.042 & \num{180908} & 0.203 & 0.033 & \num{47349} \\
  & \ckpt{60} & 0.326 & 0.076 & \num{283318} & 0.271 & 0.059 & \num{68347} \\
  & \ckpt{70} & 0.307 & 0.097 & \num{514252} & 0.324 & 0.081 & \num{136784} \\
\midrule
\multirow{5}{*}{OpenThinker3-7B}
  & \dns      & 0.137 & 0.026 & \num{6227}    & 0.122 & 0.015 & \num{3057} \\
  & \ckpt{40} & 0.138 & 0.018 & \num{306167}  & 0.142 & 0.018 & \num{105101} \\
  & \ckpt{50} & 0.160 & 0.025 & \num{369215}  & 0.146 & 0.022 & \num{127031} \\
  & \ckpt{60} & 0.160 & 0.032 & \num{579485}  & 0.151 & 0.029 & \num{192809} \\
  & \ckpt{70} & 0.210 & 0.060 & \num{1250119} & 0.167 & 0.049 & \num{479539} \\
\bottomrule
\end{tabular}
}
\end{center}
\caption{Transition dynamics between cyclic and progressive states on MATH-500 Levels 4--5 and AIME-25, measured on continuations from a \emph{shared} dense branch point. $P(\mathrm{cyc}\mid\mathrm{cyc})$ and $P(\mathrm{cyc}\mid\mathrm{prog})$ measure loop persistence and loop entry over adjacent paragraph pairs; \thd{Count} is the number of such pairs.}
\label{tab:exp5-branch-transitions-paragraph}
\end{table*}



\subsection{Comprehensive Evaluation of Transition of Pruned LRMs}
\label{app:full_transition_pruned}

We investigate whether pruning alters reasoning trajectories at moments of reconsideration, where a model may either advance its solution or revisit previously explored reasoning. We define a \emph{Decision point} as the start of a paragraph beginning with one of 11 reconsideration cues: \emph{wait}, \emph{but wait}, \emph{hmm}, \emph{alternatively}, \emph{actually}, \emph{hold on}, or \emph{let me check/verify/double-check/reconsider/re-check}. Cues are matched case-insensitively at the beginning of \texttt{\textbackslash n\textbackslash n}-separated paragraphs, followed \citet{lotfi2026quantized}. The cue set is fixed beforehand using paragraph-initial bigram frequencies from dense-model traces and is not modified based on pruned generations. \emph{Control points} are randomly sampled paragraph starts without these cues. We also assign each paragraph a local trajectory label: \textsc{Cyclic} if the Jaccard similarity between its set of 8-word n-grams and that of any preceding paragraph in the same trace is at least $0.5$, and \textsc{Progress} otherwise.

\paragraph{Teacher-Forced Preferences.} We compute $\Delta_{\mathrm{cyc}}$ (Section~\ref{sec:observation}) at decision and control points to assess whether pruning increases the tendency toward \textsc{Cyclic} reasoning at reconsideration points and ordinary paragraph starts. As shown in Table \ref{tab:full_transition}, $\Delta_{\mathrm{cyc}}$ becomes progressively less negative as pruning increases across all model families and tasks, indicating a growing preference for cyclic over progressive continuations. This pattern also holds at control points, indicating that pruning broadly shifts continuation preferences rather than selectively affecting reconsideration cues. The effect is strongest at high sparsity, whereas predictive entropy changes more gradually, suggesting a shift in continuation behavior rather than a comparable increase in uncertainty.

\paragraph{Free-Running Continuations.} We examine how pruned models continue dense reasoning traces whose prefixes are sufficient to identify the correct answer. For each eligible trace, we teacher-force the prefix up to the branch point through the pruned checkpoint and sample $N$ continuations $c$.
 
Table~\ref{tab:exp5-branch-transitions-paragraph} shows consistent shifts toward cyclic reasoning across all models and benchmarks. Loop entry, $P(\mathrm{cyc}\mid\mathrm{prog})$, increases monotonically with sparsity, reaching $0.097$-$0.105$ on MATH-500 and $0.081$-$0.085$ on AIME-25 for DS-Qwen models at 70\% sparsity, roughly 3–6$\times$ their dense rates. Loop persistence, $P(\mathrm{cyc}\mid\mathrm{cyc})$, also rises, though it peaks at 60\% in three of four DS-Qwen blocks. OT3-7B exhibits a weaker but similar increase, particularly in persistence. These trends accompany a 3–5$\times$ increase in paragraph pairs from 40\% to 70\% sparsity, suggesting that pruning promotes both entry into and persistence within cyclic reasoning across model families and benchmarks.

\subsection{Comprehensive Commitment Directions Analysis}
\label{app:commitment-direction-analysis}

Section~\ref{sec:method_consistency} shows that termination directions constructed from different pruned checkpoints of the same model have high cosine similarity. However, since these checkpoints are pruned from the same dense model, some similarity is expected regardless of what the directions encode \citep{ding2021grounding}. We therefore use hypothesis testing with three increasingly strong baselines and a transfer test to assess whether the observed cosine similarity is significant and specifically associated with termination.

\paragraph{Setup.} We consider all pairs of sparsity levels $(A,B)$ within the same model for which both $\mathcal{C}^{+}$ and $\mathcal{C}^{-}$ contain sufficient continuations, yielding 15 checkpoint pairs across four models on MATH-500 and OpenR1-Math. All checkpoints share the same branch points, with 32 continuations sampled per point (64 for DS-Qwen-14B on OpenR1-Math), using temperature $0.6$ and a $16{,}384$-token budget. To make directions comparable, we use the same number $N$ of terminating and non-terminating continuations at every sparsity level of a model, and draw them from the same set of problems at every level. Thus, $N$ is determined by the scarcest class and ranges from 52 to 200. Let $\mathbf{r}^{A}_{\ell}$ denote the direction at layer $\ell$ for checkpoint $A$. We measure similarity using the layer-averaged cosine:
\begin{equation*}
\mathrm{sim}(A,B)=\frac{1}{L}\sum_{\ell=1}^{L}\cos\bigl(\mathbf{r}^{A}_{\ell},\mathbf{r}^{B}_{\ell}\bigr).
\end{equation*}
All baselines reuse the same continuation representations $\mathbf{s}_\ell(c)$ and layers, varying only the labels or held-out split. We next compare the cosine similarities against baseline directions:

\paragraph{Random directions.} Two independent random unit vectors in $d$ dimensions have an expected absolute cosine of $\sqrt{2/(\pi d)}$, at most $0.020$ for our models (Table~\ref{tab:random-cos}). All observed similarities substantially exceed this random baseline.

\begin{table*}[t]
\centering
\begin{minipage}[t]{0.34\textwidth}
\vspace{0pt}\centering
\scriptsize
\setlength{\tabcolsep}{4pt}
\resizebox{\linewidth}{!}{%
\begin{tabular}{l r r r}
\toprule
\thd{Model} & \thd{$d$} & \thd{Random} & \thd{Observed} \\
\midrule
DS-Qwen-1.5B & 1536 & 0.020 & 0.834--0.977 \\
DS-Qwen-7B & 3584 & 0.013 & 0.870--0.960 \\
OT3-7B & 3584 & 0.013 & 0.784 \\
DS-Qwen-14B & 5120 & 0.011 & 0.870--0.881 \\
\bottomrule
\end{tabular}}
\captionof{table}{Expected absolute cosine between two random unit vectors, $\sqrt{2/(\pi d)}$, for each model's hidden size $d$. \emph{Observed}: cosine similarity between termination directions; \emph{Random}: cosine similarity with random directions.}
\label{tab:random-cos}
\end{minipage}\hfill
\begin{minipage}[t]{0.63\textwidth}
\vspace{0pt}\centering
\scriptsize
\setlength{\tabcolsep}{3pt}
\resizebox{0.6\linewidth}{!}{%
\begin{tabular}{ll r r c r r}
\toprule
\thd{Model} & \thd{Levels} & \thd{$N$} & \thd{Sim.} & \thd{Shuffled} & \thd{$p$} & \thd{Split} \\
\midrule
\multicolumn{7}{l}{\textit{OpenR1-Math}} \\
DS-Qwen-1.5B & 40 / 50 & 200 & 0.977 & 0.40 / 0.77 & 0.005 & 0.80 \\
DS-Qwen-1.5B & 40 / 60 & 200 & 0.878 & 0.36 / 0.71 & 0.005 & 0.68 \\
DS-Qwen-1.5B & 50 / 60 & 200 & 0.925 & 0.42 / 0.78 & 0.005 & 0.75 \\
DS-Qwen-7B & 40 / 50 & 52 & 0.951 & 0.40 / 0.74 & 0.005 & 0.79 \\
DS-Qwen-7B & 40 / 60 & 52 & 0.877 & 0.35 / 0.69 & 0.005 & 0.69 \\
DS-Qwen-7B & 50 / 60 & 52 & 0.953 & 0.36 / 0.71 & 0.005 & 0.81 \\
DS-Qwen-14B & 50 / 60 & 61 & 0.881 & 0.32 / 0.68 & 0.005 & 0.69 \\
\midrule
\multicolumn{7}{l}{\textit{MATH-500}} \\
DS-Qwen-1.5B & 40 / 50 & 117 & 0.954 & 0.42 / 0.79 & 0.005 & 0.70 \\
DS-Qwen-1.5B & 40 / 60 & 117 & 0.834 & 0.37 / 0.74 & 0.005 & 0.58 \\
DS-Qwen-1.5B & 50 / 60 & 117 & 0.919 & 0.37 / 0.72 & 0.005 & 0.76 \\
DS-Qwen-7B & 50 / 60 & 55 & 0.960 & 0.43 / 0.82 & 0.005 & 0.59 \\
DS-Qwen-7B & 50 / 70 & 55 & 0.870 & 0.41 / 0.76 & 0.005 & 0.35 \\
DS-Qwen-7B & 60 / 70 & 55 & 0.903 & 0.37 / 0.73 & 0.005 & 0.68 \\
DS-Qwen-14B & 60 / 70 & 152 & 0.870 & 0.39 / 0.70 & 0.005 & 0.58 \\
OT3-7B & 60 / 70 & 126 & 0.784 & 0.42 / 0.74 & 0.030 & 0.48 \\
\bottomrule
\end{tabular}}
\captionof{table}{Cosine similarity between termination directions of two sparsity levels (\%) of the same model, against two baselines. \thd{$N$}: continuations per class; \thd{Sim.}: layer-averaged cosine similarity; \thd{Shuffled}: mean / 95th percentile of the similarity when the labels of one checkpoint are permuted (200 permutations), with permutation $p$-value \thd{$p$}; \thd{Split}: similarity of a contrast between two random halves of the branch points.}
\label{tab:cos-baselines}
\end{minipage}
\end{table*}

\paragraph{Label-shuffled directions.} We first test whether any difference of means over these continuations aligns across checkpoints, regardless of the labels. For each pair, we randomly permute the terminating and non-terminating labels of checkpoint $A$ while keeping the class sizes, recompute its direction at every layer, and measure its similarity with the true direction of $B$, repeated 200 times. As shown in Table~\ref{tab:cos-baselines}, shuffled directions already reach $0.32$--$0.43$ on average and $0.68$--$0.82$ at the 95th percentile, far above random directions. Nevertheless, the real similarity exceeds the shuffled baseline in all 15 pairs ($p\le 0.030$). The similarity therefore depends on the termination labels.

\paragraph{A shared contrast unrelated to termination.} 
The label-shuffled baseline does not rule out generic alignment between checkpoints pruned from the same model. As a control, we use branch-point identity, which all checkpoints encode since they share the same branch points, but which is unrelated to termination.

Concretely, we draw a random half $\mathcal{H}\subset\mathcal{B}$ of the branch points and relabel every continuation of $A$ and $B$ by whether its branch point lies in $\mathcal{H}$, ignoring whether it terminates. Using the same $\mathcal{H}$ for both checkpoints, we compute the difference-of-means direction of this contrast at every layer for each checkpoint and measure $\mathrm{sim}(A,B)$ between them. We repeat this for 100 random halves and report the average. If shared weights alone explained the alignment, this branch-point contrast would align as strongly as the termination direction. Instead, its similarity is only $0.35$--$0.81$, which is $0.143$--$0.52$ lower than that of the termination direction in all 15 pairs. Thus, the alignment reflects the termination contrast rather than shared weights alone.

\begin{figure*}[t]
    \centering
    \includegraphics[width=0.7\textwidth]{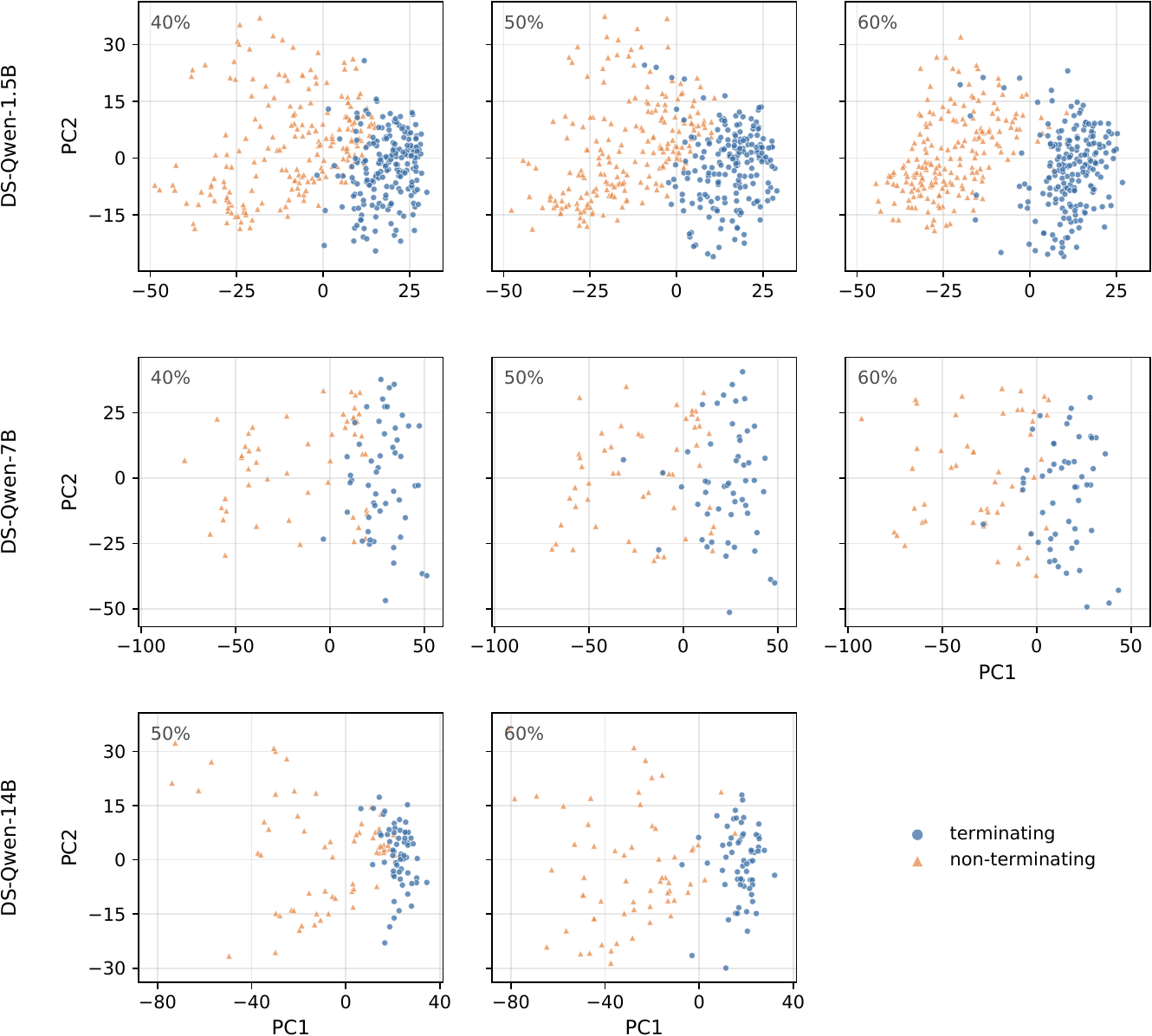}
    \caption{PCA of continuation representations at the steering layer on OpenR1-Math. For each model, PCA is fitted jointly on the continuations of all sparsity levels. Terminating continuations form a compact cluster, while non-terminating ones spread widely along the same PC1 at every sparsity level.}
    \label{fig:pca-directions}
\end{figure*}

\paragraph{Transfer across sparsity levels.} Finally, we test whether the shared direction is functional. We split the branch points into five folds, keeping all continuations of a branch point in the same fold since they share a prefix. In each fold, we construct the direction of $A$ from four folds, project the held-out continuations of $B$ onto it, and compute the ROC AUC for separating terminating from non-terminating continuations, averaged over folds and layers. Setting $A=B$ gives each direction's own AUC. As a control, we repeat the transfer with the labels of $A$ shuffled over 200 permutations. All directions achieve high within-checkpoint AUCs ($0.914$–$0.991$), while transferred directions on DS-Qwen reach $0.916$–$0.991$. Transfer AUC exceeds the 95th percentile of the shuffled-label control in 29 of 30 pairs. Together with the cosine results, these findings show that the shared direction consistently separates terminating from non-terminating continuations across sparsity levels, as also illustrated by the PCA projections in Figure ~\ref{fig:pca-directions}.

\begin{table*}[t]
\centering
\scriptsize
\setlength{\tabcolsep}{3pt}
\begin{minipage}[t]{0.49\textwidth}
\vspace{0pt}\centering
\begin{tabular}{ll r r c r}
\toprule
\multicolumn{6}{c}{\textbf{(a) OpenR1-Math}} \\
\midrule
\thd{Model} & \thd{Levels} & \thd{Transfer} & \thd{Own} & \thd{Shuffled} & \thd{$p$} \\
\midrule
DS-Qwen-1.5B & 40 $\to$ 50 & 0.973 & 0.976 & 0.839 / 0.950 & 0.005 \\
DS-Qwen-1.5B & 40 $\to$ 60 & 0.974 & 0.977 & 0.848 / 0.945 & 0.005 \\
DS-Qwen-1.5B & 50 $\to$ 40 & 0.971 & 0.970 & 0.833 / 0.940 & 0.005 \\
DS-Qwen-1.5B & 50 $\to$ 60 & 0.977 & 0.977 & 0.851 / 0.956 & 0.005 \\
DS-Qwen-1.5B & 60 $\to$ 40 & 0.960 & 0.970 & 0.827 / 0.933 & 0.010 \\
DS-Qwen-1.5B & 60 $\to$ 50 & 0.970 & 0.976 & 0.834 / 0.944 & 0.005 \\
\midrule
DS-Qwen-7B & 40 $\to$ 50 & 0.930 & 0.914 & 0.811 / 0.905 & 0.005 \\
DS-Qwen-7B & 40 $\to$ 60 & 0.916 & 0.922 & 0.803 / 0.899 & 0.010 \\
DS-Qwen-7B & 50 $\to$ 40 & 0.942 & 0.964 & 0.803 / 0.919 & 0.015 \\
DS-Qwen-7B & 50 $\to$ 60 & 0.917 & 0.922 & 0.796 / 0.903 & 0.010 \\
DS-Qwen-7B & 60 $\to$ 40 & 0.934 & 0.964 & 0.804 / 0.922 & 0.020 \\
DS-Qwen-7B & 60 $\to$ 50 & 0.919 & 0.914 & 0.793 / 0.903 & 0.020 \\
\midrule
DS-Qwen-14B & 50 $\to$ 60 & 0.971 & 0.974 & 0.828 / 0.937 & 0.010 \\
DS-Qwen-14B & 60 $\to$ 50 & 0.956 & 0.975 & 0.799 / 0.920 & 0.015 \\
\bottomrule
\end{tabular}
\end{minipage}\hfill
\begin{minipage}[t]{0.49\textwidth}
\vspace{0pt}\centering
\begin{tabular}{ll r r c r}
\toprule
\multicolumn{6}{c}{\textbf{(b) MATH-500}} \\
\midrule
\thd{Model} & \thd{Levels} & \thd{Transfer} & \thd{Own} & \thd{Shuffled} & \thd{$p$} \\
\midrule
DS-Qwen-1.5B & 40 $\to$ 50 & 0.971 & 0.976 & 0.868 / 0.956 & 0.005 \\
DS-Qwen-1.5B & 40 $\to$ 60 & 0.979 & 0.985 & 0.881 / 0.964 & 0.005 \\
DS-Qwen-1.5B & 50 $\to$ 40 & 0.972 & 0.971 & 0.858 / 0.955 & 0.005 \\
DS-Qwen-1.5B & 50 $\to$ 60 & 0.985 & 0.985 & 0.879 / 0.964 & 0.005 \\
DS-Qwen-1.5B & 60 $\to$ 40 & 0.966 & 0.971 & 0.852 / 0.951 & 0.005 \\
DS-Qwen-1.5B & 60 $\to$ 50 & 0.972 & 0.976 & 0.865 / 0.959 & 0.005 \\
\midrule
DS-Qwen-7B & 50 $\to$ 60 & 0.985 & 0.984 & 0.871 / 0.970 & 0.005 \\
DS-Qwen-7B & 50 $\to$ 70 & 0.988 & 0.991 & 0.907 / 0.981 & 0.015 \\
DS-Qwen-7B & 60 $\to$ 50 & 0.973 & 0.975 & 0.864 / 0.961 & 0.010 \\
DS-Qwen-7B & 60 $\to$ 70 & 0.991 & 0.991 & 0.888 / 0.979 & 0.005 \\
DS-Qwen-7B & 70 $\to$ 50 & 0.967 & 0.975 & 0.854 / 0.957 & 0.035 \\
DS-Qwen-7B & 70 $\to$ 60 & 0.978 & 0.984 & 0.852 / 0.963 & 0.015 \\
\midrule
DS-Qwen-14B & 60 $\to$ 70 & 0.973 & 0.983 & 0.879 / 0.960 & 0.010 \\
DS-Qwen-14B & 70 $\to$ 60 & 0.980 & 0.985 & 0.868 / 0.954 & 0.005 \\
\midrule
OT3-7B & 60 $\to$ 70 & 0.982 & 0.941 & 0.847 / 0.931 & 0.005 \\
OT3-7B & 70 $\to$ 60 & 0.733 & 0.939 & 0.686 / 0.784 & 0.204 \\
\bottomrule
\end{tabular}
\end{minipage}
\caption{Held-out ROC AUC for separating terminating from non-terminating continuations when a direction constructed at one sparsity level (\%) is applied to another. \thd{Transfer}: AUC of the source direction on the target's held-out continuations; \thd{Own}: AUC of the target's own direction; \thd{Shuffled}: mean / 95th percentile of the transfer AUC when the source labels are permuted (200 permutations), with permutation $p$-value \thd{$p$}.}
\label{tab:transfer-auc}
\end{table*}

In conclusion, our results show that: (i) The cosine similarity of termination directions is consistently higher than that of all ablation baselines; (ii) Directions transfer across sparsity levels, consistently separating terminating from non-terminating continuations; (iii) These findings support learning one termination direction per model and reusing it across sparsity levels.
\FloatBarrier
\clearpage

\section{Additional Main Experimental Results}
\label{app:full_results}

\subsection{Comprehensive Evaluation of Training-free Methods}
\label{app:full_result_comparison}

\paragraph{Overall accuracy.} Tables~\ref{tab:full-main-results-1} and~\ref{tab:full-main-results-2} show that TGAS achieves the best or tied-best accuracy on 53 of 72 checkpoint-benchmark pairs, compared with 10 for Forced-Answer. It ranks first in 20 of 24 model–sparsity settings and outperforms Base in 22, with the mean gain increasing from 1.6 points on dense models to 3.7 points at 60\% sparsity. For OT3-7B at 60\% sparsity, TGAS improves LiveCodeBench and AIME-25 accuracy by 5 points each. TGAS underperforms Base on only 8 of 72 pairs, five on AIME-25. TGAS improves termination over Base on 65 of 72 pairs, including 51 of 54 pruned settings, with gains increasing from 2.6 points on dense models to 11.2 points at 60\% sparsity. These gains generally translate into higher accuracy: among the 65 pairs with improved termination, accuracy increases on 53 and decreases on only 6. For example, on MATH-500 at 60\% sparsity, TGAS raises termination and accuracy from 21.9\% to 67.9\% and 15.5\% to 30.9\%, respectively, for DS-Qwen-1.5B. 

\paragraph{Comparison with SEAL.} SEAL improves termination by a similar margin, 5.8 points on average, compared with 5.9 for TGAS and raises it on 62 of 72 pairs, but often at the cost of accuracy. Its accuracy is below Base on 40 of 72 pairs, including 33 of the 62 pairs where it increases termination, and its average accuracy falls below Base in 14 of 24 model-sparsity settings. For example, on DS-Llama-8B at 10\% sparsity, SEAL raises AIME-25 termination from 74.2\% to 88.3\% while accuracy drops from 35.8\% to 30.8\%; on DS-Qwen-7B at 60\% sparsity, AIME-25 termination rises from 43.3\% to 60.8\% while accuracy drops from 22.5\% to 16.7\%. TGAS is more accurate than SEAL on 62 of 72 pairs and less accurate on only 3, all on AIME-25 with dense or 10\%-sparsity OT3 models.

\paragraph{Forced-Answer and No-Repeat.} Forced-Answer is most competitive at 60\% sparsity, accounting for 7 of its 10 best or tied-best results and improving average accuracy by 2.4 points over Base. However, it reduces accuracy on 27 of 36 dense and 10\%-sparsity pairs, limiting its practicality when problem difficulty is unknown. No-Repeat yields inconsistent gains, ranking best or tied-best on only 4 pairs while reducing accuracy on 41 of 72. It also underperforms Base on LiveCodeBench by 6.9 points on average, sometimes increasing termination at the expense of accuracy, e.g., 80.3\% vs.\ 75.8\% termination but 46.6\% vs.\ 62.4\% accuracy for Qwen-14B at 50\% sparsity.

\begin{table*}[t]
\centering
\scriptsize
\setlength{\tabcolsep}{4pt}
\resizebox{0.7\textwidth}{!}{%
\begin{tabular}{ll r rrr rrr}
\toprule
 & & & \multicolumn{3}{c}{\thd{Accuracy (\%)}} & \multicolumn{3}{c}{\thd{Under TGAS (\%)}} \\
\cmidrule(lr){4-6}\cmidrule(lr){7-9}
\thd{Model} & \thd{Ckpt} & \thd{\#NT} & \thd{Base} & \thd{Oracle} & \thd{TGAS} & \thd{NT$\to$Correct} & \thd{Wrong$\to$Correct} & \thd{Kept} \\
\midrule
\multirow{4}{*}{DS-Qwen-1.5B} & \dns & 22 & 75.2\sd{0.5} & 75.8\sd{1.4} & 77.1\sd{5.4} & 19.6\sd{4.0} & 43.2\sd{9.5} & 90.9\sd{1.5} \\
 & \ckpt{10} & 22 & 73.3\sd{2.7} & 73.7\sd{2.7} & 76.7\sd{0.0} & 29.0\sd{9.5} & 42.5\sd{6.5} & 90.7\sd{4.8} \\
 & \ckpt{50} & 72 & 54.8\sd{2.4} & 57.4\sd{3.0} & 61.8\sd{4.9} & 33.7\sd{1.9} & 42.2\sd{13.9} & 82.2\sd{2.3} \\
 & \ckpt{60} & 204 & 15.5\sd{3.0} & 32.6\sd{3.0} & 30.9\sd{1.1} & 25.8\sd{0.3} & 9.2\sd{5.5} & 65.8\sd{3.9} \\
\midrule
\multirow{5}{*}{DS-Qwen-7B} & \dns & 4 & 85.1\sd{1.1} & 85.1\sd{1.1} & 86.3\sd{1.9} & 16.6\sd{23.5} & 44.0\sd{3.7} & 94.2\sd{2.5} \\
 & \ckpt{10} & 6 & 84.7\sd{0.0} & 84.9\sd{0.3} & 88.2\sd{0.8} & 52.6\sd{53.1} & 45.6\sd{6.2} & 95.7\sd{0.3} \\
 & \ckpt{50} & 20 & 83.6\sd{1.1} & 84.5\sd{1.4} & 83.6\sd{0.8} & 52.3\sd{7.6} & 37.5\sd{1.2} & 91.3\sd{1.8} \\
 & \ckpt{60} & 57 & 70.2\sd{0.5} & 75.2\sd{0.5} & 73.7\sd{0.3} & 36.1\sd{2.8} & 45.4\sd{3.4} & 88.6\sd{0.9} \\
 & \ckpt{70} & 190 & 21.9\sd{0.8} & 43.7\sd{3.0} & 39.7\sd{3.0} & 31.2\sd{3.4} & 7.1\sd{6.9} & 74.6\sd{5.7} \\
\midrule
\multirow{5}{*}{DS-Qwen-14B} & \dns & 2 & 88.4\sd{0.3} & 88.4\sd{0.3} & 89.3\sd{0.5} & 48.6\sd{68.8} & 33.4\sd{16.8} & 96.5\sd{1.2} \\
 & \ckpt{10} & 4 & 87.8\sd{1.1} & 87.8\sd{1.1} & 89.7\sd{0.8} & 51.5\sd{6.6} & 54.9\sd{11.4} & 94.6\sd{2.7} \\
 & \ckpt{50} & 12 & 82.4\sd{0.5} & 82.8\sd{0.0} & 86.3\sd{0.5} & 24.3\sd{8.8} & 52.4\sd{8.2} & 95.1\sd{0.9} \\
 & \ckpt{60} & 52 & 67.6\sd{1.6} & 70.0\sd{0.3} & 72.5\sd{1.3} & 44.2\sd{5.5} & 37.6\sd{4.2} & 87.3\sd{2.5} \\
 & \ckpt{70} & 201 & 15.7\sd{1.1} & 33.0\sd{0.8} & 22.1\sd{2.6} & 17.5\sd{2.7} & 5.3\sd{5.1} & 51.2\sd{7.6} \\
\bottomrule
\end{tabular}}
\caption{Headroom and recovery on MATH-500. \thd{\#NT}: number of non-terminating Base generations. \thd{Oracle}: Base accuracy if every non-terminating generation that already states the gold answer explicitly were counted as correct. \thd{NT$\to$Correct} and \thd{Wrong$\to$Correct}: share of non-terminating and terminated-wrong Base generations that become correct under TGAS. \thd{Kept}: share of Base-correct generations that remain correct.}
\label{tab:headroom-recovery}
\end{table*}

\begin{table*}[t]
\begin{center}
\scriptsize
\begin{tabular}{lll rr rr rr}
\toprule
 & & & \multicolumn{2}{c}{MATH-500} & \multicolumn{2}{c}{AIME-25} & \multicolumn{2}{c}{LiveCodeBench} \\
\cmidrule(lr){4-5}\cmidrule(lr){6-7}\cmidrule(lr){8-9}
\thd{Model} & \thd{Ckpt} & \thd{Method} & \thd{Acc} & \thd{Term} & \thd{Acc} & \thd{Term} & \thd{Acc} & \thd{Term} \\
\midrule
\multirow{20}{*}{\makecell{Deepseek-R1-Distill\\-Qwen-1.5B}} & \multirow{5}{*}{\dns} & Base & 75.2$\pm$0.5 & 91.6$\pm$0.0 & 22.5$\pm$4.2 & 71.7$\pm$5.8 & 34.5$\pm$1.6 & 76.1$\pm$1.8 \\
 &  & No-Repeat & 71.0$\pm$0.0 & 91.2$\pm$0.5 & 22.5$\pm$4.2 & 74.2$\pm$7.4 & 27.6$\pm$1.6 & 80.8$\pm$3.9 \\
 &  & Forced-Answer & 74.2$\pm$2.4 & --- & 22.5$\pm$4.2 & --- & 35.3$\pm$1.8 & --- \\
 &  & SEAL & 73.7$\pm$2.1 & 95.8$\pm$0.6 & 18.3$\pm$1.9 & 76.7$\pm$6.1 & 35.0$\pm$3.7 & 84.7$\pm$3.3 \\
 &  & TGAS & \textbf{77.1$\pm$5.4} & 92.4$\pm$1.3 & \textbf{25.8$\pm$4.2} & 75.8$\pm$5.7 & \textbf{36.8$\pm$2.6} & 86.3$\pm$3.7 \\
\cmidrule(lr){2-9}
 & \multirow{5}{*}{\ckpt{10}} & Base & 73.3$\pm$2.7 & 91.4$\pm$1.3 & 21.7$\pm$4.3 & 71.7$\pm$1.9 & 32.9$\pm$2.6 & 76.6$\pm$3.1 \\
 &  & No-Repeat & 74.0$\pm$0.5 & 91.6$\pm$1.6 & 24.2$\pm$5.7 & 70.0$\pm$8.6 & 30.5$\pm$2.3 & 82.4$\pm$3.0 \\
 &  & Forced-Answer & 72.5$\pm$2.2 & --- & 21.7$\pm$4.3 & --- & 33.4$\pm$3.4 & --- \\
 &  & SEAL & 74.4$\pm$1.4 & 95.0$\pm$1.3 & 21.7$\pm$4.3 & 80.8$\pm$7.4 & 34.5$\pm$4.6 & 85.5$\pm$3.0 \\
 &  & TGAS & \textbf{76.7$\pm$0.0} & 95.4$\pm$1.1 & \textbf{26.7$\pm$4.7} & 75.0$\pm$6.9 & \textbf{37.1$\pm$2.6} & 83.2$\pm$2.9 \\
\cmidrule(lr){2-9}
 & \multirow{5}{*}{\ckpt{50}} & Base & 54.8$\pm$2.4 & 72.5$\pm$0.5 & 13.3$\pm$2.7 & 46.7$\pm$4.7 & 13.4$\pm$2.2 & 32.1$\pm$4.3 \\
 &  & No-Repeat & 57.8$\pm$1.9 & 80.3$\pm$1.9 & \textbf{17.5$\pm$5.7} & 51.7$\pm$7.9 & 9.5$\pm$2.6 & 40.8$\pm$6.2 \\
 &  & Forced-Answer & 59.4$\pm$3.0 & --- & 13.3$\pm$2.7 & --- & \textbf{16.6$\pm$3.1} & --- \\
 &  & SEAL & 56.5$\pm$3.1 & 82.4$\pm$1.4 & 15.8$\pm$3.2 & 60.0$\pm$6.7 & 13.2$\pm$3.5 & 46.3$\pm$4.0 \\
 &  & TGAS & \textbf{61.8$\pm$4.9} & 85.1$\pm$1.9 & 15.8$\pm$1.7 & 58.3$\pm$5.8 & 16.1$\pm$1.0 & 51.8$\pm$6.9 \\
\cmidrule(lr){2-9}
 & \multirow{5}{*}{\ckpt{60}} & Base & 15.5$\pm$3.0 & 21.9$\pm$1.9 & 1.7$\pm$1.9 & 6.7$\pm$4.7 & 0.0$\pm$0.0 & 0.5$\pm$0.6 \\
 &  & No-Repeat & 26.3$\pm$7.0 & 42.4$\pm$5.4 & 2.5$\pm$5.0 & 23.3$\pm$4.7 & 0.0$\pm$0.0 & 7.4$\pm$0.9 \\
 &  & Forced-Answer & \textbf{34.7$\pm$3.8} & --- & 2.5$\pm$3.2 & --- & 0.0$\pm$0.0 & --- \\
 &  & SEAL & 30.2$\pm$3.7 & 53.4$\pm$2.9 & 5.0$\pm$6.4 & 27.5$\pm$7.4 & 0.0$\pm$0.0 & 0.0$\pm$0.0 \\
 &  & TGAS & 30.9$\pm$1.1 & 67.9$\pm$1.3 & \textbf{5.0$\pm$1.9} & 40.0$\pm$4.7 & 0.0$\pm$0.0 & 2.4$\pm$1.8 \\
\midrule
\multirow{20}{*}{\makecell{Deepseek-R1-Distill\\-Qwen-7B}} & \multirow{5}{*}{\dns} & Base & 85.1$\pm$1.1 & 98.3$\pm$0.3 & \textbf{42.5$\pm$5.7} & 80.0$\pm$7.2 & 59.2$\pm$3.8 & 83.7$\pm$1.4 \\
 &  & No-Repeat & 83.8$\pm$0.8 & 96.8$\pm$0.8 & 34.2$\pm$5.7 & 77.5$\pm$5.0 & 45.0$\pm$1.8 & 82.4$\pm$3.4 \\
 &  & Forced-Answer & 84.2$\pm$0.3 & --- & 42.5$\pm$5.7 & --- & 53.9$\pm$3.5 & --- \\
 &  & SEAL & 84.4$\pm$1.0 & 98.1$\pm$0.7 & 36.7$\pm$2.7 & 80.8$\pm$3.2 & 57.9$\pm$1.2 & 90.0$\pm$2.6 \\
 &  & TGAS & \textbf{86.3$\pm$1.9} & 97.7$\pm$1.1 & 41.7$\pm$6.9 & 79.2$\pm$8.8 & \textbf{59.7$\pm$2.2} & 89.2$\pm$3.0 \\
\cmidrule(lr){2-9}
 & \multirow{5}{*}{\ckpt{10}} & Base & 84.7$\pm$0.0 & 97.7$\pm$1.1 & 31.7$\pm$5.8 & 80.0$\pm$7.2 & 58.4$\pm$1.8 & 82.6$\pm$2.6 \\
 &  & No-Repeat & 84.2$\pm$0.8 & 96.6$\pm$1.1 & 35.0$\pm$3.3 & 78.3$\pm$4.3 & 49.5$\pm$2.3 & 84.5$\pm$1.6 \\
 &  & Forced-Answer & 83.4$\pm$0.8 & --- & 31.7$\pm$5.8 & --- & 53.4$\pm$2.6 & --- \\
 &  & SEAL & 81.7$\pm$0.6 & 96.9$\pm$1.0 & 32.5$\pm$10.7 & 80.8$\pm$5.7 & 55.8$\pm$2.6 & 86.1$\pm$1.8 \\
 &  & TGAS & \textbf{88.2$\pm$0.8} & 99.2$\pm$0.8 & \textbf{43.3$\pm$4.7} & 80.8$\pm$1.7 & \textbf{60.3$\pm$3.4} & 90.8$\pm$3.8 \\
\cmidrule(lr){2-9}
 & \multirow{5}{*}{\ckpt{50}} & Base & 83.6$\pm$1.1 & 92.2$\pm$1.3 & 30.0$\pm$4.7 & 57.5$\pm$7.4 & \textbf{42.1$\pm$1.7} & 63.9$\pm$3.1 \\
 &  & No-Repeat & 80.5$\pm$1.6 & 91.6$\pm$2.2 & 31.7$\pm$8.8 & 65.8$\pm$3.2 & 33.7$\pm$2.7 & 66.1$\pm$5.0 \\
 &  & Forced-Answer & 81.9$\pm$0.8 & --- & 30.0$\pm$4.7 & --- & 40.8$\pm$1.6 & --- \\
 &  & SEAL & 79.4$\pm$1.1 & 93.5$\pm$1.4 & 29.2$\pm$5.7 & 70.8$\pm$5.0 & 40.3$\pm$1.6 & 75.0$\pm$3.9 \\
 &  & TGAS & \textbf{83.6$\pm$0.8} & 95.0$\pm$0.8 & \textbf{31.7$\pm$3.3} & 68.3$\pm$6.9 & 41.8$\pm$2.3 & 66.6$\pm$2.2 \\
\cmidrule(lr){2-9}
 & \multirow{5}{*}{\ckpt{60}} & Base & 70.2$\pm$0.5 & 78.2$\pm$0.5 & 22.5$\pm$6.3 & 43.3$\pm$8.6 & 14.7$\pm$4.3 & 34.2$\pm$5.0 \\
 &  & No-Repeat & 69.3$\pm$1.3 & 80.9$\pm$1.6 & 22.5$\pm$4.2 & 48.3$\pm$4.3 & 14.7$\pm$1.5 & 50.0$\pm$2.5 \\
 &  & Forced-Answer & \textbf{75.0$\pm$0.3} & --- & 22.5$\pm$3.2 & --- & 16.3$\pm$4.0 & --- \\
 &  & SEAL & 71.4$\pm$0.6 & 85.9$\pm$0.8 & 16.7$\pm$4.7 & 60.8$\pm$7.9 & 18.4$\pm$1.4 & 48.4$\pm$2.7 \\
 &  & TGAS & 73.7$\pm$0.3 & 88.5$\pm$0.3 & \textbf{22.5$\pm$3.2} & 54.2$\pm$3.2 & \textbf{19.5$\pm$3.3} & 50.5$\pm$3.0 \\
\midrule
\multirow{20}{*}{\makecell{Deepseek-R1-Distill\\-Qwen-14B}} & \multirow{5}{*}{\dns} & Base & 88.4$\pm$0.3 & 99.0$\pm$0.3 & 54.2$\pm$5.0 & 87.5$\pm$5.0 & 73.2$\pm$3.9 & 90.8$\pm$2.2 \\
 &  & No-Repeat & 87.6$\pm$0.8 & 98.1$\pm$1.1 & 45.8$\pm$11.3 & 80.8$\pm$8.3 & 56.8$\pm$1.5 & 89.5$\pm$1.9 \\
 &  & Forced-Answer & 86.1$\pm$0.3 & --- & \textbf{55.0$\pm$5.8} & --- & 63.4$\pm$1.8 & --- \\
 &  & SEAL & 88.9$\pm$0.5 & 98.5$\pm$0.6 & 45.0$\pm$3.3 & 86.7$\pm$3.8 & 72.6$\pm$2.3 & 92.4$\pm$2.3 \\
 &  & TGAS & \textbf{89.3$\pm$0.5} & 99.2$\pm$0.5 & 48.3$\pm$6.4 & 83.3$\pm$3.8 & \textbf{73.4$\pm$1.8} & 92.1$\pm$2.0 \\
\cmidrule(lr){2-9}
 & \multirow{5}{*}{\ckpt{10}} & Base & 87.8$\pm$1.1 & 98.5$\pm$0.5 & 47.5$\pm$6.9 & 84.2$\pm$5.0 & 71.8$\pm$2.3 & 89.2$\pm$2.3 \\
 &  & No-Repeat & 89.5$\pm$0.3 & 98.9$\pm$0.5 & 47.5$\pm$10.0 & 83.3$\pm$2.7 & 56.1$\pm$1.8 & 88.9$\pm$2.2 \\
 &  & Forced-Answer & 86.3$\pm$1.1 & --- & 49.2$\pm$5.0 & --- & 63.9$\pm$1.3 & --- \\
 &  & SEAL & 86.6$\pm$0.8 & 98.9$\pm$0.3 & 45.0$\pm$4.3 & 85.8$\pm$5.7 & 72.4$\pm$1.6 & 93.4$\pm$2.8 \\
 &  & TGAS & \textbf{89.7$\pm$0.8} & 98.5$\pm$0.0 & \textbf{50.8$\pm$3.2} & 87.5$\pm$1.7 & \textbf{72.9$\pm$2.3} & 94.2$\pm$2.6 \\
\cmidrule(lr){2-9}
 & \multirow{5}{*}{\ckpt{50}} & Base & 82.4$\pm$0.5 & 95.4$\pm$0.5 & 35.8$\pm$5.0 & 63.3$\pm$6.1 & 62.4$\pm$1.0 & 75.8$\pm$2.3 \\
 &  & No-Repeat & 83.6$\pm$1.1 & 95.0$\pm$1.6 & 37.5$\pm$3.2 & 64.2$\pm$6.3 & 46.6$\pm$2.9 & 80.3$\pm$3.6 \\
 &  & Forced-Answer & 81.1$\pm$1.9 & --- & 36.7$\pm$4.7 & --- & 56.6$\pm$3.3 & --- \\
 &  & SEAL & 84.4$\pm$0.8 & 98.1$\pm$0.5 & 42.5$\pm$5.7 & 70.0$\pm$3.8 & 61.3$\pm$2.8 & 78.7$\pm$2.9 \\
 &  & TGAS & \textbf{86.3$\pm$0.5} & 95.4$\pm$0.3 & \textbf{42.5$\pm$3.2} & 70.8$\pm$1.7 & \textbf{62.9$\pm$3.5} & 77.1$\pm$3.5 \\
\cmidrule(lr){2-9}
 & \multirow{5}{*}{\ckpt{60}} & Base & 67.6$\pm$1.6 & 80.2$\pm$0.5 & 26.7$\pm$4.7 & 39.2$\pm$5.0 & 37.6$\pm$5.6 & 46.6$\pm$3.9 \\
 &  & No-Repeat & 71.6$\pm$1.3 & 91.4$\pm$0.8 & 25.0$\pm$3.3 & 70.0$\pm$10.5 & 33.9$\pm$1.0 & 60.3$\pm$3.0 \\
 &  & Forced-Answer & 70.8$\pm$0.8 & --- & \textbf{27.5$\pm$3.2} & --- & 38.9$\pm$4.3 & --- \\
 &  & SEAL & 71.0$\pm$0.8 & 86.6$\pm$1.8 & 24.2$\pm$4.2 & 45.0$\pm$6.4 & 42.9$\pm$2.9 & 60.0$\pm$0.9 \\
 &  & TGAS & \textbf{72.5$\pm$1.3} & 88.9$\pm$2.2 & 25.8$\pm$3.2 & 44.2$\pm$4.2 & \textbf{42.9$\pm$2.6} & 53.7$\pm$3.1 \\
\bottomrule
\end{tabular}
\end{center}
\caption{Training-free remedies for SparseGPT-pruned LRMs with OBC calibration DeepSeek-R1-Distill-Qwen models. Acc.\ and Term.\ denote accuracy and termination rate, respectively. Forced-Answer's Term.\ is not reported, since forcing ends every generation by design.}
\label{tab:full-main-results-1}
\end{table*}

\begin{table*}[t]
\begin{center}
\scriptsize
\begin{tabular}{lll rr rr rr}
\toprule
 & & & \multicolumn{2}{c}{MATH-500} & \multicolumn{2}{c}{AIME-25} & \multicolumn{2}{c}{LiveCodeBench} \\
\cmidrule(lr){4-5}\cmidrule(lr){6-7}\cmidrule(lr){8-9}
\thd{Model} & \thd{Ckpt} & \thd{Method} & \thd{Acc} & \thd{Term} & \thd{Acc} & \thd{Term} & \thd{Acc} & \thd{Term} \\
\midrule
\multirow{20}{*}{\makecell{Deepseek-R1-Distill\\-Llama-8B}} & \multirow{5}{*}{\dns} & Base & 82.4$\pm$2.7 & 97.9$\pm$0.8 & 34.2$\pm$5.7 & 70.0$\pm$11.2 & 58.7$\pm$2.5 & 85.5$\pm$1.8 \\
 &  & No-Repeat & 83.0$\pm$0.8 & 97.7$\pm$0.0 & 35.0$\pm$4.3 & 71.7$\pm$3.3 & 44.7$\pm$3.3 & 85.0$\pm$1.3 \\
 &  & Forced-Answer & 81.3$\pm$1.6 & --- & 30.8$\pm$1.7 & --- & 53.2$\pm$1.1 & --- \\
 &  & SEAL & 80.2$\pm$1.8 & 97.7$\pm$0.3 & 32.5$\pm$5.7 & 87.5$\pm$4.2 & 57.9$\pm$2.7 & 91.6$\pm$2.3 \\
 &  & TGAS & \textbf{84.0$\pm$2.2} & 97.3$\pm$0.0 & \textbf{38.3$\pm$5.8} & 80.8$\pm$3.2 & \textbf{60.8$\pm$2.2} & 87.9$\pm$1.8 \\
\cmidrule(lr){2-9}
 & \multirow{5}{*}{\ckpt{10}} & Base & \textbf{84.0$\pm$3.8} & 96.4$\pm$1.9 & 35.8$\pm$6.9 & 74.2$\pm$1.7 & \textbf{60.3$\pm$2.2} & 87.1$\pm$0.5 \\
 &  & No-Repeat & 82.3$\pm$1.9 & 96.4$\pm$0.8 & 35.8$\pm$8.8 & 75.0$\pm$4.3 & 43.9$\pm$4.1 & 86.6$\pm$3.1 \\
 &  & Forced-Answer & 81.5$\pm$2.4 & --- & 30.0$\pm$2.7 & --- & 53.2$\pm$1.1 & --- \\
 &  & SEAL & 80.5$\pm$2.7 & 98.1$\pm$0.9 & 30.8$\pm$5.0 & 88.3$\pm$6.9 & 55.5$\pm$2.2 & 93.2$\pm$1.1 \\
 &  & TGAS & 83.6$\pm$2.7 & 97.3$\pm$0.0 & \textbf{35.8$\pm$5.7} & 79.2$\pm$11.0 & 59.5$\pm$3.7 & 90.3$\pm$4.3 \\
\cmidrule(lr){2-9}
 & \multirow{5}{*}{\ckpt{50}} & Base & 65.5$\pm$0.3 & 88.5$\pm$2.2 & 21.7$\pm$7.9 & 75.0$\pm$8.4 & 42.6$\pm$0.6 & 66.8$\pm$1.8 \\
 &  & No-Repeat & 64.7$\pm$0.8 & 90.6$\pm$1.3 & 20.8$\pm$8.8 & 76.7$\pm$2.7 & 33.2$\pm$3.0 & 75.3$\pm$4.9 \\
 &  & Forced-Answer & 64.5$\pm$0.5 & --- & 16.7$\pm$6.1 & --- & 40.8$\pm$1.3 & --- \\
 &  & SEAL & 63.0$\pm$0.6 & 89.3$\pm$1.1 & 23.3$\pm$5.4 & 76.7$\pm$9.4 & 41.3$\pm$4.1 & 75.3$\pm$3.9 \\
 &  & TGAS & \textbf{67.6$\pm$1.6} & 88.5$\pm$1.1 & \textbf{23.3$\pm$3.8} & 80.8$\pm$4.2 & \textbf{43.9$\pm$3.0} & 72.1$\pm$2.0 \\
\cmidrule(lr){2-9}
 & \multirow{5}{*}{\ckpt{60}} & Base & 30.0$\pm$3.0 & 50.6$\pm$2.4 & 3.3$\pm$2.7 & 29.2$\pm$4.2 & 10.8$\pm$1.6 & 26.3$\pm$2.3 \\
 &  & No-Repeat & 33.4$\pm$0.8 & 74.4$\pm$1.1 & 2.5$\pm$3.2 & 50.8$\pm$10.7 & 9.5$\pm$1.5 & 42.1$\pm$3.7 \\
 &  & Forced-Answer & 35.7$\pm$1.3 & --- & 3.3$\pm$2.7 & --- & \textbf{11.1$\pm$1.4} & --- \\
 &  & SEAL & 32.8$\pm$1.6 & 61.8$\pm$1.3 & 4.2$\pm$5.0 & 45.0$\pm$4.3 & 10.0$\pm$3.0 & 32.6$\pm$1.9 \\
 &  & TGAS & \textbf{36.6$\pm$1.6} & 65.6$\pm$0.5 & \textbf{7.5$\pm$1.7} & 45.0$\pm$4.3 & 10.8$\pm$1.8 & 33.7$\pm$1.5 \\
\midrule
\multirow{20}{*}{\makecell{OpenThinker3\\-1.5B}} & \multirow{5}{*}{\dns} & Base & 80.2$\pm$1.6 & 87.2$\pm$1.9 & 32.5$\pm$1.7 & 39.2$\pm$3.2 & 43.2$\pm$2.9 & 57.4$\pm$3.3 \\
 &  & No-Repeat & 80.2$\pm$2.2 & 87.4$\pm$1.6 & 35.0$\pm$5.8 & 44.2$\pm$4.2 & 42.1$\pm$2.3 & 58.2$\pm$2.3 \\
 &  & Forced-Answer & 79.2$\pm$3.0 & --- & 30.8$\pm$3.2 & --- & 41.1$\pm$2.4 & --- \\
 &  & SEAL & 80.2$\pm$2.0 & 89.7$\pm$1.3 & \textbf{37.5$\pm$5.0} & 45.8$\pm$1.7 & 41.6$\pm$2.0 & 61.3$\pm$3.5 \\
 &  & TGAS & \textbf{84.0$\pm$1.1} & 89.7$\pm$0.5 & 36.7$\pm$2.7 & 43.3$\pm$3.8 & \textbf{45.3$\pm$2.3} & 61.1$\pm$0.9 \\
\cmidrule(lr){2-9}
 & \multirow{5}{*}{\ckpt{10}} & Base & 79.6$\pm$1.3 & 87.0$\pm$0.5 & 34.2$\pm$8.3 & 40.0$\pm$2.7 & 43.4$\pm$2.3 & 57.1$\pm$2.6 \\
 &  & No-Repeat & 79.2$\pm$1.9 & 86.6$\pm$2.2 & 35.0$\pm$6.9 & 41.7$\pm$6.9 & 38.9$\pm$2.6 & 55.5$\pm$1.6 \\
 &  & Forced-Answer & 77.3$\pm$3.0 & --- & 29.2$\pm$6.3 & --- & 40.8$\pm$2.6 & --- \\
 &  & SEAL & 77.1$\pm$1.9 & 87.0$\pm$2.0 & \textbf{41.7$\pm$5.8} & 52.5$\pm$5.7 & 38.7$\pm$2.0 & 61.3$\pm$5.8 \\
 &  & TGAS & \textbf{81.7$\pm$1.9} & 88.5$\pm$0.8 & 37.5$\pm$3.2 & 46.7$\pm$9.8 & \textbf{45.5$\pm$1.6} & 61.3$\pm$2.8 \\
\cmidrule(lr){2-9}
 & \multirow{5}{*}{\ckpt{50}} & Base & 56.9$\pm$3.8 & 67.4$\pm$4.0 & \textbf{20.8$\pm$1.7} & 32.5$\pm$5.0 & 10.0$\pm$0.6 & 22.4$\pm$1.3 \\
 &  & No-Repeat & 54.4$\pm$0.3 & 67.4$\pm$1.3 & 16.7$\pm$4.7 & 25.8$\pm$3.2 & 6.3$\pm$0.9 & 23.2$\pm$4.5 \\
 &  & Forced-Answer & 57.1$\pm$3.5 & --- & 17.5$\pm$5.0 & --- & 11.1$\pm$1.8 & --- \\
 &  & SEAL & 55.0$\pm$5.1 & 73.7$\pm$2.1 & 18.3$\pm$4.3 & 34.2$\pm$1.7 & 9.2$\pm$1.0 & 26.8$\pm$2.8 \\
 &  & TGAS & \textbf{58.4$\pm$2.2} & 74.4$\pm$0.3 & 19.2$\pm$7.4 & 44.2$\pm$5.7 & \textbf{13.9$\pm$2.3} & 29.2$\pm$5.6 \\
\cmidrule(lr){2-9}
 & \multirow{5}{*}{\ckpt{60}} & Base & 17.6$\pm$1.1 & 13.4$\pm$2.2 & 0.8$\pm$1.7 & 5.8$\pm$1.7 & 0.0$\pm$0.0 & 0.0$\pm$0.0 \\
 &  & No-Repeat & 19.1$\pm$1.1 & 15.8$\pm$1.9 & 2.5$\pm$1.7 & 7.5$\pm$3.2 & 0.0$\pm$0.0 & 0.3$\pm$0.5 \\
 &  & Forced-Answer & \textbf{23.1$\pm$0.3} & --- & \textbf{3.3$\pm$0.0} & --- & 0.0$\pm$0.0 & --- \\
 &  & SEAL & 14.5$\pm$1.2 & 8.4$\pm$1.6 & 0.8$\pm$1.7 & 5.8$\pm$6.9 & 0.0$\pm$0.0 & 0.5$\pm$0.6 \\
 &  & TGAS & 22.5$\pm$2.4 & 20.6$\pm$0.8 & \textbf{3.3$\pm$0.0} & 10.8$\pm$3.2 & 0.0$\pm$0.0 & 2.6$\pm$0.6 \\
\midrule
\multirow{20}{*}{\makecell{OpenThinker3\\-7B}} & \multirow{5}{*}{\dns} & Base & 80.3$\pm$4.6 & 92.0$\pm$1.6 & 44.2$\pm$5.0 & 50.0$\pm$4.7 & 63.4$\pm$2.0 & 72.9$\pm$1.8 \\
 &  & No-Repeat & 80.9$\pm$1.1 & 91.4$\pm$0.8 & 50.0$\pm$8.2 & 55.8$\pm$10.0 & 58.2$\pm$1.6 & 71.6$\pm$2.7 \\
 &  & Forced-Answer & 80.2$\pm$2.7 & --- & 32.5$\pm$1.7 & --- & 53.7$\pm$0.9 & --- \\
 &  & SEAL & 79.4$\pm$2.5 & 93.5$\pm$1.2 & \textbf{50.8$\pm$6.9} & 53.3$\pm$7.2 & 63.4$\pm$1.3 & 75.3$\pm$1.8 \\
 &  & TGAS & \textbf{84.4$\pm$1.6} & 93.5$\pm$1.3 & 45.8$\pm$4.2 & 54.2$\pm$3.2 & \textbf{64.7$\pm$1.4} & 73.9$\pm$1.8 \\
\cmidrule(lr){2-9}
 & \multirow{5}{*}{\ckpt{10}} & Base & 81.7$\pm$3.8 & 90.8$\pm$0.5 & \textbf{50.0$\pm$0.0} & 55.0$\pm$1.9 & 61.3$\pm$2.2 & 71.8$\pm$1.3 \\
 &  & No-Repeat & 80.7$\pm$3.5 & 92.0$\pm$1.1 & 46.7$\pm$2.7 & 53.3$\pm$4.7 & 56.3$\pm$1.8 & 68.9$\pm$3.9 \\
 &  & Forced-Answer & 82.3$\pm$5.1 & --- & 35.8$\pm$3.2 & --- & 52.4$\pm$2.2 & --- \\
 &  & SEAL & 83.6$\pm$3.2 & 94.7$\pm$0.7 & 46.7$\pm$4.7 & 60.0$\pm$7.2 & 62.1$\pm$1.7 & 74.5$\pm$1.8 \\
 &  & TGAS & \textbf{84.7$\pm$0.3} & 94.3$\pm$0.5 & 48.3$\pm$5.8 & 56.7$\pm$5.4 & \textbf{63.4$\pm$0.5} & 74.2$\pm$1.8 \\
\cmidrule(lr){2-9}
 & \multirow{5}{*}{\ckpt{50}} & Base & 84.5$\pm$1.9 & 89.7$\pm$2.2 & 37.5$\pm$4.2 & 50.8$\pm$6.9 & 51.1$\pm$1.1 & 63.2$\pm$2.6 \\
 &  & No-Repeat & 80.9$\pm$0.5 & 88.5$\pm$1.1 & 39.2$\pm$5.7 & 50.8$\pm$3.2 & 47.6$\pm$4.0 & 65.8$\pm$3.0 \\
 &  & Forced-Answer & 82.3$\pm$0.3 & --- & 28.3$\pm$4.3 & --- & 47.4$\pm$1.7 & --- \\
 &  & SEAL & 81.7$\pm$1.1 & 88.9$\pm$1.2 & 35.0$\pm$1.9 & 54.2$\pm$5.0 & 51.1$\pm$3.6 & 68.7$\pm$2.6 \\
 &  & TGAS & \textbf{85.5$\pm$1.6} & 91.6$\pm$0.3 & \textbf{43.3$\pm$8.2} & 52.5$\pm$9.6 & \textbf{55.5$\pm$2.2} & 69.5$\pm$3.0 \\
\cmidrule(lr){2-9}
 & \multirow{5}{*}{\ckpt{60}} & Base & 72.3$\pm$1.3 & 83.6$\pm$1.1 & 24.2$\pm$5.7 & 47.5$\pm$6.3 & 31.3$\pm$2.6 & 54.2$\pm$1.8 \\
 &  & No-Repeat & 71.2$\pm$0.3 & 83.2$\pm$0.5 & 25.8$\pm$4.2 & 44.2$\pm$5.7 & 26.3$\pm$3.0 & 53.9$\pm$4.4 \\
 &  & Forced-Answer & 71.9$\pm$0.8 & --- & 20.8$\pm$3.2 & --- & 32.4$\pm$1.8 & --- \\
 &  & SEAL & 71.0$\pm$0.9 & 87.0$\pm$0.6 & 27.5$\pm$4.2 & 50.8$\pm$4.2 & 30.3$\pm$3.7 & 60.5$\pm$2.2 \\
 &  & TGAS & \textbf{74.8$\pm$1.1} & 84.4$\pm$0.3 & \textbf{29.2$\pm$1.7} & 50.8$\pm$6.9 & \textbf{36.3$\pm$3.0} & 58.9$\pm$3.1 \\
\bottomrule
\end{tabular}
\end{center}
\caption{Training-free remedies for SparseGPT-pruned LRMs with OBC calibration DeepSeek-R1-Distill-Llama and OpenThinker3 models. Acc.\ and Term.\ denote accuracy and termination rate, respectively. Forced-Answer's Term.\ is not reported, since forcing ends every generation by design.}
\label{tab:full-main-results-2}
\end{table*}

\subsection{Comparison of Training-Free and Training-Based Methods}
\label{app:lora}

Table~\ref{tab:lora} summarizes the results. At 50\% sparsity, training-free methods match or outperform Masked-LoRA without additional training. At 60\% sparsity, Masked-LoRA achieves stronger recovery, improving LiveCodeBench accuracy from 14.7\% to 33.4\% for DS-Qwen-7B and from 37.6\% to 51.8\% for DS-Qwen-14B, surpassing the best training-free results (19.5\% and 42.9\%). On MATH-500, it reaches 37.8\%, 74.8\%, and 79.0\% for Qwen-1.5B, 7B, and 14B, respectively. These results reveal a trade-off: training-based recovery is more effective under severe pruning but requires optimization and a separate LoRA adapter for each sparsity level, whereas TGAS remains a competitive alternative without weight updates.

\begin{table*}[t]
\scriptsize
\begin{center}
\begin{tabular}{lll rr rr rr}
\toprule
 & & & \multicolumn{2}{c}{MATH-500} & \multicolumn{2}{c}{AIME-25} & \multicolumn{2}{c}{LiveCodeBench} \\
\cmidrule(lr){4-5}\cmidrule(lr){6-7}\cmidrule(lr){8-9}
\thd{Model} & \thd{Ckpt} & \thd{Method} & \thd{Acc} & \thd{Term} & \thd{Acc} & \thd{Term} & \thd{Acc} & \thd{Term} \\
\midrule
\multirow{8}{*}{\makecell{Deepseek-R1-Distill\\-Qwen-1.5B}} & \multirow{4}{*}{\ckpt{50}} & Base & 54.8$\pm$2.4 & 72.5$\pm$0.5 & 13.3$\pm$2.7 & 46.7$\pm$4.7 & 13.4$\pm$2.2 & 32.1$\pm$4.3 \\
 &  & Forced-Answer & 59.4$\pm$3.0 & --- & 13.3$\pm$2.7 & --- & \textbf{16.6$\pm$3.1} & --- \\
 &  & TGAS & \textbf{61.8$\pm$4.9} & 85.1$\pm$1.9 & \textbf{15.8$\pm$1.7} & 58.3$\pm$5.8 & 16.1$\pm$1.0 & 51.8$\pm$6.9 \\
 &  & Masked-LoRA & 60.3$\pm$3.1 & 82.1$\pm$1.4 & 11.7$\pm$5.8 & 52.5$\pm$5.7 & 15.3$\pm$1.8 & 30.8$\pm$4.3 \\
\cmidrule(lr){2-9}
 & \multirow{4}{*}{\ckpt{60}} & Base & 15.5$\pm$3.0 & 21.9$\pm$1.9 & 1.7$\pm$1.9 & 6.7$\pm$4.7 & 0.0$\pm$0.0 & 0.5$\pm$0.6 \\
 &  & Forced-Answer & 34.7$\pm$3.8 & --- & 2.5$\pm$3.2 & --- & 0.0$\pm$0.0 & --- \\
 &  & TGAS & 30.9$\pm$1.1 & 67.9$\pm$1.3 & 5.0$\pm$1.9 & 40.0$\pm$4.7 & 0.0$\pm$0.0 & 2.4$\pm$1.8 \\
 &  & Masked-LoRA & \textbf{37.8$\pm$3.7} & 62.2$\pm$2.9 & \textbf{5.8$\pm$4.2} & 45.8$\pm$5.7 & \textbf{0.3$\pm$0.5} & 5.3$\pm$1.5 \\
\midrule
\multirow{12}{*}{\makecell{Deepseek-R1-Distill\\-Qwen-7B}} & \multirow{4}{*}{\ckpt{50}} & Base & 83.6$\pm$1.1 & 92.2$\pm$1.3 & 30.0$\pm$4.7 & 57.5$\pm$7.4 & \textbf{42.1$\pm$1.7} & 63.9$\pm$3.1 \\
 &  & Forced-Answer & 81.9$\pm$0.8 & --- & 30.0$\pm$4.7 & --- & 40.8$\pm$1.6 & --- \\
 &  & TGAS & \textbf{83.6$\pm$0.8} & 95.0$\pm$0.8 & \textbf{31.7$\pm$3.3} & 68.3$\pm$6.9 & 41.8$\pm$2.3 & 66.6$\pm$2.2 \\
 &  & Masked-LoRA & 81.7$\pm$1.1 & 89.3$\pm$1.4 & 29.2$\pm$3.2 & 52.5$\pm$4.2 & 41.6$\pm$1.4 & 60.0$\pm$2.3 \\
\cmidrule(lr){2-9}
 & \multirow{4}{*}{\ckpt{60}} & Base & 70.2$\pm$0.5 & 78.2$\pm$0.5 & 22.5$\pm$6.3 & 43.3$\pm$8.6 & 14.7$\pm$4.3 & 34.2$\pm$5.0 \\
 &  & Forced-Answer & \textbf{75.0$\pm$0.3} & --- & 22.5$\pm$3.2 & --- & 16.3$\pm$4.0 & --- \\
 &  & TGAS & 73.7$\pm$0.3 & 88.5$\pm$0.3 & 22.5$\pm$3.2 & 54.2$\pm$3.2 & 19.5$\pm$3.3 & 50.5$\pm$3.0 \\
 &  & Masked-LoRA & 74.8$\pm$0.6 & 85.5$\pm$0.8 & \textbf{26.7$\pm$2.7} & 51.7$\pm$7.9 & \textbf{33.4$\pm$2.3} & 52.6$\pm$3.3 \\
\cmidrule(lr){2-9}
 & \multirow{4}{*}{\ckpt{70}} & Base & 21.9$\pm$0.8 & 24.8$\pm$0.5 & 5.0$\pm$3.3 & 19.2$\pm$3.2 & 0.0$\pm$0.0 & 0.0$\pm$0.0 \\
 &  & Forced-Answer & 43.3$\pm$1.3 & --- & 5.0$\pm$3.3 & --- & 0.0$\pm$0.0 & --- \\
 &  & TGAS & 39.7$\pm$1.1 & 61.1$\pm$0.5 & 5.8$\pm$4.2 & 23.3$\pm$8.6 & 0.0$\pm$0.0 & 5.0$\pm$1.0 \\
 &  & Masked-LoRA & \textbf{56.1$\pm$1.1} & 76.3$\pm$0.5 & \textbf{15.0$\pm$3.3} & 45.8$\pm$5.0 & \textbf{0.5$\pm$1.1} & 20.0$\pm$3.0 \\
\midrule
\multirow{12}{*}{\makecell{Deepseek-R1-Distill\\-Qwen-14B}} & \multirow{4}{*}{\ckpt{50}} & Base & 82.4$\pm$0.5 & 95.4$\pm$0.5 & 35.8$\pm$5.0 & 63.3$\pm$6.1 & 62.4$\pm$1.0 & 75.8$\pm$2.3 \\
 &  & Forced-Answer & 81.1$\pm$1.9 & --- & 36.7$\pm$4.7 & --- & 56.6$\pm$3.3 & --- \\
 &  & TGAS & \textbf{86.3$\pm$0.5} & 95.4$\pm$0.3 & \textbf{42.5$\pm$3.2} & 70.8$\pm$1.7 & \textbf{62.9$\pm$3.5} & 77.1$\pm$3.5 \\
 &  & Masked-LoRA & 82.1$\pm$1.0 & 92.4$\pm$0.8 & 32.5$\pm$5.0 & 49.2$\pm$1.7 & 60.0$\pm$0.9 & 69.7$\pm$2.3 \\
\cmidrule(lr){2-9}
 & \multirow{4}{*}{\ckpt{60}} & Base & 67.6$\pm$1.6 & 80.2$\pm$0.5 & 26.7$\pm$4.7 & 39.2$\pm$5.0 & 37.6$\pm$5.6 & 46.6$\pm$3.9 \\
 &  & Forced-Answer & 70.8$\pm$0.8 & --- & \textbf{27.5$\pm$3.2} & --- & 38.9$\pm$4.3 & --- \\
 &  & TGAS & 72.5$\pm$1.3 & 88.9$\pm$2.2 & 25.8$\pm$3.2 & 44.2$\pm$4.2 & 42.9$\pm$2.6 & 53.7$\pm$3.1 \\
 &  & Masked-LoRA & \textbf{79.0$\pm$1.3} & 88.9$\pm$1.2 & 24.2$\pm$5.0 & 46.7$\pm$7.2 & \textbf{51.8$\pm$1.6} & 65.0$\pm$3.4 \\
\cmidrule(lr){2-9}
 & \multirow{4}{*}{\ckpt{70}} & Base & 15.6$\pm$1.1 & 23.1$\pm$0.3 & 3.3$\pm$2.7 & 10.8$\pm$3.2 & 0.0$\pm$0.0 & 0.0$\pm$0.0 \\
 &  & Forced-Answer & 31.9$\pm$1.3 & --- & 5.8$\pm$4.2 & --- & 0.0$\pm$0.0 & --- \\
 &  & TGAS & 22.1$\pm$1.2 & 37.4$\pm$0.3 & 5.8$\pm$1.7 & 16.7$\pm$2.7 & 0.0$\pm$0.0 & 0.3$\pm$0.5 \\
 &  & Masked-LoRA & \textbf{58.0$\pm$1.2} & 72.5$\pm$0.3 & \textbf{12.5$\pm$1.7} & 34.2$\pm$5.0 & \textbf{17.9$\pm$3.3} & 33.2$\pm$2.5 \\
\bottomrule
\end{tabular}
\end{center}
\caption{Training-free remedies vs.\ Masked-LoRA for SparseGPT-pruned LRMs with OBC calibration DeepSeek-R1-Distill-Qwen models. Forced-Answer's \thd{Term} is not reported, since forcing ends every generation by design.}
\label{tab:lora}
\end{table*}

\subsection{Headroom and Recovery Analysis}
\label{app:headroom}

Pruned models often fail to terminate even after their reasoning has already reached the correct answer. We ask two questions: how much accuracy would be regained if these generations simply committed to the answer they already state, and how many non-terminating generations TGAS actually turns into correct ones.

\paragraph{Setup.} We evaluate DS-Qwen-1.5B, 7B, and 14B on 262 level 4-5 MATH-500 problems, from dense to 70\% sparsity, using temperature $0.6$ and a $16{,}384$-token budget. We define oracle accuracy as Base accuracy plus the fraction of non-terminating generations that already state the gold answer \emph{Conclusively} (the same rule in Appendix \ref{app:presence}). For recovery, we measure the fraction of initially incorrect non-terminating and terminated-wrong generations corrected by TGAS on the same problems and seeds, and the fraction of Base-correct generations that remain correct.

\paragraph{Headroom.} At 50\% sparsity, the oracle is at most $2.6$ points above Base (Table~\ref{tab:headroom-recovery}). Headroom grows under heavier pruning: committing to an already stated answer would raise DS-Qwen-1.5B at 60\% from 15.5\% to 32.6\%, and DS-Qwen-7B and 14B at 70\% from 21.9\% to 43.7\% and from 15.7\% to 33.0\%. TGAS captures 90\% and 70\% of this headroom for DS-Qwen-1.5B and 7B at 60\%, exceeds the oracle for DS-Qwen-14B, and captures 82\% and 37\% for DS-Qwen-7B and 14B at 70\%.

\paragraph{Recovery.} At 60\% sparsity, TGAS turns 25.8\%, 36.1\% and 44.2\% of non-terminating generations into correct ones for DS-Qwen-1.5B, 7B and 14B, against 17.1\% (pooled over models) for a rerun of the unsteered model with another seed. In contrast, it does not help generations that terminated with a wrong answer: pooled over models, 31.2\% of them become correct under TGAS and 31.9\% under the rerun. TGAS therefore restores termination rather than correcting wrong answers, consistent with the intended behavior of its steering method.

\clearpage
\newpage

\section{Additional Ablation Studies}
\label{app:full_ablation}

\begin{table}[t]
\begin{center}
\tabsmall
\setlength{\tabcolsep}{3pt}
\resizebox{0.9\columnwidth}{!}{%
\begin{tabular}{ll rr rr}
\toprule
 & & \multicolumn{2}{c}{$\Delta_{\mathrm{cyc}}$ at decisions} & \multicolumn{2}{c}{\thd{DPs/1k}} \\
\cmidrule(lr){3-4}\cmidrule(lr){5-6}
\thd{Model} & \thd{Ckpt} & \thd{U} & \thd{S} & \thd{U} & \thd{S} \\
\midrule
\multirow{4}{*}{\makecell{Deepseek-R1-Distill\\-Qwen-1.5B}} & \dns & $\num{-14.9}\pm$0.5 & $\num{-16.4}\pm$0.6 & 8.1 & 5.6 \\
 & \ckpt{10} & $\num{-15.3}\pm$0.5 & $\num{-16.5}\pm$0.6 & 8.0 & 5.3 \\
 & \ckpt{50} & $\num{-10.2}\pm$0.4 & $\num{-12.7}\pm$0.5 & 9.2 & 5.2 \\
 & \ckpt{60} & $\num{-5.6}\pm$0.3 & $\num{-6.3}\pm$0.4 & 23.7 & 10.5 \\
\midrule
\multirow{4}{*}{\makecell{Deepseek-R1-Distill\\-Qwen-7B}} & \dns & $\num{-18.2}\pm$0.6 & $\num{-20.5}\pm$0.6 & 7.1 & 4.7 \\
 & \ckpt{10} & $\num{-18.2}\pm$0.5 & $\num{-19.5}\pm$0.6 & 7.4 & 4.5 \\
 & \ckpt{50} & $\num{-15.3}\pm$0.5 & $\num{-17.1}\pm$0.5 & 9.0 & 5.4 \\
 & \ckpt{60} & $\num{-12.0}\pm$0.4 & $\num{-15.4}\pm$0.5 & 12.3 & 5.8 \\
\midrule
\multirow{4}{*}{\makecell{Deepseek-R1-Distill\\-Qwen-14B}} & \dns & $\num{-16.1}\pm$0.5 & $\num{-18.1}\pm$0.5 & 6.6 & 4.8 \\
 & \ckpt{10} & $\num{-16.0}\pm$0.4 & $\num{-18.2}\pm$0.5 & 6.7 & 4.8 \\
 & \ckpt{50} & $\num{-14.7}\pm$0.4 & $\num{-17.9}\pm$0.5 & 7.2 & 5.5 \\
 & \ckpt{60} & $\num{-12.7}\pm$0.4 & $\num{-14.0}\pm$0.4 & 11.5 & 6.8 \\
\bottomrule
\end{tabular}}
\end{center}
\caption{$\Delta_{\mathrm{cyc}}$ at decision points on MATH-500 L4--5 with and without TGAS. \thd{U}: unsteered traces; \thd{S}: steered traces scored with the steering hook. More negative values indicate stronger preference for moving on. \thd{DPs/1k}: decision points per 1{,}000 generated tokens.}
\label{tab:steer-deltacyc}
\end{table}

\subsection{Behavioral Changes in Pruned Models After Steering}
\label{app:behavior}

Section~\ref{subsec:behavior} showed that pruning makes re-treading increasingly competitive at decision points. $\Delta_{\mathrm{cyc}}=\log P(\textsc{cyclic})-\log P(\textsc{progress})$, measured over the next 32 tokens after a reconsideration cue (\emph{"Wait"}, \emph{"Alternatively"}, \dots), approaches zero as sparsity increases. We ask whether TGAS reverses this trend. We repeat the measurement on steered DeepSeek-R1-Distill-Qwen models, using decision points from the steered traces. 

Table~\ref{tab:steer-deltacyc} reports MATH-500 results for the three DeepSeek-R1-Distill-Qwen models. TGAS makes $\Delta_{\mathrm{cyc}}$ more negative in all 12 settings, with shifts of $0.7$-$3.4$. More consistently, it reduces decision points by 24\%-56\%, with the largest reductions at 60\% sparsity: from $23.7$ to $10.5$, $12.3$ to $5.8$, and $11.5$ to $6.8$ per 1k tokens for the 1.5B, 7B, and 14B models, respectively. For the latter two, this restores reconsideration frequency to approximately dense-model levels. Overall, TGAS primarily suppresses repeated reconsideration and promotes earlier termination.

\subsection{Ablation Study of Forced-Answer} 
\label{app:ablation-force-answer}

Table~\ref{tab:ablation-forcing} shows that the effectiveness of Forced-Answer depends strongly on the generation budget. Short budgets can truncate useful reasoning in capable models, e.g., dense Qwen-14B drops from 54.2\% to 25.0\% on AIME-25 at $D=4{,}096$, while longer budgets are less effective for heavily pruned models. At $D=16{,}384$, forcing only at the token limit is generally safe but provides limited gains. Overall, no single budget works across models and sparsity levels. In practice, selecting $D$ requires estimating the difficulty of each problem, since a budget sufficient for difficult problems may unnecessarily truncate easier ones. This dependence on problem difficulty is a major practical drawback of Forced-Answer.

\begin{table*}[t]
\scriptsize
\begin{center}
\setlength{\tabcolsep}{3pt}
\resizebox{0.7\textwidth}{!}{%
\begin{tabular}{ll rrrr rrr}
\toprule
 & & \multicolumn{4}{c}{AIME-25} & \multicolumn{3}{c}{LiveCodeBench} \\
\cmidrule(lr){3-6}\cmidrule(lr){7-9}
\thd{Model} & \thd{Ckpt} & Base & $D$=4096 & $D$=8192 & $D$=16384 & Base & $D$=4096 & $D$=8192 \\
\midrule
\multirow{5}{*}{\makecell{Deepseek-R1-Distill\\-Qwen-1.5B}} & \dns & 22.5$\pm$4.2 & 19.2$\pm$1.7 & 22.5$\pm$4.2 & 22.5$\pm$4.2 & 34.5$\pm$1.6 & 35.0$\pm$1.8 & 35.3$\pm$1.8 \\
 & \ckpt{10} & 21.7$\pm$4.3 & 16.7$\pm$4.7 & 21.7$\pm$6.9 & 21.7$\pm$4.3 & 32.9$\pm$2.6 & 33.7$\pm$1.9 & 33.4$\pm$3.4 \\
 & \ckpt{50} & 13.3$\pm$2.7 & 12.5$\pm$3.2 & 14.2$\pm$1.7 & 13.3$\pm$2.7 & 13.4$\pm$2.2 & 19.2$\pm$4.6 & 16.6$\pm$3.1 \\
 & \ckpt{60} & 1.7$\pm$1.9 & 2.5$\pm$3.2 & 2.5$\pm$3.2 & 2.5$\pm$3.2 & 0.0$\pm$0.0 & 0.0$\pm$0.0 & 0.0$\pm$0.0 \\
 & \ckpt{70} & 0.0$\pm$0.0 & 0.0$\pm$0.0 & 0.0$\pm$0.0 & 0.0$\pm$0.0 & 0.0$\pm$0.0 & 0.0$\pm$0.0 & 0.0$\pm$0.0 \\
\midrule
\multirow{5}{*}{\makecell{Deepseek-R1-Distill\\-Qwen-7B}} & \dns & 42.5$\pm$5.7 & 25.8$\pm$3.2 & 39.2$\pm$4.2 & 42.5$\pm$5.7 & 59.2$\pm$3.8 & 50.3$\pm$2.2 & 53.9$\pm$3.5 \\
 & \ckpt{10} & 31.7$\pm$5.8 & 24.2$\pm$4.2 & 30.8$\pm$3.2 & 31.7$\pm$5.8 & 58.4$\pm$1.8 & 50.0$\pm$2.0 & 53.4$\pm$2.6 \\
 & \ckpt{50} & 30.0$\pm$4.7 & 23.3$\pm$2.7 & 25.8$\pm$5.0 & 30.0$\pm$4.7 & 42.1$\pm$1.7 & 40.5$\pm$1.8 & 40.8$\pm$1.6 \\
 & \ckpt{60} & 22.5$\pm$3.2 & 15.8$\pm$4.2 & 20.8$\pm$1.7 & 22.5$\pm$3.2 & 14.7$\pm$4.3 & 15.8$\pm$3.1 & 16.3$\pm$4.0 \\
 & \ckpt{70} & 5.0$\pm$3.3 & 6.7$\pm$0.0 & 5.8$\pm$1.7 & 5.0$\pm$3.3 & 0.0$\pm$0.0 & 0.0$\pm$0.0 & 0.0$\pm$0.0 \\
\midrule
\multirow{5}{*}{\makecell{Deepseek-R1-Distill\\-Qwen-14B}} & \dns & 54.2$\pm$5.0 & 25.0$\pm$5.8 & 44.2$\pm$3.2 & 55.0$\pm$5.8 & 73.2$\pm$3.9 & 52.9$\pm$1.0 & 63.4$\pm$1.8 \\
 & \ckpt{10} & 47.5$\pm$6.9 & 26.7$\pm$2.7 & 39.2$\pm$3.2 & 49.2$\pm$5.0 & 71.8$\pm$2.3 & 53.9$\pm$3.0 & 63.9$\pm$1.3 \\
 & \ckpt{50} & 35.8$\pm$5.0 & 26.7$\pm$2.7 & 33.3$\pm$4.7 & 36.7$\pm$4.7 & 62.4$\pm$1.0 & 50.0$\pm$4.7 & 56.6$\pm$3.3 \\
 & \ckpt{60} & 26.7$\pm$4.7 & 20.8$\pm$3.2 & 25.0$\pm$3.3 & 27.5$\pm$3.2 & 37.6$\pm$5.6 & 30.5$\pm$1.5 & 38.9$\pm$4.3 \\
 & \ckpt{70} & 3.3$\pm$2.7 & 5.0$\pm$4.3 & 5.8$\pm$4.2 & 5.8$\pm$4.2 & 0.0$\pm$0.0 & 0.0$\pm$0.0 & 0.0$\pm$0.0 \\
\bottomrule
\end{tabular}
}
\end{center}
\caption{Effect of the Forced-Answer budget. Accuracy (\%) after forcing an answer at $D$ reasoning tokens, compared with the unforced checkpoint (\textit{Base}). Generations finishing before $D$ are kept unchanged, and at $D=16{,}384$ only capped generations are forced.}
\label{tab:ablation-forcing}
\end{table*}

\subsection{Ablation Study of No-Repeat} 
\label{app:ablation-no-repeat}

Table~\ref{tab:ablation-ngram} shows that No-Repeat is sensitive to the blocked $n$-gram size and that improved termination does not necessarily translate into accuracy gains. With $n=8$, accuracy drops across nearly all settings, including dense models, with particularly severe degradation on LiveCodeBench: DeepSeek-R1-Distill-Qwen-14B falls from 73.2\% to 19.5\%, and the 7B model from 59.2\% to 15.0\%. Although $n=8$ improves termination for heavily pruned 1.5B models and on LiveCodeBench, accuracy remains near zero in these settings, indicating that most additional terminations are incorrect. It also reduces termination for the 7B model on AIME-25 at every pruned checkpoint, from 43.3\% to 35.8\% at 60\% sparsity. 

These results suggest that blocking short repeated spans can disrupt valid reasoning and code patterns. Increasing the block size to $n=32$ mitigates accuracy degradation and preserves AIME-25 performance for most checkpoints, while improving termination at moderate sparsity, e.g., 39.2\% to 70.0\% for the 14B model at 60\% sparsity. However, these gains yield limited accuracy recovery, and substantial degradation persists on LiveCodeBench, including for dense models. Overall, No-Repeat can suppress repetitive generation but does not reliably guide the model toward correct answers, making it the least effective of the three remedies.

\begin{table*}[t]
\begin{center}
\setlength{\tabcolsep}{2.5pt}
\resizebox{0.8\textwidth}{!}{%
\begin{tabular}{ll rrrrrr rrrrrr}
\toprule
 & & \multicolumn{6}{c}{AIME-25} & \multicolumn{6}{c}{LiveCodeBench} \\
\cmidrule(lr){3-8}\cmidrule(lr){9-14}
 & & \multicolumn{2}{c}{Base} & \multicolumn{2}{c}{$n$=8} & \multicolumn{2}{c}{$n$=32} & \multicolumn{2}{c}{Base} & \multicolumn{2}{c}{$n$=8} & \multicolumn{2}{c}{$n$=32} \\
\cmidrule(lr){3-4}\cmidrule(lr){5-6}\cmidrule(lr){7-8}\cmidrule(lr){9-10}\cmidrule(lr){11-12}\cmidrule(lr){13-14}
\thd{Model} & \thd{Ckpt} & \thd{Acc} & \thd{Term} & \thd{Acc} & \thd{Term} & \thd{Acc} & \thd{Term} & \thd{Acc} & \thd{Term} & \thd{Acc} & \thd{Term} & \thd{Acc} & \thd{Term} \\
\midrule
\multirow{4}{*}{\makecell{DS-Qwen\\-1.5B}} & \dns & 22.5$\pm$4.2 & 71.7$\pm$5.8 & 7.5$\pm$4.2 & 68.3$\pm$7.9 & 22.5$\pm$4.2 & 74.2$\pm$7.4 & 34.5$\pm$1.6 & 76.1$\pm$1.8 & 8.9$\pm$1.4 & 79.7$\pm$1.0 & 27.6$\pm$1.6 & 80.8$\pm$3.9 \\
 & \ckpt{10} & 21.7$\pm$4.3 & 71.7$\pm$1.9 & 6.7$\pm$2.7 & 66.7$\pm$4.7 & 24.2$\pm$5.7 & 70.0$\pm$8.6 & 32.9$\pm$2.6 & 76.6$\pm$3.1 & 8.9$\pm$2.6 & 77.9$\pm$4.8 & 30.5$\pm$2.3 & 82.4$\pm$3.0 \\
 & \ckpt{50} & 13.3$\pm$2.7 & 46.7$\pm$4.7 & 3.3$\pm$0.0 & 46.7$\pm$9.8 & 17.5$\pm$5.7 & 51.7$\pm$7.9 & 13.4$\pm$2.2 & 32.1$\pm$4.3 & 1.8$\pm$1.0 & 40.3$\pm$5.0 & 9.5$\pm$2.6 & 40.8$\pm$6.2 \\
 & \ckpt{60} & 1.7$\pm$1.9 & 6.7$\pm$4.7 & 0.0$\pm$0.0 & 42.5$\pm$13.4 & 2.5$\pm$5.0 & 23.3$\pm$4.7 & 0.0$\pm$0.0 & 0.5$\pm$0.6 & 0.0$\pm$0.0 & 31.6$\pm$3.6 & 0.0$\pm$0.0 & 7.4$\pm$0.9 \\
\midrule
\multirow{5}{*}{\makecell{DS-Qwen\\-7B}} & \dns & 42.5$\pm$5.7 & 80.0$\pm$7.2 & 23.3$\pm$7.2 & 76.7$\pm$8.2 & 34.2$\pm$5.7 & 77.5$\pm$5.0 & 59.2$\pm$3.8 & 83.7$\pm$1.4 & 15.0$\pm$1.6 & 82.1$\pm$2.3 & 45.0$\pm$1.8 & 82.4$\pm$3.4 \\
 & \ckpt{10} & 31.7$\pm$5.8 & 80.0$\pm$7.2 & 29.2$\pm$4.2 & 79.2$\pm$3.2 & 35.0$\pm$3.3 & 78.3$\pm$4.3 & 58.4$\pm$1.8 & 82.6$\pm$2.6 & 17.1$\pm$4.1 & 84.5$\pm$2.8 & 49.5$\pm$2.3 & 84.5$\pm$1.6 \\
 & \ckpt{50} & 30.0$\pm$4.7 & 57.5$\pm$7.4 & 20.8$\pm$3.2 & 48.3$\pm$6.9 & 31.7$\pm$8.8 & 65.8$\pm$3.2 & 42.1$\pm$1.7 & 63.9$\pm$3.1 & 10.0$\pm$2.2 & 67.6$\pm$3.1 & 33.7$\pm$2.7 & 66.1$\pm$5.0 \\
 & \ckpt{60} & 22.5$\pm$3.2 & 43.3$\pm$8.6 & 12.5$\pm$5.0 & 35.8$\pm$3.2 & 22.5$\pm$4.2 & 48.3$\pm$4.3 & 14.7$\pm$4.3 & 34.2$\pm$5.0 & 4.2$\pm$1.7 & 40.5$\pm$5.0 & 14.7$\pm$1.5 & 50.0$\pm$2.5 \\
 & \ckpt{70} & 5.0$\pm$3.3 & 19.2$\pm$3.2 & 0.0$\pm$0.0 & 2.5$\pm$3.2 & 4.2$\pm$1.7 & 15.0$\pm$3.3 & 0.0$\pm$0.0 & 0.0$\pm$0.0 & 0.0$\pm$0.0 & 3.9$\pm$1.6 & 0.0$\pm$0.0 & 1.6$\pm$1.4 \\
\midrule
\multirow{5}{*}{\makecell{DS-Qwen\\-14B}} & \dns & 54.2$\pm$5.0 & 87.5$\pm$5.0 & 33.3$\pm$4.7 & 82.5$\pm$5.7 & 45.8$\pm$11.3 & 80.8$\pm$8.3 & 73.2$\pm$3.9 & 90.8$\pm$2.2 & 19.5$\pm$4.4 & 93.2$\pm$2.5 & 56.8$\pm$1.5 & 89.5$\pm$1.9 \\
 & \ckpt{10} & 47.5$\pm$6.9 & 84.2$\pm$5.0 & 40.8$\pm$6.3 & 87.5$\pm$7.4 & 47.5$\pm$10.0 & 83.3$\pm$2.7 & 71.8$\pm$2.3 & 89.2$\pm$2.3 & 18.7$\pm$3.7 & 94.2$\pm$1.4 & 56.1$\pm$1.8 & 88.9$\pm$2.2 \\
 & \ckpt{50} & 35.8$\pm$5.0 & 63.3$\pm$6.1 & 28.3$\pm$4.3 & 66.7$\pm$11.9 & 37.5$\pm$3.2 & 64.2$\pm$6.3 & 62.4$\pm$1.0 & 75.8$\pm$2.3 & 11.6$\pm$3.3 & 83.9$\pm$2.8 & 46.6$\pm$2.9 & 80.3$\pm$3.6 \\
 & \ckpt{60} & 26.7$\pm$4.7 & 39.2$\pm$5.0 & 17.5$\pm$1.7 & 45.0$\pm$5.8 & 25.0$\pm$3.3 & 70.0$\pm$10.5 & 37.6$\pm$5.6 & 46.6$\pm$3.9 & 9.5$\pm$1.9 & 63.9$\pm$4.6 & 33.9$\pm$1.0 & 60.3$\pm$3.0 \\
 & \ckpt{70} & 3.3$\pm$2.7 & 10.8$\pm$3.2 & 1.7$\pm$1.9 & 10.0$\pm$4.7 & 5.0$\pm$5.8 & 24.2$\pm$10.7 & 0.0$\pm$0.0 & 0.0$\pm$0.0 & 0.0$\pm$0.0 & 7.1$\pm$2.3 & 0.0$\pm$0.0 & 1.8$\pm$1.3 \\
\bottomrule
\end{tabular}}
\end{center}
\caption{No-Repeat $n$-gram size. \thd{Acc} is accuracy (\%) and \thd{Term} the share of generations that terminated (\%). All three columns of a dataset come from the same evaluation pass, so \textit{Base} here is that pass's unblocked run.}
\label{tab:ablation-ngram}
\end{table*}

\subsection{Ablation Study of Steering Strength} 

Table~\ref{tab:ablation-alpha} shows that the steering strength $\alpha$ provides direct control over termination, with stronger effects under higher sparsity. For DeepSeek-R1-Distill-Qwen-7B at 60\% sparsity, LiveCodeBench termination increases from 1.8\% at $\alpha=-2$ to 34.2\% without steering and 57.9\% at $\alpha=+2$. On AIME-25, it similarly increases from 13.3\% to 43.3\% and 55.0\%, respectively. The effect is smaller for dense and lightly pruned checkpoints, which already terminate reliably. Reversing the steering direction is particularly harmful: at 60\% sparsity, LiveCodeBench accuracy drops from 14.7\% to 1.1\% when $\alpha$ changes from 0 to $-2$. These results indicate that the steering direction directly affects the model's stopping behavior.

However, increasing $\alpha$ too strong beyond zero primarily improves termination rather than accuracy. For DeepSeek-R1-Distill-Qwen-7B, positive $\alpha$ values produce little change in LiveCodeBench accuracy and even reduce the accuracy of AIME-25 at 60\% sparsity: increasing the value of $\alpha$ from 1 to 2 reduces accuracy from 22.5\% to 20\%. OpenThinker3-7B exhibits a similar but weaker trend, with termination increasing modestly while accuracy changes only slightly. Overall, $\alpha$ is an effective control for termination, stronger steering mainly improves stopping behavior rather than solution accuracy, while reversing the steering direction can substantially degrade both.

\begin{table*}[t]
\begin{center}
\setlength{\tabcolsep}{4pt}
\resizebox{\textwidth}{!}{%
\begin{tabular}{lll rr rr rr rr rr rr rr}
\toprule
 & & & \multicolumn{2}{c}{$\alpha=-2$} & \multicolumn{2}{c}{$\alpha=-1.5$} & \multicolumn{2}{c}{$\alpha=-1$} & \multicolumn{2}{c}{$\alpha=0$} & \multicolumn{2}{c}{$\alpha=+1$} & \multicolumn{2}{c}{$\alpha=+1.5$} & \multicolumn{2}{c}{$\alpha=+2$} \\
\cmidrule(lr){4-5}\cmidrule(lr){6-7}\cmidrule(lr){8-9}\cmidrule(lr){10-11}\cmidrule(lr){12-13}\cmidrule(lr){14-15}\cmidrule(lr){16-17}
\thd{Model} & \thd{Test} & \thd{Ckpt} & \thd{Acc} & \thd{Term} & \thd{Acc} & \thd{Term} & \thd{Acc} & \thd{Term} & \thd{Acc} & \thd{Term} & \thd{Acc} & \thd{Term} & \thd{Acc} & \thd{Term} & \thd{Acc} & \thd{Term} \\
\midrule
\multirow{8}{*}{\makecell{Deepseek-R1-Distill\\-Qwen-7B}} & \multirow{4}{*}{AIME} & \dns & 31.7$\pm$4.3 & 70.0$\pm$6.1 & 37.5$\pm$5.0 & 83.3$\pm$7.2 & 37.5$\pm$4.2 & 81.7$\pm$1.9 & 42.5$\pm$5.7 & 80.0$\pm$7.2 & 41.7$\pm$6.9 & 79.2$\pm$8.8 & 42.5$\pm$6.9 & 82.5$\pm$3.2 & 36.7$\pm$4.2 & 83.3$\pm$5.0 \\ 
 &  & \ckpt{10} & 32.5$\pm$3.2 & 70.0$\pm$4.7 & 37.5$\pm$9.2 & 76.7$\pm$3.8 & 37.5$\pm$4.2 & 80.0$\pm$6.1 & 31.7$\pm$5.8 & 80.0$\pm$7.2 & 40.0$\pm$6.1 & 76.7$\pm$8.2 & 43.3$\pm$3.2 & 85.0$\pm$9.6 & 36.7$\pm$2.7 & 88.3$\pm$7.2 \\
 &  & \ckpt{50} & 27.5$\pm$6.3 & 33.3$\pm$7.2 & 25.8$\pm$5.7 & 45.8$\pm$10.7 & 30.8$\pm$6.9 & 55.0$\pm$4.3 & 30.0$\pm$4.7 & 57.5$\pm$7.4 & 30.0$\pm$4.7 & 63.3$\pm$4.7 & 25.8$\pm$4.2 & 64.2$\pm$4.2 & 25.8$\pm$3.2 & 72.5$\pm$4.3 \\
 &  & \ckpt{60} & 10.0$\pm$2.7 & 13.3$\pm$4.7 & 14.2$\pm$4.2 & 16.7$\pm$4.7 & 15.8$\pm$5.7 & 25.8$\pm$3.2 & 22.5$\pm$6.3 & 43.3$\pm$8.6 & 22.5$\pm$1.7 & 49.2$\pm$4.2 & 20.8$\pm$7.4 & 55.0$\pm$3.3 & 20.0$\pm$3.2 & 63.3$\pm$8.2 \\
\cmidrule(lr){2-17}
 & \multirow{4}{*}{LCB} & \dns & 55.3$\pm$2.2 & 73.4$\pm$2.0 & 57.4$\pm$2.2 & 80.5$\pm$3.7 & 55.8$\pm$2.7 & 81.1$\pm$1.5 & 59.2$\pm$3.8 & 83.7$\pm$1.4 & 59.7$\pm$2.2 & 89.2$\pm$3.0 & 59.7$\pm$1.3 & 90.0$\pm$3.3 & 56.1$\pm$1.4 & 91.3$\pm$1.3 \\
 &  & \ckpt{10} & 56.1$\pm$1.8 & 75.5$\pm$3.6 & 55.8$\pm$1.5 & 78.7$\pm$2.3 & 56.3$\pm$1.1 & 78.7$\pm$3.8 & 58.4$\pm$1.8 & 82.6$\pm$2.6 & 58.7$\pm$2.5 & 86.6$\pm$1.6 & 57.6$\pm$2.3 & 89.5$\pm$2.3 & 56.8$\pm$1.8 & 92.6$\pm$1.3 \\
 &  & \ckpt{50} & 27.1$\pm$4.0 & 35.0$\pm$4.2 & 32.1$\pm$2.5 & 44.7$\pm$2.0 & 37.6$\pm$2.6 & 53.7$\pm$2.3 & 42.1$\pm$1.7 & 63.9$\pm$3.1 & 38.2$\pm$4.3 & 67.4$\pm$2.3 & 42.1$\pm$0.9 & 76.3$\pm$3.3 & 37.1$\pm$3.5 & 80.8$\pm$1.8 \\
 &  & \ckpt{60} & 1.1$\pm$0.9 & 1.8$\pm$0.5 & 3.7$\pm$1.1 & 7.1$\pm$0.5 & 8.2$\pm$2.8 & 14.7$\pm$0.0 & 14.7$\pm$4.3 & 34.2$\pm$5.0 & 20.9$\pm$2.2 & 49.5$\pm$3.1 & 16.8$\pm$3.7 & 53.7$\pm$4.5 & 16.3$\pm$2.3 & 57.9$\pm$2.7 \\
\midrule
\multirow{8}{*}{\makecell{OpenThinker3\\-7B}} & \multirow{4}{*}{AIME} & \dns & 47.5$\pm$5.7 & 49.2$\pm$4.2 & 45.8$\pm$6.9 & 49.2$\pm$5.0 & 42.5$\pm$1.7 & 45.8$\pm$4.2 & 44.2$\pm$5.0 & 50.0$\pm$4.7 & 42.5$\pm$4.2 & 47.5$\pm$3.2 & 45.8$\pm$4.2 & 54.2$\pm$3.2 & 43.3$\pm$5.4 & 50.0$\pm$5.4 \\
 &  & \ckpt{10} & 41.7$\pm$4.3 & 43.3$\pm$4.7 & 47.5$\pm$3.2 & 48.3$\pm$4.3 & 42.5$\pm$4.2 & 47.5$\pm$6.3 & 50.0$\pm$0.0 & 55.0$\pm$1.9 & 48.3$\pm$7.9 & 55.0$\pm$6.9 & 48.3$\pm$5.8 & 56.7$\pm$5.4 & 45.8$\pm$4.2 & 52.5$\pm$5.0 \\
 &  & \ckpt{50} & 35.0$\pm$3.3 & 40.8$\pm$3.2 & 30.0$\pm$9.4 & 41.7$\pm$10.4 & 31.7$\pm$4.3 & 40.0$\pm$6.1 & 37.5$\pm$4.2 & 50.8$\pm$6.9 & 37.5$\pm$6.3 & 48.3$\pm$7.9 & 39.2$\pm$7.4 & 53.3$\pm$8.6 & 39.2$\pm$10.0 & 50.0$\pm$7.2 \\
 &  & \ckpt{60} & 19.2$\pm$7.4 & 35.8$\pm$8.8 & 24.2$\pm$4.2 & 42.5$\pm$3.2 & 23.3$\pm$3.8 & 39.2$\pm$6.9 & 24.2$\pm$5.7 & 47.5$\pm$6.3 & 24.2$\pm$5.0 & 50.0$\pm$10.5 & 29.2$\pm$1.7 & 50.8$\pm$6.9 & 22.5$\pm$5.0 & 51.7$\pm$6.4 \\
\cmidrule(lr){2-17}
 & \multirow{4}{*}{LCB} & \dns & 60.8$\pm$2.3 & 69.5$\pm$0.9 & 61.3$\pm$2.3 & 71.3$\pm$3.4 & 61.6$\pm$3.0 & 71.8$\pm$4.4 & 63.4$\pm$2.0 & 72.9$\pm$1.8 & 62.1$\pm$1.5 & 74.2$\pm$0.6 & 64.7$\pm$1.4 & 73.9$\pm$1.8 & 61.3$\pm$2.3 & 72.1$\pm$2.8 \\
 &  & \ckpt{10} & 61.8$\pm$1.3 & 68.9$\pm$1.4 & 61.6$\pm$1.4 & 70.8$\pm$3.6 & 61.8$\pm$1.0 & 69.7$\pm$2.2 & 61.3$\pm$2.2 & 71.8$\pm$1.3 & 60.5$\pm$0.6 & 71.1$\pm$1.1 & 63.4$\pm$0.5 & 74.2$\pm$1.8 & 63.2$\pm$1.9 & 75.0$\pm$3.4 \\
 &  & \ckpt{50} & 52.6$\pm$0.9 & 64.7$\pm$2.2 & 52.4$\pm$2.3 & 63.7$\pm$5.3 & 53.2$\pm$2.2 & 64.2$\pm$2.3 & 51.1$\pm$1.1 & 63.2$\pm$2.6 & 51.8$\pm$1.8 & 66.8$\pm$1.4 & 52.4$\pm$3.5 & 67.1$\pm$3.7 & 53.4$\pm$1.0 & 67.6$\pm$2.3 \\
 &  & \ckpt{60} & 24.7$\pm$2.5 & 49.5$\pm$2.1 & 28.4$\pm$2.3 & 54.2$\pm$3.9 & 30.0$\pm$1.4 & 54.7$\pm$2.9 & 31.3$\pm$2.6 & 54.2$\pm$1.8 & 32.6$\pm$3.1 & 58.9$\pm$3.7 & 36.3$\pm$3.0 & 58.9$\pm$3.1 & 38.2$\pm$2.2 & 60.0$\pm$3.5 \\
\bottomrule
\end{tabular}}
\end{center}
\caption{Effect of steering strength $\alpha$ on accuracy and termination. Each model is steered with its main direction scaled by $\alpha$, absolute steering coefficient; $\alpha=0$ is the unsteered checkpoint and negative $\alpha$ pushes against the direction. \thd{Acc} is accuracy (\%) and \thd{Term} the share of generations that terminated (\%).}
\label{tab:ablation-alpha}
\end{table*}

\subsection{Ablation on Number of Sampled Created Directions}
\label{app:number_sample_create_direction}

Table~\ref{tab:ablation-sample-size} shows that a small number of samples is sufficient to obtain an effective steering direction. Using 50 samples per class achieves comparable steering performance to 100 samples or the full set for both DeepSeek-R1-Distill-Qwen-7B and OpenThinker3-7B. In particular, the termination rate remains stable across sample sizes: for DeepSeek-R1-Distill-Qwen-7B at 60\% sparsity on LiveCodeBench, it stays within 52.9\%-57.6\%, compared with 34.2\% without steering. These results suggest that 50 samples per class provide sufficient signal to identify an effective steering direction.

\begin{table*}[t]
\begin{center}
\tabsmall
\begin{tabular}{lll rr rr rr rr}
\toprule
 & & & \multicolumn{2}{c}{n50} & \multicolumn{2}{c}{n75} & \multicolumn{2}{c}{n100} & \multicolumn{2}{c}{Full} \\
\cmidrule(lr){4-5}\cmidrule(lr){6-7}\cmidrule(lr){8-9}\cmidrule(lr){10-11}
\thd{Model} & \thd{Test} & \thd{Ckpt} & \thd{Acc} & \thd{Term} & \thd{Acc} & \thd{Term} & \thd{Acc} & \thd{Term} & \thd{Acc} & \thd{Term} \\
\midrule
\multirow{8}{*}{\makecell{Deepseek-R1-Distill\\-Qwen-7B}} & \multirow{4}{*}{AIME} & \dns & 38.3$\pm$6.4 & 82.5$\pm$9.2 & 36.7$\pm$9.0 & 79.2$\pm$5.7 & 42.5$\pm$4.2 & 80.0$\pm$3.8 & 41.7$\pm$6.9 & 79.2$\pm$8.8 \\
 &  & \ckpt{10} & 35.8$\pm$3.2 & 83.3$\pm$8.6 & 38.3$\pm$4.3 & 80.8$\pm$3.2 & 35.8$\pm$1.7 & 80.8$\pm$3.2 & 43.3$\pm$4.7 & 80.8$\pm$1.7 \\
 &  & \ckpt{50} & 31.7$\pm$4.3 & 69.2$\pm$3.2 & 28.3$\pm$6.4 & 62.5$\pm$5.0 & 30.8$\pm$8.8 & 69.2$\pm$4.2 & 31.7$\pm$3.3 & 68.3$\pm$6.9 \\
 &  & \ckpt{60} & 18.3$\pm$4.3 & 55.0$\pm$5.8 & 20.0$\pm$3.8 & 57.5$\pm$3.2 & 25.0$\pm$6.4 & 62.5$\pm$3.2 & 22.5$\pm$3.2 & 54.2$\pm$3.2 \\
\cmidrule(lr){2-11}
 & \multirow{4}{*}{LCB} & \dns & 58.2$\pm$2.3 & 87.4$\pm$3.1 & 57.9$\pm$2.3 & 86.3$\pm$0.9 & 57.9$\pm$2.3 & 88.4$\pm$2.6 & 59.7$\pm$2.2 & 89.2$\pm$3.0 \\
 &  & \ckpt{10} & 57.9$\pm$1.5 & 87.1$\pm$4.1 & 53.2$\pm$2.0 & 87.9$\pm$3.0 & 56.6$\pm$2.0 & 89.7$\pm$1.8 & 60.3$\pm$3.4 & 90.8$\pm$3.8 \\
 &  & \ckpt{50} & 38.9$\pm$5.8 & 73.2$\pm$3.9 & 40.8$\pm$2.3 & 74.5$\pm$1.8 & 37.9$\pm$2.6 & 73.9$\pm$3.8 & 41.8$\pm$2.3 & 66.6$\pm$2.2 \\
 &  & \ckpt{60} & 16.3$\pm$2.2 & 57.6$\pm$3.1 & 16.1$\pm$2.2 & 53.9$\pm$5.0 & 16.3$\pm$1.4 & 52.9$\pm$3.0 & 19.5$\pm$3.3 & 50.5$\pm$3.0 \\
\midrule
\multirow{8}{*}{\makecell{OpenThinker3\\-7B}} & \multirow{4}{*}{AIME} & \dns & 41.7$\pm$3.3 & 49.2$\pm$3.2 & 44.2$\pm$6.3 & 50.0$\pm$4.7 & 43.3$\pm$4.7 & 48.3$\pm$1.9 & 45.8$\pm$4.2 & 54.2$\pm$3.2 \\
 &  & \ckpt{10} & 45.0$\pm$6.4 & 51.7$\pm$6.9 & 42.5$\pm$5.7 & 47.5$\pm$1.7 & 43.3$\pm$7.2 & 56.7$\pm$5.4 & 48.3$\pm$5.8 & 56.7$\pm$5.4 \\
 &  & \ckpt{50} & 37.5$\pm$1.7 & 50.8$\pm$1.7 & 39.2$\pm$5.0 & 55.8$\pm$10.0 & 35.0$\pm$3.3 & 46.7$\pm$2.7 & 43.3$\pm$8.2 & 52.5$\pm$9.6 \\
 &  & \ckpt{60} & 25.8$\pm$5.7 & 54.2$\pm$4.2 & 23.3$\pm$3.8 & 49.2$\pm$10.0 & 25.8$\pm$4.2 & 52.5$\pm$3.2 & 29.2$\pm$1.7 & 50.8$\pm$6.9 \\
\cmidrule(lr){2-11}
 & \multirow{4}{*}{LCB} & \dns & 63.7$\pm$1.1 & 73.7$\pm$3.3 & 61.6$\pm$2.0 & 72.9$\pm$3.5 & 62.4$\pm$1.0 & 73.9$\pm$3.3 & 64.7$\pm$1.4 & 73.9$\pm$1.8 \\
 &  & \ckpt{10} & 62.4$\pm$5.3 & 73.7$\pm$4.0 & 64.7$\pm$2.0 & 75.0$\pm$3.8 & 63.7$\pm$1.8 & 72.9$\pm$2.5 & 63.4$\pm$0.5 & 74.2$\pm$1.8 \\
 &  & \ckpt{50} & 50.0$\pm$2.8 & 64.7$\pm$1.4 & 50.0$\pm$0.6 & 65.5$\pm$1.3 & 54.2$\pm$1.4 & 68.7$\pm$1.8 & 55.5$\pm$2.2 & 69.5$\pm$3.0 \\
 &  & \ckpt{60} & 32.6$\pm$2.9 & 60.8$\pm$3.6 & 33.4$\pm$3.8 & 58.2$\pm$2.2 & 32.1$\pm$1.8 & 57.1$\pm$2.2 & 36.3$\pm$3.0 & 58.9$\pm$3.1 \\
\bottomrule
\end{tabular}
\end{center}
\caption{TGAS with directions built from different numbers of samples per class. \thd{Acc} is accuracy (\%) and \thd{Term} the share of generations that terminated (\%). \thd{n50}, \thd{n75} and \thd{n100} use 50, 75 and 100 samples per class; \thd{Full} is the direction used in the main experiments. We use the same $\alpha$ values as in the main experiments for each configuration.}
\label{tab:ablation-sample-size}
\end{table*}

\subsection{Generalization Across Pruning Methods}
\label{app:different_pruning_method}

Table~\ref{tab:ablation-pruners} shows that the termination failure is not specific to one pruning method. Under both RAC (SparseGPT) and C4-calibrated Wanda, the termination rate of Base drops monotonically as sparsity increases, for every model and benchmark. For DeepSeek-R1-Distill-Qwen-7B pruned with RAC, AIME-25 termination falls from 79.2\% at 10\% sparsity to 32.5\% at 50\% and 1.7\% at 60\%, and LiveCodeBench termination falls from 85.0\% to 37.4\% and 2.4\%. The same trend holds for DeepSeek-R1-Distill-Qwen-14B with RAC and for Wanda-pruned checkpoints. Accuracy degrades alongside termination in all cases. 

TGAS remains effective under both pruners: it raises termination over Base on 16 of 18 checkpoint-benchmark pairs and improves accuracy on all 18, achieving the best accuracy on 15. The gains are largest where termination has collapsed, e.g., on DeepSeek-R1-Distill-Qwen-7B with RAC at 60\% sparsity, LiveCodeBench termination rises from 2.4\% to 21.8\% and accuracy from 2.4\% to 13.4\%, and on DeepSeek-R1-Distill-Qwen-14B with RAC at 60\% sparsity, AIME-25 termination rises from 21.7\% to 47.5\% and LiveCodeBench accuracy from 6.1\% to 18.2\%. Under Wanda, TGAS improves DeepSeek-R1-Distill-Qwen-7B at 50\% sparsity from 47.5\% to 64.2\% termination on AIME-25 and from 31.8\% to 39.2\% accuracy on LiveCodeBench. These results indicate that TGAS recovers part of the accuracy lost to pruning by restoring termination, independent of the pruning method used.

\begin{table}[ht]
\centering
\scriptsize
\setlength{\tabcolsep}{4pt}
\begin{tabular}{ll rr}
\toprule
\thd{Model} & \thd{SEAL} & \thd{Layer mean} & \thd{Unit-avg} \\
\midrule
\multirow{2}{*}{DS-Qwen-1.5B} & dense     & 0.172 & 0.108 \\
                              & \ckpt{50} & 0.245 & 0.269 \\
\multirow{2}{*}{DS-Qwen-7B}   & dense     & 0.283 & 0.187 \\
                              & \ckpt{50} & 0.306 & 0.265 \\
\multirow{2}{*}{DS-Qwen-14B}  & dense     & 0.269 & 0.221 \\
                              & \ckpt{50} & 0.244 & 0.214 \\
DS-Llama-8B                   & dense     & 0.163 & 0.153 \\
OT3-7B                        & dense     & 0.177 & 0.180 \\
OT3-1.5B                      & dense     & 0.146 & 0.157 \\
\bottomrule
\end{tabular}
\caption{Cosine between the steering directions of SEAL and TGAS, averaged over decoder layers. \thd{Layer mean}: cosine of the unnormalised layer averages; \thd{Unit-avg}: each layer normalised before averaging.}
\label{tab:seal-tgas}
\end{table}

\begin{table*}[t]
\scriptsize
\begin{center}
\begin{tabular}{llll rr rr}
\toprule
 & & & & \multicolumn{2}{c}{AIME-25} & \multicolumn{2}{c}{LiveCodeBench} \\
\cmidrule(lr){5-6}\cmidrule(lr){7-8}
\thd{Model} & \thd{Pruner} & \thd{Ckpt} & \thd{Method} & \thd{Acc} & \thd{Term} & \thd{Acc} & \thd{Term} \\
\midrule
\multirow{12}{*}{\makecell{Deepseek-R1-Distill\\-Qwen-7B}} & \multirow{12}{*}{RAC} & \multirow{4}{*}{10\%} & Base & 35.8$\pm$5.0 & 79.2$\pm$4.2 & 56.3$\pm$1.8 & 85.0$\pm$3.8 \\
 &  &  & No-Repeat & 38.3$\pm$5.8 & 80.0$\pm$5.4 & 47.4$\pm$3.1 & 85.8$\pm$1.1 \\
 &  &  & Forced-Answer & 33.3$\pm$4.7 & --- & 53.7$\pm$1.2 & --- \\
 &  &  & TGAS & \textbf{39.2$\pm$3.2} & 85.0$\pm$6.9 & \textbf{58.7$\pm$2.6} & 89.7$\pm$3.0 \\
\cmidrule(lr){3-8}
 &  & \multirow{4}{*}{50\%} & Base & 24.2$\pm$5.7 & 32.5$\pm$5.7 & 32.6$\pm$1.7 & 37.4$\pm$2.2 \\
 &  &  & No-Repeat & 25.0$\pm$3.3 & 79.2$\pm$7.9 & 27.9$\pm$2.8 & 80.0$\pm$3.6 \\
 &  &  & Forced-Answer & 23.3$\pm$3.8 & --- & \textbf{37.1$\pm$1.8} & --- \\
 &  &  & TGAS & \textbf{25.8$\pm$1.7} & 59.2$\pm$7.4 & 34.5$\pm$4.0 & 47.9$\pm$1.1 \\
\cmidrule(lr){3-8}
 &  & \multirow{4}{*}{60\%} & Base & 0.0$\pm$0.0 & 1.7$\pm$1.9 & 2.4$\pm$0.5 & 2.4$\pm$0.5 \\
 &  &  & No-Repeat & 5.8$\pm$3.2 & 46.7$\pm$11.5 & 10.5$\pm$2.3 & 58.9$\pm$3.7 \\
 &  &  & Forced-Answer & \textbf{10.0$\pm$4.7} & --- & 6.1$\pm$2.3 & --- \\
 &  &  & TGAS & 6.7$\pm$5.4 & 10.8$\pm$7.9 & \textbf{13.4$\pm$2.3} & 21.8$\pm$2.3 \\
\midrule
\multirow{8}{*}{\makecell{Deepseek-R1-Distill\\-Qwen-14B}} & \multirow{8}{*}{RAC} & \multirow{4}{*}{50\%} & Base & 26.7$\pm$2.7 & 70.0$\pm$4.7 & 48.7$\pm$2.0 & 66.1$\pm$1.8 \\
 &  &  & No-Repeat & 28.3$\pm$3.3 & 70.8$\pm$6.9 & 42.6$\pm$2.8 & 75.0$\pm$1.8 \\
 &  &  & Forced-Answer & 24.2$\pm$4.2 & --- & 47.6$\pm$1.0 & --- \\
 &  &  & TGAS & \textbf{30.0$\pm$2.7} & 68.3$\pm$4.3 & \textbf{55.8$\pm$4.1} & 68.7$\pm$3.3 \\
\cmidrule(lr){3-8}
 &  & \multirow{4}{*}{60\%} & Base & 3.3$\pm$2.7 & 21.7$\pm$6.4 & 6.1$\pm$1.0 & 8.2$\pm$1.3 \\
 &  &  & No-Repeat & 3.3$\pm$2.7 & 45.0$\pm$7.9 & 18.7$\pm$2.3 & 61.8$\pm$2.9 \\
 &  &  & Forced-Answer & 3.3$\pm$2.7 & --- & \textbf{21.1$\pm$1.9} & --- \\
 &  &  & TGAS & \textbf{5.8$\pm$6.9} & 47.5$\pm$8.8 & 18.2$\pm$2.3 & 26.8$\pm$1.4 \\
\midrule
\multirow{12}{*}{\makecell{Deepseek-R1-Distill\\-Qwen-7B}} & \multirow{12}{*}{C4 Wanda} & \multirow{4}{*}{40\%} & Base & 34.2$\pm$3.2 & 67.5$\pm$6.9 & 49.5$\pm$1.7 & 73.7$\pm$3.5 \\
 &  &  & No-Repeat & 31.7$\pm$4.3 & 75.8$\pm$3.2 & 41.6$\pm$4.6 & 77.9$\pm$1.2 \\
 &  &  & Forced-Answer & 30.0$\pm$4.7 & --- & 49.7$\pm$2.0 & --- \\
 &  &  & TGAS & \textbf{35.0$\pm$1.9} & 75.8$\pm$5.0 & \textbf{52.1$\pm$0.6} & 81.6$\pm$4.7 \\
\cmidrule(lr){3-8}
 &  & \multirow{4}{*}{50\%} & Base & 19.2$\pm$1.7 & 47.5$\pm$9.2 & 31.8$\pm$5.3 & 52.9$\pm$5.6 \\
 &  &  & No-Repeat & 15.8$\pm$3.2 & 61.7$\pm$1.9 & 26.6$\pm$4.7 & 67.4$\pm$2.4 \\
 &  &  & Forced-Answer & 19.2$\pm$1.7 & --- & 34.7$\pm$4.3 & --- \\
 &  &  & TGAS & \textbf{21.7$\pm$4.3} & 64.2$\pm$7.4 & \textbf{39.2$\pm$4.0} & 58.7$\pm$3.6 \\
\cmidrule(lr){3-8}
 &  & \multirow{4}{*}{60\%} & Base & 0.0$\pm$0.0 & 0.0$\pm$0.0 & 0.3$\pm$0.5 & 1.6$\pm$1.1 \\
 &  &  & No-Repeat & 0.0$\pm$0.0 & 9.2$\pm$3.2 & 0.5$\pm$0.6 & 11.1$\pm$1.4 \\
 &  &  & Forced-Answer & 0.0$\pm$0.0 & --- & 0.3$\pm$0.5 & --- \\
 &  &  & TGAS & \textbf{1.7$\pm$1.9} & 1.7$\pm$3.3 & \textbf{2.1$\pm$0.9} & 14.5$\pm$1.6 \\
\midrule
\multirow{8}{*}{\makecell{OpenThinker3\\-7B}} & \multirow{8}{*}{C4 Wanda} & \multirow{4}{*}{40\%} & Base & 37.5$\pm$5.7 & 53.3$\pm$8.2 & 50.5$\pm$1.9 & 64.2$\pm$2.9 \\
 &  &  & No-Repeat & 39.2$\pm$6.9 & 58.3$\pm$6.9 & 45.5$\pm$4.1 & 65.3$\pm$2.1 \\
 &  &  & Forced-Answer & 30.8$\pm$3.2 & --- & 46.6$\pm$2.0 & --- \\
 &  &  & TGAS & \textbf{40.0$\pm$2.7} & 53.3$\pm$2.7 & \textbf{55.0$\pm$2.0} & 67.1$\pm$1.8 \\
\cmidrule(lr){3-8}
 &  & \multirow{4}{*}{50\%} & Base & 17.5$\pm$3.2 & 40.8$\pm$5.7 & 29.7$\pm$2.0 & 58.9$\pm$0.9 \\
 &  &  & No-Repeat & 18.3$\pm$4.3 & 44.2$\pm$5.7 & 24.2$\pm$4.8 & 54.7$\pm$4.5 \\
 &  &  & Forced-Answer & 15.0$\pm$4.3 & --- & 31.3$\pm$3.8 & --- \\
 &  &  & TGAS & \textbf{20.0$\pm$3.8} & 41.7$\pm$6.4 & \textbf{31.6$\pm$3.6} & 60.5$\pm$3.9 \\
\bottomrule
\end{tabular}
\end{center}
\caption{Training-free remedies on checkpoints pruned with SparseGPT-pruned LRMs with RAC calibration and with C4 calibration (Wanda prune). \thd{Acc} is accuracy (\%) and \thd{Term} the share of generations that terminated (\%). Forced-Answer's \thd{Term} is not reported, since forcing ends every generation by design. Best accuracy per checkpoint and dataset in bold.}
\label{tab:ablation-pruners}
\end{table*}

\subsection{Cross-Testing with Different Directions}
\label{app:different-direction}

\begin{figure*}[t]
\centering
\begin{minipage}[c]{0.60\textwidth}
    \centering
    \includegraphics[width=\linewidth]{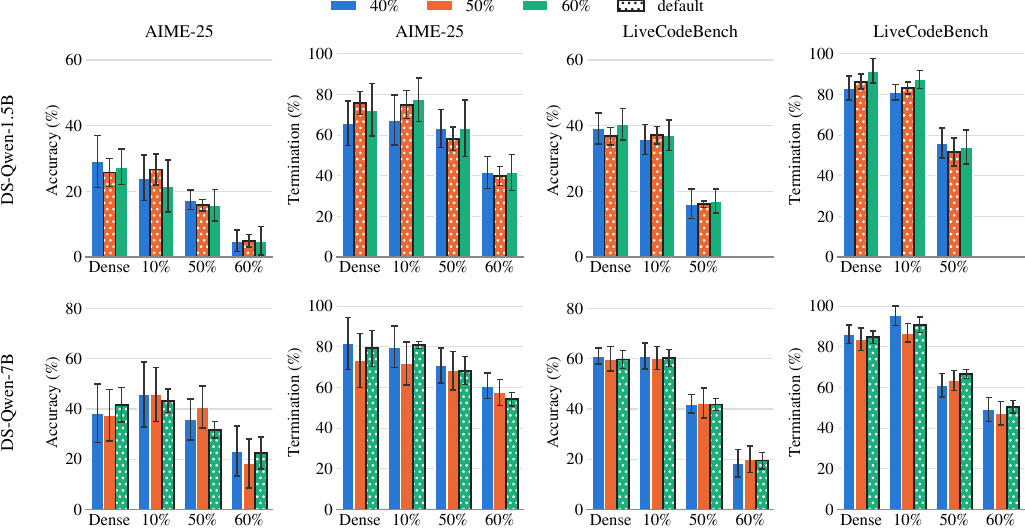}
    \caption{Accuracy and termination rate of DS-Qwen-1.5B and 7B when steered with directions constructed at different sparsity ratios. Error bars show the standard deviation over runs. The three directions give comparable results at every evaluated checkpoint.}
    \label{fig:direction-source}
\end{minipage}\hfill
\begin{minipage}[c]{0.37\textwidth}
    \centering
    \scriptsize
    \setlength{\tabcolsep}{3pt}
    \resizebox{\linewidth}{!}{%
    \begin{tabular}{ll rr lrr}
    \toprule
     & & \multicolumn{2}{c}{Acc (pooled)} & \multicolumn{3}{c}{Acc (best ckpt)} \\
    \cmidrule(lr){3-4}\cmidrule(lr){5-7}
    \thd{Model} & \thd{Bench} & \thd{$\Delta$} & \thd{$p$} & \thd{Ckpt} & \thd{$\Delta$} & \thd{$p$} \\
    \midrule
    \multirow{2}{*}{DS-Qwen-7B} & AIME-25 & +1.5 & 0.171 & \ckpt{10} & +2.9 & 0.261 \\
     & LCB & \textbf{+1.5} & \textbf{0.016} & \ckpt{60} & \textbf{+4.4} & \textbf{0.003} \\
    \midrule
    \multirow{2}{*}{OT3-7B} & AIME-25 & \textbf{+2.1} & \textbf{0.046} & \ckpt{50} & \textbf{+5.4} & \textbf{0.002} \\
     & LCB & \textbf{+1.1} & \textbf{0.026} & \ckpt{60} & \textbf{+2.3} & \textbf{0.030} \\
    \bottomrule
    \end{tabular}}
    \captionof{table}{Significance of TGAS accuracy gains over the base model ($K{=}16$). $\Delta$: TGAS minus base (points); $p$: two-sided problem-level sign-flip test. Pooled: dense, 10\%, 50\% and 60\% checkpoints combined. Best ckpt: largest gain, ties broken by lower $p$. Bold: $p<0.05$.}
    \label{tab:significance}
\end{minipage}
\end{figure*}

\paragraph{Sensitivity to the Sparsity Level of the Direction.} TGAS constructs the direction at one sparsity level and reuses it across checkpoints. To test this choice, we construct additional directions at 40\% and 60\% sparsity for DS-Qwen-1.5B (default: 50\%) and at 40\% and 50\% for DS-Qwen-7B (default: 60\%), following Section~\ref{sec:method_direction}, and evaluate them on the dense, 10\%, 50\% and 60\% checkpoints, selecting $\alpha$ using the same procedure on a held-out OpenR1-Math split. As shown in Figure~\ref{fig:direction-source}, all directions perform comparably: averaged over the eight benchmark-checkpoint settings, accuracy varies by at most $0.6$ points across directions of the same model (20.4--20.9\% for DS-Qwen-1.5B and 40.1--40.7\% for DS-Qwen-7B), and termination by at most $4.1$ points. Larger differences in individual settings appear only on AIME-25 (up to about 9 points for DS-Qwen-7B), whose 30 problems give standard deviations of 8--13 points. TGAS is therefore not sensitive to the sparsity level at which the direction is constructed, consistent with Section~\ref{sec:method_consistency}.

\subsection{Comparison with SEAL}
\label{app:seal-comparison}

\paragraph{TGAS vs SEAL direction similarities.} Table~\ref{tab:seal-tgas} compares TGAS and SEAL directions across layers and model families. All nine comparisons show weak positive alignment, with cosine similarities of 0.146--0.306 under Layer-mean averaging and 0.108--0.269 under Unit-avg averaging. Although alignment is slightly stronger for Qwen-7B and 14B, the results indicate that the two methods share only a limited directional component and are largely distinct.

\paragraph{SEAL with a Direction from the 50\% Checkpoint.} To test whether SEAL is limited by building its direction on the dense model, we also construct its vector on the 50\% checkpoint and tune it in the same way as the dense vector, selecting the steering layer and strength on the held-out OpenR1-Math split (Table~\ref{tab:seal-vec}). Averaged over the 16 settings of DS-Qwen-1.5B and 7B on AIME-25 and LiveCodeBench, accuracy changes only from 26.9\% to 27.6\%, with similar termination (65.9\% and 64.9\%). Both remain below TGAS (30.2\% accuracy at 65.8\% termination), which achieves the best accuracy in 14 of the 15 settings where any method scores above zero. The SEAL direction is thus stable with respect to the checkpoint it is built on, but steering away from reflection recovers less accuracy than steering toward termination.

\begin{table*}[t]
\centering
\scriptsize
\setlength{\tabcolsep}{3pt}
\resizebox{\textwidth}{!}{%
\begin{tabular}{ll rr rr rr rr rr rr}
\toprule
 & & \multicolumn{6}{c}{\thd{AIME-25}} & \multicolumn{6}{c}{\thd{LiveCodeBench}} \\
\cmidrule(lr){3-8}\cmidrule(lr){9-14}
 & & \multicolumn{2}{c}{\thd{SEAL (dense)}} & \multicolumn{2}{c}{\thd{SEAL (50\%)}} & \multicolumn{2}{c}{\thd{TGAS}} & \multicolumn{2}{c}{\thd{SEAL (dense)}} & \multicolumn{2}{c}{\thd{SEAL (50\%)}} & \multicolumn{2}{c}{\thd{TGAS}} \\
\cmidrule(lr){3-4}\cmidrule(lr){5-6}\cmidrule(lr){7-8}\cmidrule(lr){9-10}\cmidrule(lr){11-12}\cmidrule(lr){13-14}
\thd{Model} & \thd{Ckpt} & \thd{Acc} & \thd{Term} & \thd{Acc} & \thd{Term} & \thd{Acc} & \thd{Term} & \thd{Acc} & \thd{Term} & \thd{Acc} & \thd{Term} & \thd{Acc} & \thd{Term} \\
\midrule
\multirow{4}{*}{DS-Qwen-1.5B}
 & \dns      & 18.3\sd{1.9} & 76.7\sd{6.1} & 19.5\sd{2.9} & 75.8\sd{8.3} & \textbf{25.8\sd{4.2}} & 75.8\sd{5.7} & 35.0\sd{3.7} & 84.7\sd{3.3} & 35.5\sd{2.2} & 86.6\sd{3.0} & \textbf{36.8\sd{2.6}} & 86.3\sd{3.7} \\
 & \ckpt{10} & 21.7\sd{4.3} & 80.8\sd{7.4} & 25.0\sd{6.9} & 76.7\sd{8.6} & \textbf{26.7\sd{4.7}} & 75.0\sd{6.9} & 34.5\sd{4.6} & 85.5\sd{3.0} & 33.4\sd{3.6} & 81.3\sd{2.2} & \textbf{37.1\sd{2.6}} & 83.2\sd{2.9} \\
 & \ckpt{50} & 15.8\sd{3.2} & 60.0\sd{6.7} & \textbf{16.7\sd{2.7}} & 57.5\sd{5.7} & 15.8\sd{1.7} & 58.3\sd{5.8} & 13.2\sd{3.5} & 46.3\sd{4.0} & 15.3\sd{1.8} & 45.8\sd{5.0} & \textbf{16.1\sd{1.0}} & 51.8\sd{6.9} \\
 & \ckpt{60} & \textbf{5.0\sd{6.4}} & 27.5\sd{7.4} & 3.3\sd{2.7} & 24.2\sd{7.4} & \textbf{5.0\sd{1.9}} & 40.0\sd{4.7} & 0.0\sd{0.0} & 0.0\sd{0.0} & 0.0\sd{0.0} & 0.0\sd{0.0} & 0.0\sd{0.0} & 2.4\sd{1.8} \\
\midrule
\multirow{4}{*}{DS-Qwen-7B}
 & \dns      & 36.7\sd{2.7} & 80.8\sd{3.2} & 38.3\sd{1.9} & 84.2\sd{6.9} & \textbf{41.7\sd{6.9}} & 79.2\sd{8.8} & 57.9\sd{1.2} & 90.0\sd{2.6} & 56.3\sd{2.5} & 90.5\sd{1.5} & \textbf{59.7\sd{2.2}} & 89.2\sd{3.0} \\
 & \ckpt{10} & 32.5\sd{10.7} & 80.8\sd{5.7} & 32.5\sd{5.7} & 84.2\sd{1.7} & \textbf{43.3\sd{4.7}} & 80.8\sd{1.7} & 55.8\sd{2.6} & 86.1\sd{1.8} & 56.3\sd{0.6} & 86.8\sd{2.2} & \textbf{60.3\sd{3.4}} & 90.8\sd{3.8} \\
 & \ckpt{50} & 29.2\sd{5.7} & 70.8\sd{5.0} & 30.0\sd{2.7} & 70.0\sd{4.7} & \textbf{31.7\sd{3.3}} & 68.3\sd{6.9} & 40.3\sd{1.6} & 75.0\sd{3.9} & 40.4\sd{1.8} & 70.8\sd{1.6} & \textbf{41.8\sd{2.3}} & 66.6\sd{2.2} \\
 & \ckpt{60} & 16.7\sd{4.7} & 60.8\sd{7.9} & 21.7\sd{4.3} & 56.7\sd{0.0} & \textbf{22.5\sd{3.2}} & 54.2\sd{3.2} & 18.4\sd{1.4} & 48.4\sd{2.7} & 16.8\sd{3.4} & 46.8\sd{8.2} & \textbf{19.5\sd{3.3}} & 50.5\sd{3.0} \\
\bottomrule
\end{tabular}}
\caption{SEAL with steering vectors built on the dense model or on the 50\% checkpoint, compared with TGAS. \thd{Acc}: accuracy (\%); \thd{Term}: termination rate (\%). Best accuracy per checkpoint and benchmark in bold.}
\label{tab:seal-vec}
\end{table*}

\subsection{Comparison with Plan-CoT}
\label{app:Plan-CoT-comparison}

Plan-CoT \citep{sheng2025reasoning} shows that LRMs plan their reasoning length before generation: question activations at the \texttt{<think>} position encode a direction whose magnitude scales with problem difficulty and the number of reasoning tokens. The direction is extracted by difference-in-means between hard and easy questions,
\begin{equation*}
\mathbf{r}_{\ell} = \frac{1}{4}\sum_{i=2}^{5}
(\boldsymbol{\mu}_{\ell,i} - \boldsymbol{\mu}_{\ell,1})
\end{equation*}
where
$\boldsymbol{\mu}_{\ell,i}$ is the mean layer-$\ell$ activation at
\texttt{<think>} over MATH questions of difficulty level $i$, and is applied as
\begin{equation*}
\mathbf{h}_{\ell} \leftarrow \mathbf{h}_{\ell} + \lambda\,\mathbf{r}_{\ell}
\end{equation*}
at every layer and token position; $\lambda<0$ shortens reasoning by raising the \texttt{</think>} logit. Plan-CoT therefore differs from TGAS in what the direction contrasts: it captures pre-generation difficulty planning, whereas TGAS contrasts post-branch continuations that terminate with those that loop.

Following the original setup, we construct the Plan-CoT vector on MATH-500 questions grouped by difficulty level, either from the dense model (Plan-CoT$_{\text{dense}}$) or from the 50\%-sparse checkpoint (Plan-CoT$_{50\%}$), and apply it with negative coefficients to counter non-termination. We search $\lambda \in \{-0.5, -0.4, -0.3, -0.2, -0.1\}$ and report the best value per checkpoint and benchmark. The selected Plan-CoT coefficient is either $-0.1$ or $-0.2$, while the selected TGAS strength ranges from $+0.5$ to $+1.5$.

Table~\ref{tab:tgas-vs-plancot} compares the TGAS and Plan-CoT on DeepSeek-R1-Distill-Qwen-7B and OpenThinker3-7B. In accuracy, TGAS is best or tied in 14 of 16 checkpoint benchmark pairs. TGAS also achieves the highest termination rate in 14 of 16 pairs over AIME-25 and LCB, with
the largest gains at pruned checkpoints: on DeepSeek-R1-Distill-Qwen-7B, the termination rate of AIME-25 reaches 68.3\% vs.\ 54.2\% at 50\% sparsity and 54.2\% vs.\ 45.0\% at 60\% sparsity. The Plan-CoT vector built from the 50\%-sparse checkpoint achieves comparable results to that from the dense model, but still underperforms TGAS.

\begin{table*}[t]
\centering
\scriptsize
\setlength{\tabcolsep}{4pt}
\resizebox{0.9\textwidth}{!}{%
\begin{tabular}{ll ccc ccc}
\toprule
& & \multicolumn{3}{c}{\textbf{AIME-25}} & \multicolumn{3}{c}{\textbf{LiveCodeBench}} \\
\cmidrule(lr){3-5} \cmidrule(lr){6-8}
\textbf{Model} & \textbf{Sparsity}
& TGAS & Plan-CoT$_{\text{dense}}$ & Plan-CoT$_{50\%}$
& TGAS & Plan-CoT$_{\text{dense}}$ & Plan-CoT$_{50\%}$ \\
\midrule
\multicolumn{8}{l}{\textit{Accuracy (\%)}} \\
\midrule
\multirow{4}{*}{DS-Qwen-7B}
 & Dense & $\mathbf{41.7}_{\pm 6.9}$ {\scriptsize(+1)}  & $37.5_{\pm 3.2}$ {\scriptsize($-$0.2)} & $39.2_{\pm 5.0}$ {\scriptsize($-$0.2)}
         & $59.7_{\pm 2.2}$ {\scriptsize(+1)} & $58.7_{\pm 2.8}$ {\scriptsize($-$0.1)} & $\mathbf{60.3}_{\pm 3.0}$ {\scriptsize($-$0.2)} \\
 & 10\%  & $\mathbf{43.3}_{\pm 4.7}$ {\scriptsize(+1.25)} & $40.0_{\pm 2.7}$ {\scriptsize($-$0.1)} & $38.3_{\pm 4.3}$ {\scriptsize($-$0.2)}
         & $\mathbf{60.3}_{\pm 3.4}$ {\scriptsize(+1.25)} & $\mathbf{60.3}_{\pm 2.3}$ {\scriptsize($-$0.1)} & $57.9_{\pm 3.5}$ {\scriptsize($-$0.2)} \\
 & 50\%  & $\mathbf{31.7}_{\pm 3.3}$ {\scriptsize(+0.5)} & $29.2_{\pm 3.2}$ {\scriptsize($-$0.1)} & $\mathbf{31.7}_{\pm 7.9}$ {\scriptsize($-$0.2)}
         & $\mathbf{41.8}_{\pm 2.3}$ {\scriptsize(+0.5)}  & $40.3_{\pm 1.3}$ {\scriptsize($-$0.1)} & $40.0_{\pm 3.1}$ {\scriptsize($-$0.2)} \\
 & 60\%  & $22.5_{\pm 3.2}$ {\scriptsize(+1.25)}          & $22.5_{\pm 3.2}$ {\scriptsize($-$0.1)} & $\mathbf{25.0}_{\pm 1.9}$ {\scriptsize($-$0.1)}
         & $\mathbf{19.5}_{\pm 3.3}$ {\scriptsize(+1.25)} & $15.5_{\pm 1.8}$ {\scriptsize($-$0.2)} & $13.7_{\pm 1.2}$ {\scriptsize($-$0.2)} \\
\cmidrule(lr){1-8}
\multirow{4}{*}{OT3-7B}
 & Dense & $\mathbf{45.8}_{\pm 4.2}$ {\scriptsize(+1.5)} & $43.3_{\pm 6.1}$ {\scriptsize($-$0.2)} & $45.0_{\pm 6.4}$ {\scriptsize($-$0.2)}
         & $\mathbf{64.7}_{\pm 1.4}$ {\scriptsize(+1.5)} & $62.9_{\pm 2.5}$ {\scriptsize($-$0.1)} & $62.1_{\pm 2.6}$ {\scriptsize($-$0.1)} \\
 & 10\%  & $\mathbf{48.3}_{\pm 5.8}$ {\scriptsize(+1.5)}  & $44.2_{\pm 3.2}$ {\scriptsize($-$0.1)} & $45.8_{\pm 4.2}$ {\scriptsize($-$0.1)}
         & $\mathbf{63.4}_{\pm 0.5}$ {\scriptsize(+1.5)}  & $62.1_{\pm 1.5}$ {\scriptsize($-$0.1)} & $62.1_{\pm 2.3}$ {\scriptsize($-$0.1)} \\
 & 50\%  & $\mathbf{43.3}_{\pm 8.2}$ {\scriptsize(+1.25)} & $39.2_{\pm 7.4}$ {\scriptsize($-$0.2)} & $39.2_{\pm 1.7}$ {\scriptsize($-$0.1)}
         & $\mathbf{55.5}_{\pm 2.2}$ {\scriptsize(+1.25)} & $52.1_{\pm 2.0}$ {\scriptsize($-$0.1)} & $52.9_{\pm 2.8}$ {\scriptsize($-$0.1)} \\
 & 60\%  & $\mathbf{29.2}_{\pm 1.7}$ {\scriptsize(+1.5)}   & $26.7_{\pm 2.7}$ {\scriptsize($-$0.2)} & $24.2_{\pm 5.7}$ {\scriptsize($-$0.2)}
         & $\mathbf{36.3}_{\pm 3.0}$ {\scriptsize(+1.5)} & $34.5_{\pm 3.5}$ {\scriptsize($-$0.1)} & $34.5_{\pm 2.8}$ {\scriptsize($-$0.1)} \\
\midrule
\multicolumn{8}{l}{\textit{Termination rate (\%)}} \\
\midrule
\multirow{4}{*}{DS-Qwen-7B}
 & Dense & $\mathbf{79.2}_{\pm 8.8}$ {\scriptsize(+1)}  & $75.0_{\pm 5.8}$ {\scriptsize($-$0.2)} & $77.5_{\pm 1.7}$ {\scriptsize($-$0.2)}
         & $\mathbf{89.2}_{\pm 3.0}$ {\scriptsize(+1)} & $86.6_{\pm 2.9}$ {\scriptsize($-$0.1)} & $82.6_{\pm 2.5}$ {\scriptsize($-$0.2)} \\
 & 10\%  & $\mathbf{80.8}_{\pm 1.7}$ {\scriptsize(+1.25)} & $77.5_{\pm 5.7}$ {\scriptsize($-$0.1)} & $75.8_{\pm 6.9}$ {\scriptsize($-$0.2)}
         & $\mathbf{90.8}_{\pm 3.8}$ {\scriptsize(+1.25)} & $85.0_{\pm 0.5}$ {\scriptsize($-$0.1)} & $81.8_{\pm 4.9}$ {\scriptsize($-$0.2)} \\
 & 50\%  & $\mathbf{68.3}_{\pm 6.9}$ {\scriptsize(+0.5)} & $54.2_{\pm 4.2}$ {\scriptsize($-$0.1)} & $52.5_{\pm 6.9}$ {\scriptsize($-$0.2)}
         & $66.6_{\pm 2.2}$ {\scriptsize(+0.5)}  & $\mathbf{67.6}_{\pm 2.0}$ {\scriptsize($-$0.1)} & $63.4_{\pm 3.4}$ {\scriptsize($-$0.2)} \\
 & 60\%  & $\mathbf{54.2}_{\pm 3.2}$ {\scriptsize(+1.25)} & $45.0_{\pm 8.8}$ {\scriptsize($-$0.1)} & $44.2_{\pm 5.7}$ {\scriptsize($-$0.1)}
         & $\mathbf{50.5}_{\pm 3.0}$ {\scriptsize(+1.25)} & $47.4_{\pm 3.1}$ {\scriptsize($-$0.2)} & $42.6_{\pm 3.0}$ {\scriptsize($-$0.2)} \\
\cmidrule(lr){1-8}
\multirow{4}{*}{OT3-7B}
 & Dense & $\mathbf{54.2}_{\pm 3.2}$ {\scriptsize(+1.5)} & $47.5_{\pm 3.2}$ {\scriptsize($-$0.2)} & $50.0_{\pm 7.2}$ {\scriptsize($-$0.2)}
         & $\mathbf{73.9}_{\pm 1.8}$ {\scriptsize(+1.5)} & $72.4_{\pm 3.0}$ {\scriptsize($-$0.1)} & $70.3_{\pm 2.9}$ {\scriptsize($-$0.1)} \\
 & 10\%  & $\mathbf{56.7}_{\pm 5.4}$ {\scriptsize(+1.5)}  & $51.7_{\pm 6.4}$ {\scriptsize($-$0.1)} & $50.8_{\pm 7.4}$ {\scriptsize($-$0.1)}
         & $\mathbf{74.2}_{\pm 1.8}$ {\scriptsize(+1.5)}  & $71.8_{\pm 1.0}$ {\scriptsize($-$0.1)} & $71.6_{\pm 3.6}$ {\scriptsize($-$0.1)} \\
 & 50\%  & $\mathbf{52.5}_{\pm 9.6}$ {\scriptsize(+1.25)} & $47.5_{\pm 8.8}$ {\scriptsize($-$0.2)} & $49.2_{\pm 5.7}$ {\scriptsize($-$0.1)}
         & $\mathbf{69.5}_{\pm 3.0}$ {\scriptsize(+1.25)} & $65.5_{\pm 2.6}$ {\scriptsize($-$0.1)} & $66.1_{\pm 3.6}$ {\scriptsize($-$0.1)} \\
 & 60\%  & $50.8_{\pm 6.9}$ {\scriptsize(+1.5)}            & $\mathbf{51.7}_{\pm 6.4}$ {\scriptsize($-$0.2)} & $48.3_{\pm 7.9}$ {\scriptsize($-$0.2)}
         & $\mathbf{58.9}_{\pm 3.1}$ {\scriptsize(+1.5)} & $56.3_{\pm 3.0}$ {\scriptsize($-$0.1)} & $56.8_{\pm 3.9}$ {\scriptsize($-$0.1)} \\
\bottomrule
\end{tabular}}
\caption{TGAS at its best steering strength $\alpha$ (shown in parentheses) compared with
Plan-CoT at best steering coefficient, using a steering vector extracted from the dense model
(Plan-CoT$_{\text{dense}}$) or from the 50\%-sparse model (Plan-CoT$_{50\%}$). Best value per row and benchmark in bold.}
\label{tab:tgas-vs-plancot}
\end{table*}

\subsection{Statistical Significance of TGAS Gains}
\label{app:significant_test}

\paragraph{Significance testing.} We evaluate TGAS against the base model on AIME-25 and LiveCodeBench (LCB) using DS-Qwen-7B and OT3-7B, with \(K=16\) samples per problem. The steering strength \(\alpha\) is fixed using a held-out OpenR1-Math split. We assess significance using a two-sided problem-level sign-flip permutation test with 20,000 permutations. Pooled results aggregate the dense, OBC at 10\%, 50\% and 60\% sparsity checkpoints. 

\paragraph{TGAS yields significant gains across model-benchmark pairs.}
Table~\ref{tab:significance} shows significant pooled accuracy gains in three of four model--benchmark pairs: DS-Qwen-7B on LCB (+1.5 points, $p=0.016$) and OT3-7B on AIME-25 (+2.1, $p=0.046$) and LCB (+1.1, $p=0.026$). The largest checkpoint-level gains occur at pruned checkpoints, reaching +4.4 points for DS-Qwen-7B at 60\% sparsity on LCB ($p=0.003$) and +5.4 for OT3-7B at 50\% sparsity on AIME-25 ($p=0.002$). The pooled gain for DS-Qwen-7B on AIME-25 is not statistically significant (+1.5, $p=0.171$). For this pair, TGAS primarily affects termination behavior rather than accuracy (Table~\ref{tab:full-main-results-1}).

\paragraph{TGAS reduces generation cost without sacrificing accuracy.}
Shorter generations are only useful if they do not come at the cost of accuracy. Across six models, four checkpoints and three benchmarks, TGAS matches or exceeds the base model in 64 of 72 settings and raises mean accuracy from 46.5\% to 49.1\% (Tables~\ref{tab:full-main-results-1} and~\ref{tab:full-main-results-2}). Both effects grow with sparsity: from the dense model to 60\% sparsity, the mean accuracy gain rises from +1.6 to +3.7 points, while the token reduction rises from 4.4\% to 12.3\%, or 1{,}423 tokens per generation (Table~\ref{tab:tokens-main}). As pruning makes non-termination more frequent, steering towards termination removes more wasted tokens and recovers more answers otherwise lost to the budget. Overall, TGAS makes pruned LRMs cheaper to run while keeping accuracy at or above the base model.

\end{document}